\newif\ifdraft\draftfalse
\newif\ifarxiv\arxivtrue
\documentclass{article}

\usepackage{fullpage}
\usepackage[square, numbers, sort]{natbib}
\usepackage[utf8]{inputenc} % allow utf-8 input
\usepackage[T1]{fontenc}    % use 8-bit T1 fonts
\usepackage{hyperref}       % hyperlinks
\usepackage{booktabs}       % professional-quality tables
\usepackage{longtable}
\usepackage{amsfonts}       % blackboard math symbols
\usepackage{nicefrac}       % compact symbols for 1/2, etc.
\usepackage{microtype}      % microtypography

\usepackage{mathtools}
\usepackage{bbm, bm}
\usepackage{tabularx}

\usepackage{algorithmic} 
\usepackage[ruled]{algorithm2e}

\SetAlFnt{\small}
\SetAlCapFnt{\small}
\SetAlCapNameFnt{\small}
\SetAlCapHSkip{0pt}

\usepackage{amsthm}
\usepackage{authblk}
\usepackage{array, multirow, adjustbox}
\usepackage{arydshln}
\usepackage{blindtext}
\usepackage{bm}
\usepackage{booktabs} % commands to create good-looking tables
\usepackage{comment}
\usepackage{enumitem}
\usepackage{graphicx}
\usepackage{hyperref}
\usepackage{multicol}
\usepackage{subcaption}
\usepackage{tikz} % nice language for creating drawings and diagrams
\usepackage{url}
\usepackage{xcolor}

\newcommand{\RR}{\mathbb{R}} % real numbers
\DeclareMathOperator*{\argmin}{arg\,min}

\newcommand{\cX}{\mathcal{X}} 
\newcommand{\cY}{\mathcal{Y}} 
\newcommand{\cH}{\mathcal{H}} 
 
\newcommand{\xz}{x_0} %x_0
\newcommand{\newx}{x'} %modficiation to x_0
\newcommand{\xstar}{{x^*}} %optimal recourse
\newcommand{\cost}{c}%cost function 
\theoremstyle{plain}
\newtheorem{theorem}{Theorem}[section]
\newtheorem{proposition}[theorem]{Proposition}

\newtheorem{corollary}[theorem]{Corollary}
\theoremstyle{definition}

\newtheorem{assumption}[theorem]{Assumption}
\theoremstyle{remark}

\title{Optimizing Recourse Under Predictive Multiplicity}
\author{Shahin Jabbari\thanks{Drexel University, shahin@drexel.edu}}
\date{}

\begin{document}

\maketitle

\begin{abstract}
\emph{Algorithmic recourse} identifies modifications that can change an unfavorable model prediction into a favorable one. Under \emph{predictive multiplicity}, similarly performing models can disagree on individual predictions, so recourse valid for one model may fail for another. We formulate robust recourse under predictive multiplicity as a budget-constrained minimization of the worst-case validity loss across a competing set of models, which can include models with different architectures or from different hypothesis classes. We propose an oracle-efficient algorithm called ORPM and prove sufficient conditions for its convergence. We evaluate ORPM on one synthetic and five real datasets, and competing sets consist of linear, neural, mixed, and decision-tree models. Our experiments show decreasing optimization-gap bounds and high validity in many tested settings, with favorable performance against baselines designed for specific model classes or architectures. We also quantify how requiring a common action across competing models changes recourse success at a given budget and examine disparities in this success across demographic groups.
\end{abstract}

\section{Introduction}
\label{sec:intro}

\emph{Algorithmic recourse} provides individuals who receive
unfavorable decisions from automated decision-making systems
(e.g., loan denial) with guidance on modifying their features
to achieve a favorable outcome (e.g., loan approval)~\citep{UstunSL19, WachterMR18}.
In practice, decision-makers may have access to competing
set of models with similar predictive performance but different
decisions for specific individuals, a phenomenon known as
\emph{predictive multiplicity}~\citep{breiman2001statistical, MarxCU20}.
Therefore, for an individual, the modifications 
can depend on which similarly accurate model a decision-maker selects. 
Recourse computed for one model may therefore become invalid
for other models in the set, undermining its
reliability~\citep{PawelczykBK20a}. 

\emph{Robust algorithmic recourse} addresses this
unreliability by accounting for model
uncertainty~\citep{UpadhyayJL21, GuoJC+23, NguyenBN23}.
Some approaches consider bounded changes in model
parameters~\citep{UpadhyayJL21}, while others target
particular model classes~\citep{DuttaLMTM22}.
Recently, \citet{TurbalVS25} studied recourse under 
model multiplicity and proposed an ellipsoidal
approximation of the Rashomon set for linear models and
neural networks sharing a fixed feature extractor, with
variation confined to the last-layer parameters. 
However, competing models may have different architectures
or even belong to different classes~\citep{BehzadCDT25}.
This motivates designing a framework that directly considers 
diverse competing sets without requiring a shared architecture or
closeness in parameters.

\ifarxiv\paragraph{Our Contributions and Results.}
\else
\noindent\textbf{Our Contributions and Results.}
\fi
We formulate robust recourse under predictive multiplicity 
as a budget-constrained
minimization of the worst-case validity loss across
competing models, allowing models with different architectures or
from different classes. To solve this problem, we develop an oracle 
efficient algorithm, called
\emph{Oracle-Efficient Robust Recourse under Predictive
Multiplicity} (ORPM), which is inspired by dynamics for computing
equilibria in two-player games~\citep{FreundS96}.
Under bounded losses and controlled global oracle error,
we prove convergence to an approximate equilibrium for
a uniform mixture of oracle responses. In the convex
setting, the same guarantee applies to their average point.
We show how an oracle can be designed for differentiable and tree-based competing sets.

Empirically, using one synthetic and five real-world datasets, we
evaluate ORPM on linear, neural, mixed, and tree-based competing sets.
Our convergence experiments show that validity 
can be achieved before the prescribed optimization-gap target is certified. 
ORPM achieves high joint validity in many tested settings
and often outperforms or matches prior approaches, even in settings
where the competing set only contains models with the same architecture.

We further examine how the recourse budget and size or complexity of the competing set affect validity. We observe that larger budgets generally improve validity, while acceptance by more diverse models can substantially reduce it. On competing sets that contain models with the same model class and architecture, we show that ORPM competes favorably with prior approaches that are specialized for such competing sets. We also empirically study how multiplicity affects the validity and disparity of recourse. These results characterize both ORPM's practical strengths and highlight the challenges of providing recourse that is robust to predictive multiplicity.

\section{Related Work}
\label{sec:related-work}

\ifarxiv\paragraph{Algorithmic Recourse.}
\else
\noindent\textbf{Algorithmic Recourse.}
\fi
Algorithmic recourse provides low-cost modifications needed to change a model's prediction for an input with an undesirable label~\citep{WachterMR18, UstunSL19}. Since its introduction, researchers have proposed various recourse formulations and studied aspects such as feasibility, fairness, improvement, and competition~\citep{PawelczykDHKL23, KarimiBBV20, PoyiadziSSBF20, KennyAB+26, CheonWF+25, Russell19, Jabbari26a, SegalGD24, KonigFF+25, KonigFG23, AvasaralaGJ+26}. See~\citep{VermaDH20, KarimiBSV23} for surveys.

\ifarxiv\paragraph{Predictive Multiplicity.}
\else
\noindent\textbf{Predictive Multiplicity.}
\fi
Predictive multiplicity refers to the existence of multiple models that achieve similar predictive performance yet differ in their individual-level predictions~\citep{FisherRD19, DAmourHM+22, SemenovaR19, SemenovaRP22}. A Rashomon set is a set of models from the same model class that perform similarly to a reference model~\citep{breiman2001statistical}. In this work, we call the finite collection of similar-performing models from potentially different model classes the competing set.~\citet{MarxCU20} formalize this phenomenon for classification, and~\citet{Watson-DanielsPU23} extend the analysis to probabilistic classifiers.  Subsequent works quantify predictive multiplicity and the size of Rashomon sets~\citep{HsuC22, SemenovaRP22}, efficiently explore Rashomon sets~\citep{HsuLH024, XinZC+22} and study the effects of multiplicity in downstream tasks~\citep{DuNW25, JainWC+25}.

\ifarxiv\paragraph{Recourse Under Predictive Multiplicity.}
\else
\noindent\textbf{Recourse Under Predictive Multiplicity.}
\fi
Several works study the challenges posed by predictive multiplicity for algorithmic recourse~\citep{PoyiadziSSBF20, PawelczykBK20a, AlfanoGL+25}.~\citet{TurbalVS25} study robust recourse over an ellipsoidal approximation of the Rashomon set of linear models and neural networks with a fixed feature extractor and different last-layer parameters.~\citet{LeofanteBR23} define robust counterfactuals for models with compatible inputs and outputs. Their product-network construction focuses on homogeneous neural networks, with a complexity extension to piecewise-linear models.~\citet{NooraniDHD25} study ensemble recourse through cost and worst-case loss and introduce an entropic-risk formulation whose limiting case recovers worst-case robustness. We study budget-constrained worst-case loss; our approach is oracle-efficient and can be applied to a competing set that contains models from different model classes. We compare our approach to~\cite{TurbalVS25} and a baseline (MinL2) that adapts the joint neural feasibility formulation~\cite{LeofanteBR23}.

\ifarxiv\paragraph{Robust Recourse.}
\else
\noindent\textbf{Robust Recourse.}
\fi
Robust recourse studies how the validity of recourse can be preserved under model uncertainty. An early influential paper proposes ROAR, designed to find valid recourses under bounded shifts in model parameters~\citep{UpadhyayJL21}. Similar to recourse, various formulations for robust recourse have been explored~\citep{YetukuriHVUL24, JiangLR+24a, NguyenBN23, CheonWF+25, HammanNMMD23, GuoJC+23, KayasthaGJ26, KyawKJ25, KonigFF+25, JiangLL+23, BuiNYN25, NguyenBN+22}. Specifically, for tree-based models, OCEAN uses mixed-integer programs to generate recourse for a fixed tree ensemble~\citep{ParmentierV21}, while FOCUS uses differentiable approximations, enabling gradient-based search~\citep{LucicOH+22}. Robustness is addressed through a stability criterion (RobX)~\citep{DuttaLMTM22} and probabilistic constraints to target the randomness of the ensemble~\citep{ForelPV24}. See~\citep{JiangLR+24a} for a survey. Our formulation permits a specified finite set of compatible models from different classes without requiring proximity in parameter space. Our contribution concerns the budgeted formulation, oracle implementations, conditional guarantees, and empirical analysis, rather than introducing heterogeneous-model robustness itself. We compare our approach to many of these prior robust approaches in Section~\ref{sec:exp}.

\ifarxiv\paragraph{Equilibrium Computation in Two-Player Games.}
\else
\noindent\textbf{Equilibrium Computation in Two-Player Games.}
\fi
Computing equilibrium using dynamics involving best-response and no-regret approaches has a long history, including the celebrated work of~\citet{FreundS96} and subsequent works~\citep{RakhlinS13, DaskalakisISZ18, AroraHK12}. The oracle-efficient approach has also been extensively studied in the algorithmic fairness literature~\citep{KearnsNRW18, AgarwalBD+18, Sharifi-MalvajerdiKR19, DianaGG+21}. Our analysis specializes the oracle-based robust-optimization framework of~\cite{ChenLSS17} to recourse.
\section{Problem Formulation}
\label{sec:problem-formulation}
Consider a binary classification task where the input space is $\cX \subseteq \RR^d$ (representing individuals), and the label space is $\cY = \{ 0,1 \}$, where label $0$ ($1$) denotes an unfavorable (favorable) outcome, e.g., loan denial (loan approval).  Let $D$ denote the distribution over $\cX \times \cY$. Let $\cH \subseteq [0,1]^{\cX}$ be the hypothesis class of probabilistic models, and $h\in \cH$ be a probabilistic classifier deployed by a decision-maker, where $h: \cX\to[0,1]$. The probability output of the classifier $h$ can be converted to a deterministic label in $\cY$ by thresholding, and we use $\hat{y}_h(x)$ to denote this deterministic label.

Let $\cost: \cX \times \cX \to \RR^{\geq 0}$ be a cost function quantifying the cost of modifying any input to another. We assume $\cost(x,x)=0$. For any given $\xz\in \cX$ and $h\in \cH$, if the $\hat{y}_h(\xz)=0$, then the decision-maker's goal in algorithmic recourse~\citep{UstunSL19, WachterMR18} is to find an $\newx\in \cX$ such that $\hat{y}_h(\newx)=1$ (i.e., a \emph{valid} recourse) and $c(\xz, \newx)\leq \delta$, where $\delta > 0$.  In this formulation, $\delta$ is a (potentially instance-dependent) budget parameter and demonstrates the maximum cost an individual is willing and able to incur to receive the favorable label. This formulation captures the limits on the resources available for implementing recourse, and even the least costly valid recourse may be unaffordable. We treat $\delta$ as fixed for each optimization but vary $\delta$ across experiments to study the trade-off between cost and validity~\cite{PawelczykLBK23}.

We consider a decision-maker that uses a budget-constrained relaxed formulation of recourse~\citep{UpadhyayJL21}:
\begin{align}
    \label{eq:recourse}
    \newx \in \argmin_{x \in \cX} \ell(h(x), 1) \text{ s.t. } \cost(\xz, x) \leq \delta,
\end{align}
where $\ell: \RR \times \RR \to \RR^{\geq 0}$ is a surrogate loss function chosen such that $\ell(p, 1)$ is strictly decreasing in $p\in [0,1]$ (e.g., squared loss). Therefore, minimizing the loss is a \emph{proxy} for maximizing validity.

In the presence of model multiplicity~\citep{breiman2001statistical, MarxCU20}, there exists a set of competing models that can be used by the decision-maker. We assume this set has $n$ models and use $H = \{h_1, \ldots, h_n\} \subseteq \cH$ to refer to it. Our formulation allows a finite but otherwise arbitrary competing set. In particular, we allow the models in $H$ to belong to different model classes. Let $h \in H$ denote the current model used by the decision-maker to label individual $\xz \in \cX$. When $\hat{y}_h(\xz)=0$, the goal in recourse in the presence of model multiplicity is \emph{ideally} to find a \emph{robust} recourse $\xstar\in \cX$ that is valid under every model in the competing set, i.e., to satisfy ${\hat{y}}_{h}(\xstar)=1$ for all $h\in H$, and again $\xstar$ needs to satisfy $\cost(\xz, \xstar)\leq \delta$. 

However, such a valid recourse need not exist: either because the budget may be insufficient to reach a favorable point for all models or the favorable decision regions of different models in the competing set may have no common intersection. Despite this, similar to the case of a single classifier, a relaxed version of the problem can be written as:
\begin{align}
    \label{eq:recourse-pure-strategy}
    \xstar \in \argmin_{x \in \cX} \max_{h \in H} \ell(h(x), 1) \text{ s.t. } \cost(\xz, x) \leq \delta,
\end{align}
where $\ell$ is the surrogate loss defined above. Minimizing the first term in Equation~\ref{eq:recourse-pure-strategy} is a \emph{proxy} for maximizing the worst-case validity across all competing models. Under the monotonicity assumption, this is equivalent to maximizing the lowest score across all models in the competing set, within the budget. We emphasize that while the main motivation of our work is to generate a recourse robust to model multiplicity, our formulation does not require architectural similarity or parameter-space proximity among the models in $H$.

\section{Algorithm and Analysis}
\label{sec:algorithm}

To solve the robust recourse problem defined in Equation~\ref{eq:recourse-pure-strategy}, we model the optimization as a zero-sum game between the decision-maker who generates the recourse and an adversary who selects the worst-case model for the generated recourse. Let $K:=\{x\in\cX:\cost(\xz,x)\leq\delta\}$ denote the feasible recourse set and $\mathcal{L}(x,w):=\sum_{i=1}^n w_i\ell(h_i(x),1).$ We allow the adversary to select a distribution $w\in\Delta(H)=\{w\in\RR^n:w_i\geq0,\ \sum_{i=1}^n w_i=1\}$ over the competing models and solve
$\label{eq:recourse-mixed-strategy}
\xstar\in\argmin_{x\in K}\max_{w\in\Delta(H)}\mathcal{L}(x,w)$ instead.
This is without loss of generality, as the objective is linear in $w$, so its maximum over the simplex is attained by placing all mass on a worst-case model.
Our approach separates the game dynamics from the optimization of recourse for a fixed model distribution, 
utilizing an approximate oracle.

\begin{assumption}[Approximate Oracle]
\label{assumpt:oracle}
Given any distribution $w\in\Delta(H)$, an
$\epsilon$-approximate oracle
$\mathcal{O}_{\xz,\ell,\cost}$ returns a feasible recourse
$\tilde{x}\in K$ satisfying
$
\label{eq:oracle}
\mathcal{L}(\tilde{x},w)
\leq
\inf_{x\in K}\mathcal{L}(x,w)+\epsilon,
$
where $\epsilon\geq0$ bounds the additive global
error.
\end{assumption}

For notational convenience, we suppress the dependence
of the oracle on $H$, $\delta$, and $\epsilon$. 
The error bound concerns global optimality and may be 
established analytically or using
optimization certificates.

We propose the \textbf{O}racle-Efficient Robust \textbf{R}ecourse under \textbf{P}redictive \textbf{M}ultiplicity (ORPM) algorithm. In each round, the decision-maker plays an approximate best response to $w_t$ by invoking the oracle (line~\ref{alg:bestresponse}). The adversary then uses multiplicative weights~\citep{AroraHK12} to shift probability mass toward models with higher loss at the current response (lines~\ref{alg:update-v}--\ref{alg:normalize}).

\begin{algorithm}[ht!]
\caption{\textbf{O}racle-Efficient Robust \textbf{R}ecourse under \textbf{P}redictive \textbf{M}ultiplicity (ORPM)}
\label{alg:orpm}
\begin{algorithmic}[1]
\STATE \textbf{Inputs:} Initial instance $\xz$, loss $\ell$, cost $\cost$, competing set $H=\{h_1,\ldots,h_n\}$, oracle $\mathcal{O}_{\xz,\ell,\cost}$, budget $\delta$, learning rate $\eta$, number of rounds $T$.
\STATE \textbf{Initialize:} $w_{1,i}=1/n$ for $i=1,\ldots,n$.
\FOR{$t=1$ to $T$}
    \STATE $x_t\leftarrow\mathcal{O}_{\xz,\ell,\cost}(w_t)$. \label{alg:bestresponse}\hfill{\hspace{-5mm}$\triangleright$ Query the oracle for the optimal recourse for current weights}
    \STATE $v_{t,i}\leftarrow\ell(h_i(x_t),1)$ for $i=1,\ldots,n$. \label{alg:update-v}\hfill{$\triangleright$ Compute the loss vector for the generated recourse}
    \STATE $\widetilde{w}_{t+1,i}\leftarrow w_{t,i}\exp(\eta v_{t,i})$ for $i=1,\ldots,n$. \label{alg:update-weights}\hfill{$\triangleright$ Update weights to reward  models with high loss}
    \STATE $w_{t+1,i}\leftarrow\widetilde{w}_{t+1,i}/\sum_{j=1}^n\widetilde{w}_{t+1,j}$ for $i=1,\ldots,n$. \label{alg:normalize}\hfill{$\triangleright$ Normalize the weights to sum to 1}
\ENDFOR
\STATE \textbf{Return:} Response sequence $(x_1,\ldots,x_T)$ and $\bar{w}=T^{-1}\sum_{t=1}^T w_t$; obtain a deterministic recourse using the return rules below.
\end{algorithmic}
\end{algorithm}
ORPM makes $T$ oracle calls, $nT$ model-loss evaluations, and $O(T(n+d))$ additional arithmetic operations. The cost of each oracle call depends on the competing set and the weighted recourse problem.

\ifarxiv\paragraph{Convergence Guarantees.}
\else
\noindent\textbf{Convergence Guarantees.}
\fi
Let $x_{1:T}:=(x_1,\ldots,x_T)$ denote the sequence of oracle
responses, and let $\bar{x}:=T^{-1}\sum_{t=1}^T x_t$ and
$\bar{w}:=T^{-1}\sum_{t=1}^T w_t$.
Also let $\epsilon_t$ denote the global error bound for the
oracle call at round $t$, i.e.,
$\mathcal{L}(x_t,w_t)
\leq
\inf_{x\in K}\mathcal{L}(x,w_t)+\epsilon_t.$
We allow these bounds to vary across calls. If the same
accuracy guarantee $\epsilon$ holds for every call, then
$\epsilon_t\leq\epsilon$ for every $t$.
Define the mixture gap as:
$\operatorname{Gap}_{\mathrm{mix}}(x_{1:T},\bar{w})
:=
\max_{w\in\Delta(H)}
\tfrac1T\sum_{t=1}^T\mathcal{L}(x_t,w)
-
\inf_{x\in K}\mathcal{L}(x,\bar{w}).
%\notag\\
%&=
%\max_{1\leq i\leq n}
%\frac1T\sum_{t=1}^T\ell(h_i(x_t),1)
%-
%\inf_{x\in K}\mathcal{L}(x,\bar{w}).
$
Consider the strategy pair in which the decision-maker
selects one of $x_1,\ldots,x_T$ uniformly at random and
the adversary independently selects a model according
to $\bar{w}$. The mixture gap measures the sum of the
players' possible improvements: the adversary can increase
the expected loss by changing its distribution,
while the decision-maker can decrease it by changing its
recourse. In each case, the other player's
strategy remains fixed. We 
require the following assumption.

\begin{assumption}[Bounded Loss]
\label{assumpt:bounded-loss}
There exists $B>0$ such that $\ell(h_i(x),1)\in[0,B]$
for every $x\in K$ and $i\in \{1, \ldots, n\}$.
\end{assumption}
For probabilistic predictions, squared loss satisfies this
assumption with $B=1$. Our main result is as follows.

\begin{theorem}[Oracle-Dependent Convergence]
\label{thm:convergence}
Under Assumptions~\ref{assumpt:oracle}
and~\ref{assumpt:bounded-loss}, let
$\bar\epsilon_T:=T^{-1}\sum_{t=1}^T\epsilon_t$.
For $n\geq2$ and any $\eta>0$, the responses and average
weights produced by Algorithm~\ref{alg:orpm} satisfy
$
\label{eq:general-regret-bound}
\operatorname{Gap}_{\mathrm{mix}}(x_{1:T},\bar{w})
\leq
\tfrac{\ln n}{\eta T}
+\tfrac{\eta B^2}{8}
+\bar\epsilon_T.
$
In particular, setting $\eta=B^{-1}\sqrt{\tfrac{\ln n}{T}}$ gives
$
\label{eq:mixture-rate}
\operatorname{Gap}_{\mathrm{mix}}(x_{1:T},\bar{w})
\leq
2B\sqrt{\tfrac{\ln n}{T}}+\bar\epsilon_T.
$
For $n=1$, the mixture gap is at most $\bar\epsilon_T$.
\end{theorem}

All proofs are deferred to Appendix~\ref{sec:appx-proof}. 
Theorem~\ref{thm:convergence} does not require convexity of the models' losses
or the feasible set. It applies to nonconvex problems whenever
the oracle has controlled global error (e.g.,
oracles with suitable optimization certificates). With a
uniform error bound $\epsilon$, the gap is at most
$\epsilon+O(\sqrt{\ln n/T})$. Allowing the
decision-maker to randomize can improve the objective relative
to any single recourse, but this guarantee alone
does not establish convergence to an optimal deterministic
recourse. We next study the \emph{convex setting} where we can guarantee
convergence to deterministic recourse. We need the following
assumption.

\begin{assumption}[Convexity and Compactness]
\label{assumpt:convex-compact}
The feasible recourse set $K$ is compact and convex, and
the mapping $x\mapsto\ell(h_i(x),1)$ is continuous and convex on $K$
for every $i\in\{1,\ldots,n\}$.
\end{assumption}
Norm-based cost constraints and convex feature restrictions
yield convex feasible sets. Under
Assumption~\ref{assumpt:convex-compact}, together with
linearity in $w$, the minimax theorem guarantees an
equilibrium for the deterministic recourse game~\citep{Sion1958}.
Define the point gap as
$\label{eq:daulity-gap}
\operatorname{Gap}(x,w)
:=
\max_{u\in\Delta(H)}\mathcal{L}(x,u)
-
\inf_{z\in K}\mathcal{L}(z,w).$

\begin{corollary}[Convergence of the Average Recourse]
\label{cor:convex-convergence}
Suppose Assumptions~\ref{assumpt:oracle},
\ref{assumpt:bounded-loss}, and
\ref{assumpt:convex-compact} hold. For $n\geq2$, run Algorithm~\ref{alg:orpm} with $\eta=B^{-1}\sqrt{\tfrac{\ln n}{T}}$. 
Then the average
$\bar{x}=T^{-1}\sum_{t=1}^T x_t$ of the response sequence returned by Algorithm~\ref{alg:orpm} is feasible and satisfies
$
\label{eq:convex-point-rate}
\operatorname{Gap}(\bar{x},\bar{w})
\leq
\operatorname{Gap}_{\mathrm{mix}}(x_{1:T},\bar{w})
\leq
2B\sqrt{\tfrac{\ln n}{T}}+\bar\epsilon_T.
$
\end{corollary}

The distinction between Theorem~\ref{thm:convergence} and Corollary~\ref{cor:convex-convergence} separates
the role of global oracle accuracy from that of convexity:
the former controls convergence of the response distribution,
while the latter allows us to return its average responses 
with the same guarantee. In Corollary~\ref{cor:convex-convergence}, if $\epsilon_t\leq\epsilon$ for all $t$, the final term
may be replaced by $\epsilon$.

\ifarxiv\paragraph{Oracle Design.}
\else
\noindent\textbf{Oracle Design.}
\fi
We next describe how an oracle can be implemented for convex and specific
non-convex settings. Additional details can be found in Appendix~\ref{sec:app-exp-details}.

\begin{enumerate}
    \item \textbf{Convex Setting}: 
For affine logits, 
both squared and logistic loss are strictly decreasing functions of the logit. 
Consequently, their deterministic minimax problems have the same optimizers and admit the exact reformulation 
(maximize the smallest logit across the competing models)
that, combined with an $L_2$ budget and affine feature constraints, leads to a second-order cone program~\cite{BV2014}. 
Logistic loss also makes each weighted response problem convex, allowing supporting-plane lower bounds to certify its 
numerical optimization error. Weighted squared-probability loss need not be convex, and replacing it changes the response problems even though the deterministic minimax optimizers coincide.
\item \textbf{Differentiable Nonconvex Setting}:
For differentiable nonconvex competing sets, our implementation uses local SLSQP response optimization. 
These local solves do not establish the global oracle-error bounds 
required by Theorem~\ref{thm:convergence}. 
We therefore assess their performance empirically.
See Section~\ref{sec:exp}.

\item \textbf{Tree-Based Nonconvex Setting}:
Decision trees have piecewise-constant
predictions, so their input gradients do not provide useful
search directions. Following the mixed-integer modeling
approach of~\citet{ParmentierV21}, for competing sets containing decision trees, our oracle encodes the
competing tree models jointly and minimizes their weighted
loss subject to budget and feature constraints. If a
feasible response has weighted loss
$u_t=\mathcal{L}(x_t,w_t)$ and the solver provides a valid
global lower bound $b_t$, then $\epsilon_t=u_t-b_t$ is a
global oracle-error bound. Thus, even a time-limited solver
can support Theorem~\ref{thm:convergence}, although the
resulting bound may be loose. The solver's bounds apply to the budget and feature
constraints represented in the optimization problem,
up to its numerical tolerances. For a deterministic tree-based recourse, we return the
candidate with the smallest worst-model loss among the
original input and all oracle responses, breaking ties
by cost. Combining this candidate's worst-model loss
with the solver's global lower bounds lets us bound
how far its loss is from the optimal deterministic
recourse loss, without requiring convexity. See Appendices~\ref{sec:appx-proof}~and~\ref{sec:app-exp-additional}.

\end{enumerate}
\section{Experiments}
\label{sec:exp}
We empirically study the performance of ORPM in convex and non-convex settings. Code and data are available 
\ifarxiv\href{https://github.com/shahin-jabbari/RecoursePredictiveMultiplicity}{here}. 
\else\href{https://anonymous.4open.science/r/RecoursePredictiveMultiplicity-3A2C/README.txt}{here}. 
\fi

\subsection{Datasets}
\label{sec:exp-data}
We used six datasets: one synthetic dataset and five real-world datasets. The synthetic dataset is two-dimensional. There are equal observations for each class, and they are generated from a multivariate normal distribution with means $\mu_0=[-3,-3]$ and $\mu_1=[3,3]$, respectively, and a shared covariance matrix $\Sigma=5\mathbb{I}$. The real-world datasets include Correctional Offender Management Profiling for Alternative Sanctions (\href{https://github.com/propublica/compas-analysis}{COMPAS}), Polish Companies Bankruptcy (\href{https://archive.ics.uci.edu/dataset/365/polish+companies+bankruptcy+data}{Polish}), German Credit (\href{https://archive.ics.uci.edu/dataset/144/statlog+german+credit+data}{German}), Give Me Some Credit (\href{https://www.kaggle.com/competitions/GiveMeSomeCredit/data}{GMC}), and ACSIncome~\citep{DingHM+21} from \href{https://github.com/socialfoundations/folktables}{Folktables}.
The datasets range from 2 to 67 features. For COMPAS, Polish, and German, we build on the preprocessing code associated with~\citep{TurbalVS25}. 
Detailed information about the datasets, splitting, and standardization procedures is described in Appendix~\ref{sec:app-exp-datasets}.

\subsection{Implementation Details}
\label{sec:exp-implementation}
We used logistic regression, neural networks with one or two hidden
layers, and decision trees as models. We constructed models for the competing
sets using randomized model
training (on a subset of features to generate diversity) or from a Hessian-based ellipsoidal approximation to the Rashomon set~\citep{TurbalVS25}.
We then filtered these models by validation accuracy to form competing sets. Evaluation used
repeated stratified training, validation, and test splits, with feature
standardization fitted on the training split. We generated recourse only
for held-out instances rejected by at least half of the models in the competing set.

We used the $L^2$-norm in standardized feature space as the recourse
cost $\cost$, enforcing no changes on dataset-specific immutable features. 
The differentiable trade-off and same-architecture neural experiments use raw logistic loss and local SLSQP response optimization. 
Tree experiments use squared probability loss and time-limited mixed-integer response optimization. 
The convex duality-gap experiment uses normalized logistic loss. Budgets and compositions/size of the competing set
are specified with each experiment. Unless stated otherwise, experiments use 10 seeds and 20 instances per seed. 
Results are averaged first
across instances within each seed and then across seeds. 
For the stratified cost/disparity results, we weight subgroup success rates by the 
study-specific group proportions within each seed, then average equally across seeds. 
These weighted estimates need not occur in fixed increments of 0.05 or 0.005.
See Appendix~\ref{sec:app-exp-details} for more details.

\ifarxiv\paragraph{Baselines.}
\else
\noindent\textbf{Baselines.}
\fi
We next discuss the main baselines used in our trade-off experiments. We also use additional experiment-specific baselines, 
which we introduce throughout Section~\ref{sec:exp-findings}.
For differentiable competing sets, we compared ORPM with 3 baselines: 
(1) a budget-constrained
implementation of ROAR~\citep{UpadhyayJL21}, which optimizes against a bounded parameter uncertainty set,
(2) ElliCE~\citep{TurbalVS25}, which uses a Hessian-based ellipsoid to approximate the Rashomon set, and (3) ADV,
a baseline built on the idea of adversarial training that optimizes against the largest
loss among the competing models. For neural networks, ROAR uses local linear logit surrogates, 
whereas ElliCE considers last-layer parameter uncertainty with the feature extractor fixed. 
For tree-based competing sets, we compared ORPM with 4 baselines: (1) ADV-surrogate, which is a differentiable tree approximations inspired by FOCUS~\citep{LucicOH+22}, (2) ROAR-surrogate, which applies ROAR to local linear logit surrogates, (3) RobX-constrained which adapts RobX~\citep{DuttaLMTM22} to the budget and immutable-feature constraints, and (4) OCEAN-style which implements minimum-cost recourse for the deployed model using the mixed-integer formulation approach of~\citep{ParmentierV21}. 
See Appendix~\ref{sec:app-exp-details} for more details.

\subsection{Findings}
\label{sec:exp-findings}
\ifarxiv\paragraph{Duality Gap.}
\else
\noindent\textbf{Duality Gap.}
\fi
\label{sec:exp-duality}
We ran ORPM on all datasets using 5 seeds and 5 instances per seed for 1000 rounds. We experimented on competing sets of size 6, 12, and 24 and three types of competing sets: the first type only contains logistic regression models (linear), the second contains an equal number of one-layer and two-layer neural networks (neural), and the third contains an equal number of logistic regression, one-layer, and two-layer neural networks (mixed). We used different budgets for each dataset. See Appendix~\ref{sec:app-exp-duality}. Upon termination and after several checkpoint rounds, we measured the validity of recourse and upper and lower bounds on the averaged-point gap (using, e.g., convex bounds) with a gap of $\epsilon=0.005$ as the target. 

\begin{figure}[htbp]
\centering
\begin{subfigure}[t]{0.49\textwidth}
\centering\includegraphics[width=\linewidth]{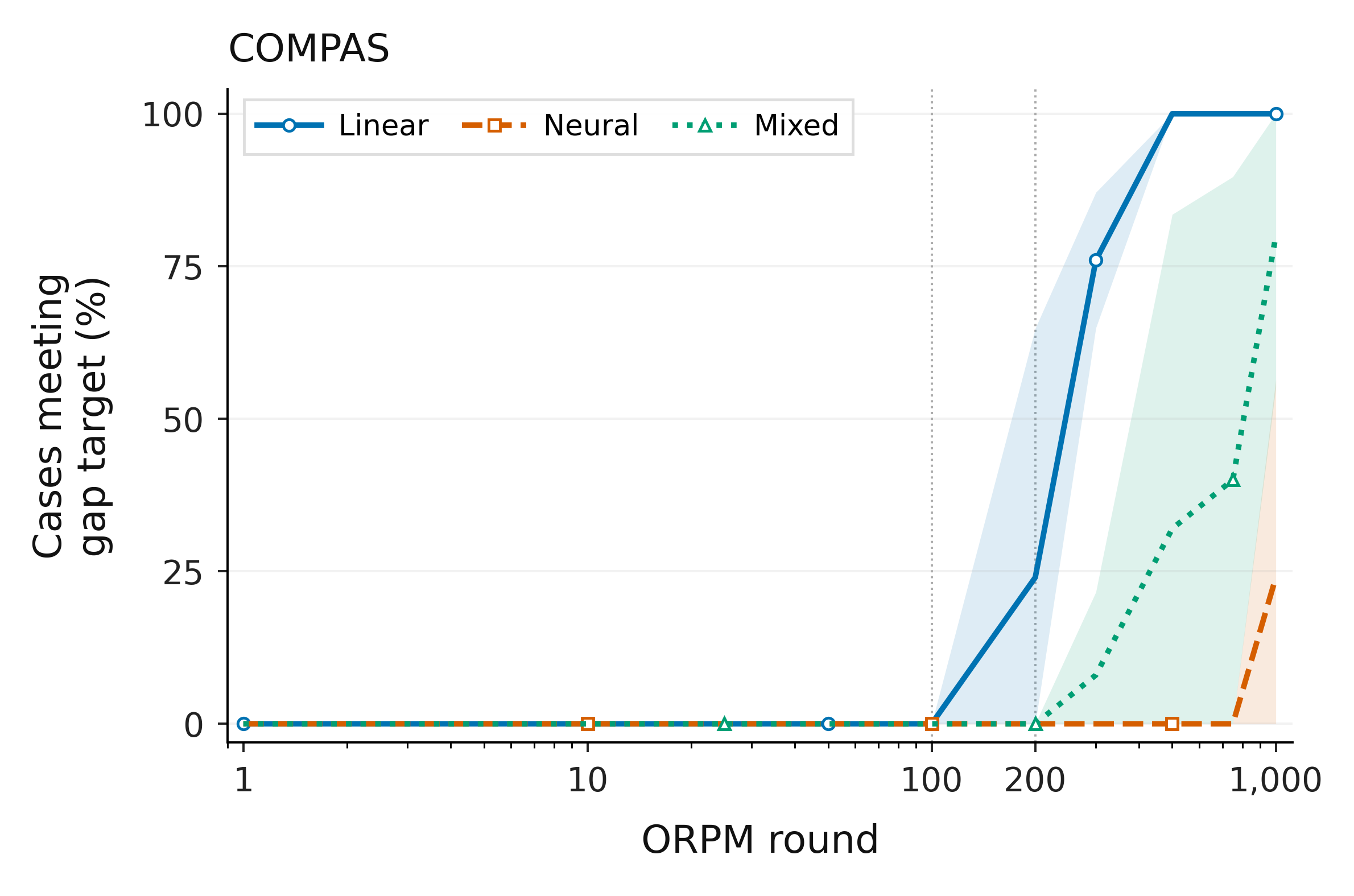}
\caption{COMPAS dataset, 6 competing models, $\delta=2$. \label{fig:conv-compas}}
\end{subfigure}\hfill
\begin{subfigure}[t]{0.49\textwidth}
\centering\includegraphics[width=\linewidth]{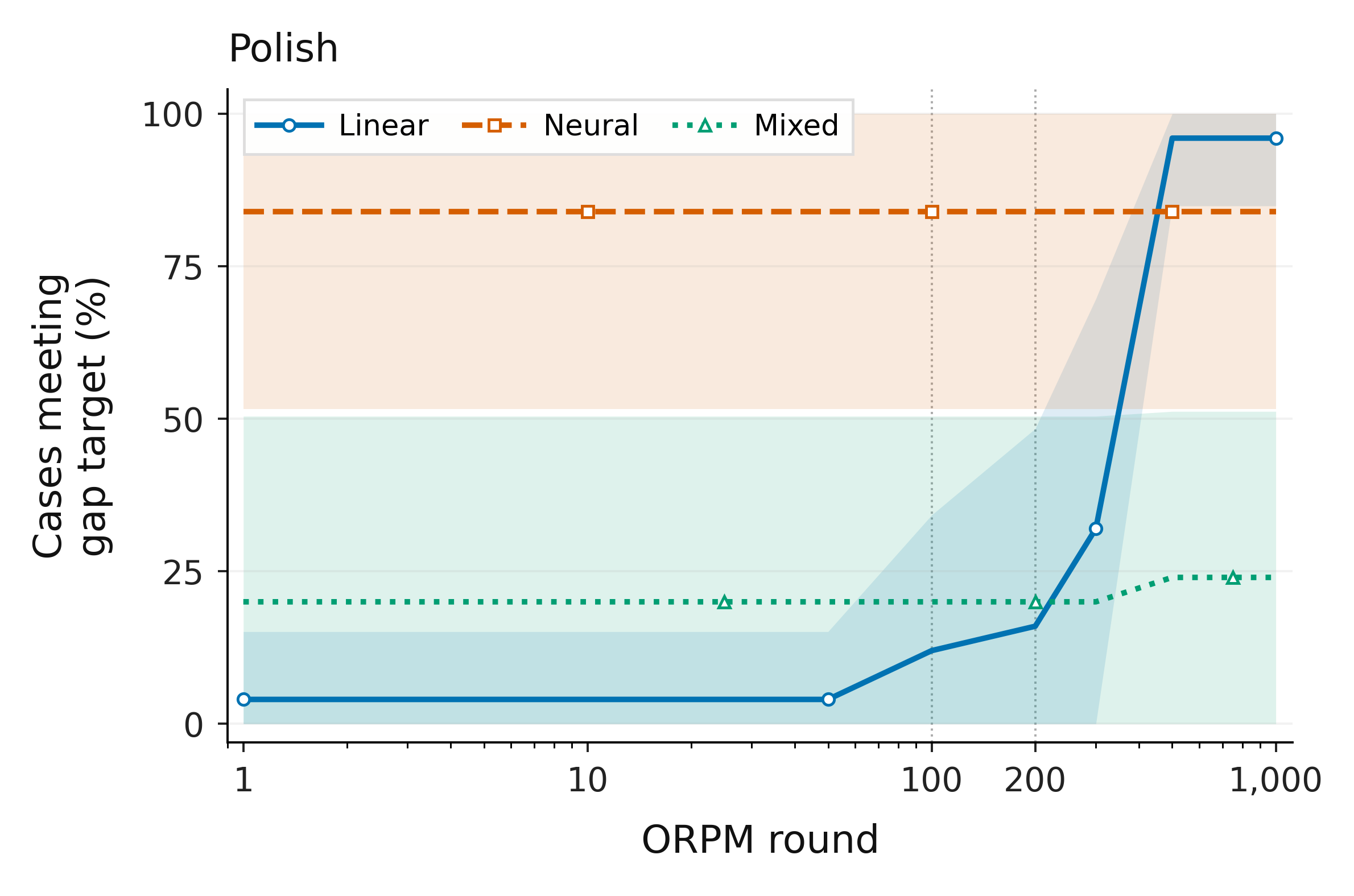}
\caption{Polish dataset, 24 competing models, $\delta=2$.}
\end{subfigure}
\caption{Fraction of instances whose certified upper bound on the averaged-point gap is at most 0.005 at each checkpoint. Each subfigure corresponds to a different dataset. The number of models and budget are mentioned in the caption.\label{fig:conv-main}}
\end{figure}

Figure~\ref{fig:conv-main} depicts the fraction of instances that achieve their upper bound on the gap and meet this stringent threshold. This fraction is averaged over seeds and instances and is depicted (Y-axis) versus a number of rounds (X-axis) for COMPAS (left panel) and Polish (right panel). In each subfigure, each curve tracks a competing set of linear, neural, or mixed models. The size of the competing set and the budget are mentioned in the caption. We observe that the fraction of recourses with a duality gap upper bound at or below the threshold is generally higher for linear competing sets by the end of 1000 rounds, even though other competing sets can initially achieve a higher success fraction. Across 54 dataset/competing set/budget configurations, when increasing the number of models from 6 to 24, the fraction of instances achieving the target gap has decreased in 22 comparisons, remained unchanged in 28 comparisons, and only increased in 4. The effect of budget is less uniform; increasing the budget increased the fraction in 23 comparisons, left it unchanged in 22, and decreased it in 9 comparisons. See Figure~\ref{fig:conv-app} in Appendix~\ref{sec:app-exp-duality} for other datasets.

To understand whether more rounds are required to improve the validity of ORPM, we compared the validity of recourse returned by ORPM at rounds 100 and 200 with the validity of recourse at round 1000. Outside GMC, the validity is identical from round 200 to round 1000 for 132 out of 135 dataset/model class/model count/budget configurations, and the increase in the other 3 configurations is around 4\%. In 100 rounds, in 129 of the 135 configurations, validity is the same as in 1000 rounds. In GMC, however, the validity gains are significantly higher when scaling from 100 or 200 rounds to 1000 rounds. Based on these observations, we used a default choice of 200 for all datasets except GMC and 1000 for GMC for the remainder of the experiments on linear, neural network, and mixed models.

\begin{figure}[htb!]
\centering
\begin{subfigure}[t]{0.49\textwidth}
\centering\includegraphics[width=\linewidth]{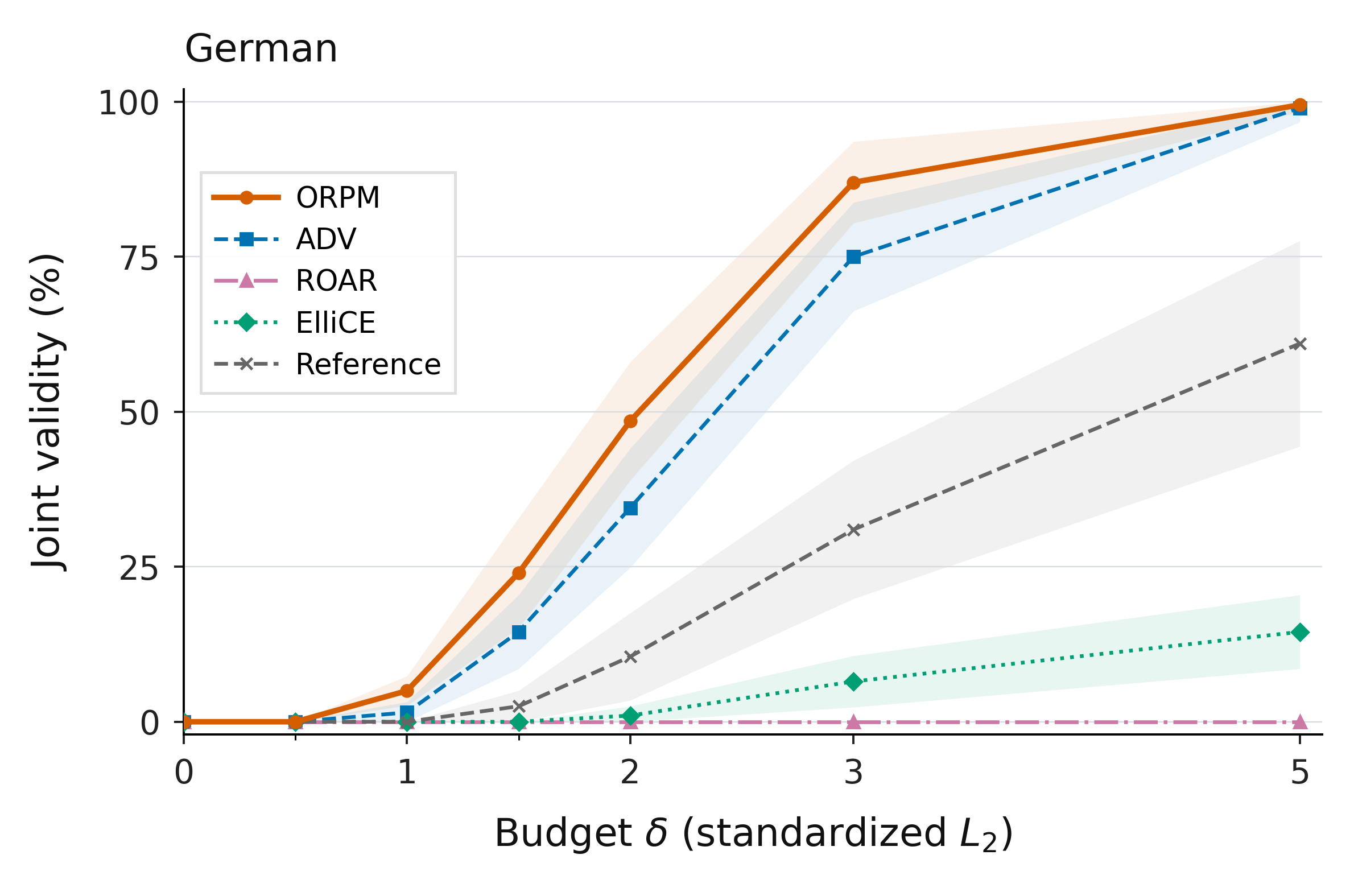}
\caption{German dataset, mixed competing set of 12 models. \label{fig:trade-off-main-german}}
\end{subfigure}\hfill
\begin{subfigure}[t]{0.49\textwidth}
\centering\includegraphics[width=\linewidth]{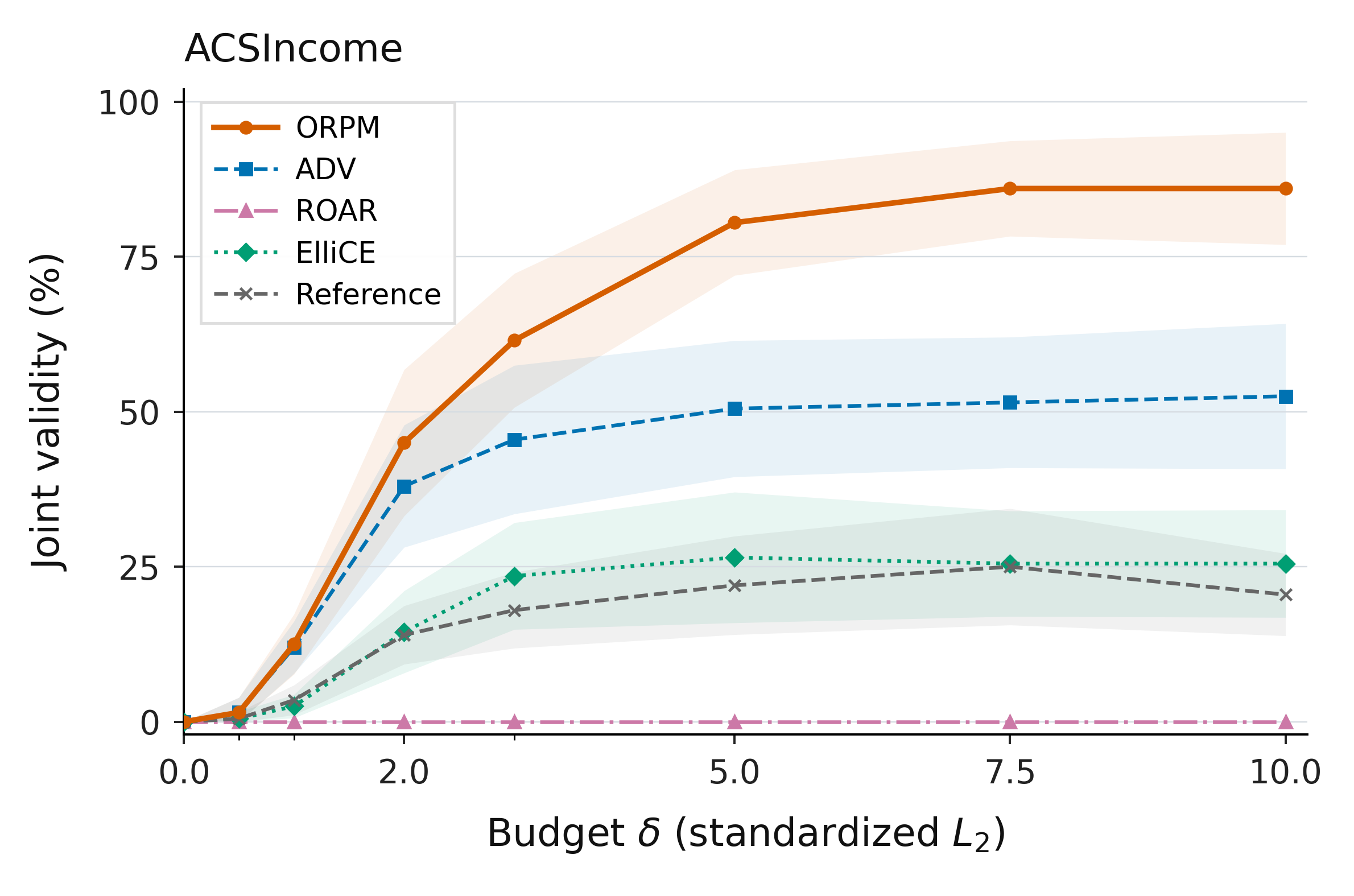}
\caption{ACSIncome dataset, neural competing set of 24 models. \label{fig:trade-off-main-folk}}
\end{subfigure}
\caption{The trade-off between validity and budget. Each subfigure corresponds to a dataset and competing set as described in the caption. In each subfigure, the curves track the validity of ORPM and baselines (ROAR~\cite{UpadhyayJL21}, ElliCE~\cite{TurbalVS25}, ADV, and Reference).
\label{fig:trade-off-main}}
\end{figure}

\ifarxiv\paragraph{Tradeoff Between Validity and Budget.}
\else
\noindent\textbf{Tradeoff Between Validity and Budget.}
\fi
\label{sec:exp-cost-validity-trade-off}
We next study the validity of recourse returned by ORPM, where we define a recourse as \emph{jointly valid} (or simply valid) if all the models provide a favorable prediction. In particular, we use the same combinations of models, model counts, and datasets as in the convergence experiments, but this time we use 10 seeds and compute recourse for 20 instances per seed, and all results are averaged over instances and seeds. We expand the set of budgets used compared to the convergence experiments (see Appendix~\ref{sec:app-exp-cost-validity-tradeoff}).

Figure~\ref{fig:trade-off-main} depicts the trade-off between the budget (X-axis) and the validity of recourse (Y-axis).  Each subfigure corresponds to a combination of dataset, competing set, and the number of competing models as indicated in the caption. In addition to ORPM, we measure the validity for baselines ROAR~\cite{UpadhyayJL21}, ElliCE~\cite{TurbalVS25}, and ADV. We also add another baseline called \emph{Reference}, where we simply optimize for recourse for the reference model in the competing set without taking into account whether such recourse would be valid for other models in the competing set. Results for other datasets are in Figure~\ref{fig:trade-off-app} in Appendix~\ref{sec:app-exp-cost-validity-tradeoff}.

We observe that in all datasets except GMC, ORPM achieves the highest validity over the baseline across all budgets. For GMC, other approaches outperform ORPM for mixed and neural network models, but ORPM outperforms for linear models (see Figure~\ref{fig:trade-off-app}). ADV generally has the second-highest validity after ORPM, while ElliCE performs better on linear than on nonlinear sets. ROAR and ElliCE address different uncertainty sets, and our results assess their stated adaptations on our finite competing sets.

\begin{figure}[htb!]
\centering
\begin{subfigure}[t]{0.49\textwidth}
\centering\includegraphics[width=\linewidth]{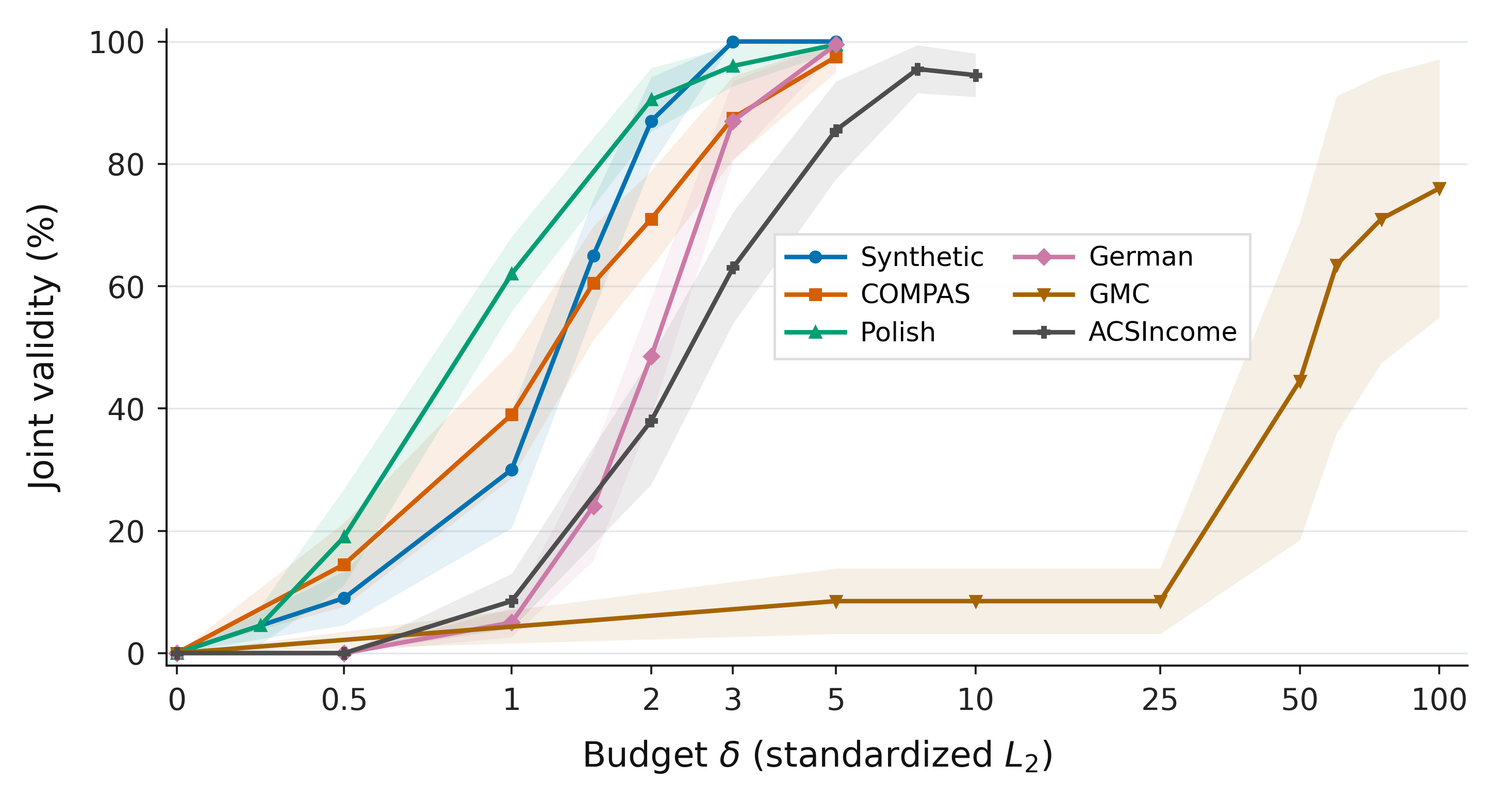}
\caption{The effect of budget on ORPM's validity for a mixed competing set of 12 models.\label{fig:parameters-main-budget}}
\end{subfigure}\hfill
\begin{subfigure}[t]{0.49\textwidth}
\centering\includegraphics[width=\linewidth]{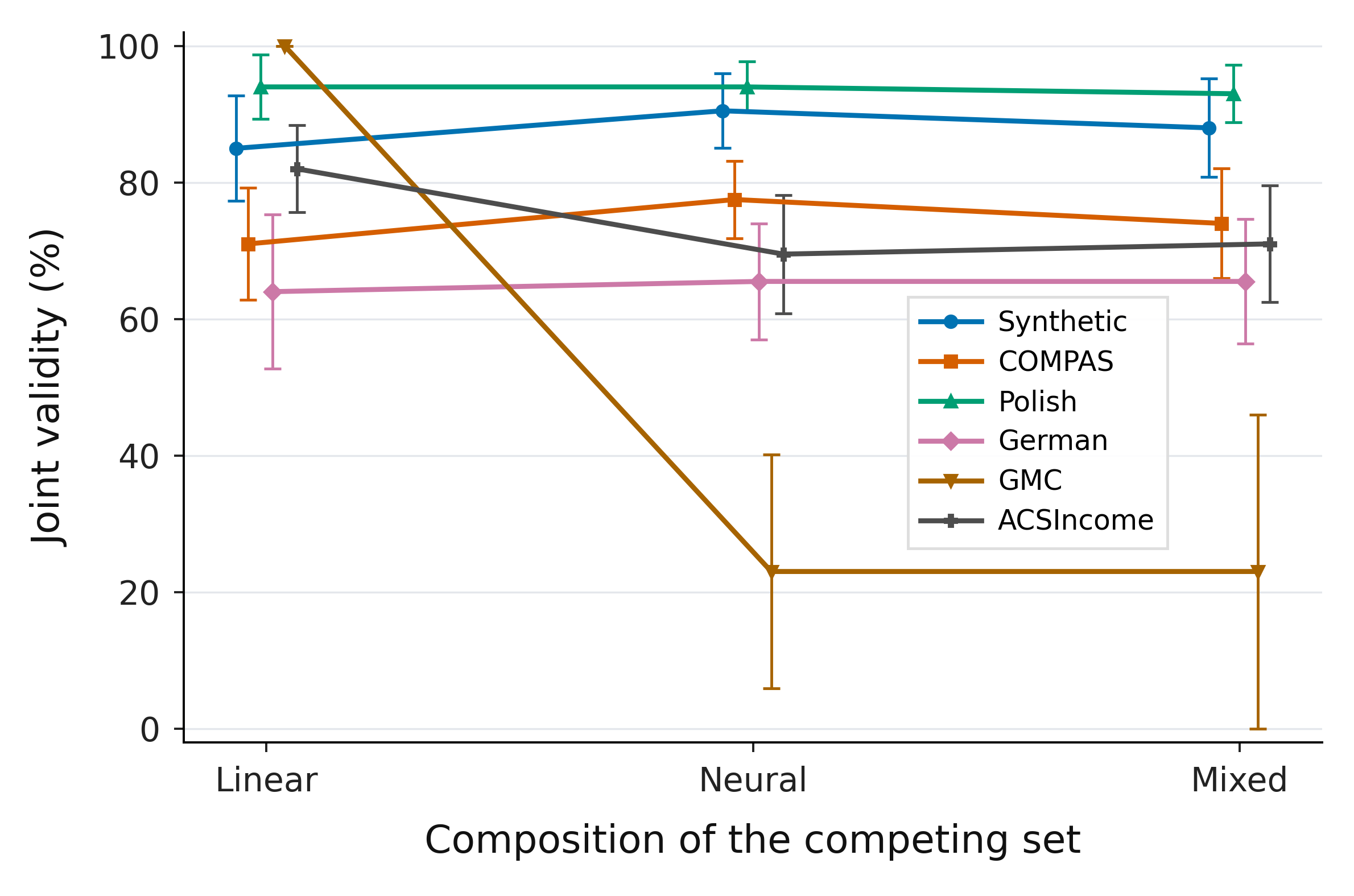}
\caption{The effect of composition of the competing set of 6 models on ORPM's validity.\label{fig:parameters-main-model-class}}
\end{subfigure}
\caption{The effect of budget and composition of the competing set on validity. Each curve corresponds to a different dataset.\label{fig:parameters-main}}
\end{figure}

Increasing the budget will generally result in ORPM achieving perfect 
(or near-perfect) validity in several settings. An example of this behavior is provided in 
Figure~\ref{fig:parameters-main-budget}. In this figure, for a mixed competing set 
of 12 models, we track the validity of ORPM across each dataset as the budget increases. 
However, residual failures remain in some configurations, including ACSIncome with 24 neural models 
(see e.g., Figure~\ref{fig:trade-off-main-folk}).
For many settings, we observe 3 regimes: a low budget regime where the validity 
remains low, a medium budget regime where the validity sharply increases, and a high budget 
regime where the validity is near perfect and the rate of validity improvement is significantly 
lower compared to the medium budget regime. 
See Figure~\ref{fig:parameters-app-budget} 
in Appendix~\ref{sec:app-exp-cost-validity-tradeoff} for more details.

The validity generally decreases when the competing set contains both linear and neural models
compared to the case where it contains only linear or neural models. In general, achieving
high validity for neural/mixed models is more difficult than for linear models. Figure~\ref{fig:parameters-main-model-class}
tracks the validity of recourse returned by ORPM (Y-axis) for different compositions of competing sets containing 6 classifiers (X-axis).
We also observe that validity drops most significantly on GMC, but the drop is generally milder on other datasets.
These observations are fairly robust across competing sets of different sizes. Observed validity generally decreases with competing-set size, although notable exceptions occur, including GMC. Moreover, the validity degradation is more pronounced for neural and mixed competing sets, compared to linear competing sets.
See Figure~\ref{fig:parameters-app-size} 
in Appendix~\ref{sec:app-exp-cost-validity-tradeoff} for more details.

\ifarxiv\paragraph{Homogeneous Competing Sets.}
\else
\noindent\textbf{Homogeneous Competing Sets and Generalization to Unseen Models.}
\fi
We next study the validity of the provided recourse when the competing set contains homogeneous models and compare with prior approaches that are specifically designed to handle such competing sets. In particular, we focus on two types of competing sets: one only containing linear models and another consisting of neural networks with the same architecture but different weights. 

\begin{table}[htb!]
\centering
\begin{subtable}[t]{0.49\textwidth}
\centering
\captionsetup{font=footnotesize,skip=4pt}
\footnotesize
\setlength{\tabcolsep}{1.5pt}
\renewcommand{\arraystretch}{1.0}
\begin{tabular}{@{}lrlrr@{}}
\toprule
Dataset & $\delta$ & Set & ORPM & ElliCE \\
\midrule
Synthetic & 2 & Seen & $\mathbf{79.0} \pm 8.4$ & $69.5 \pm 7.2$ \\
 &  & Unseen & $\mathbf{78.5} \pm 8.3$ & $70.0 \pm 7.5$ \\
 & 5 & Seen & $\mathbf{100.0} \pm 0.0$ & $\mathbf{100.0} \pm 0.0$ \\
 &  & Unseen & $\mathbf{100.0} \pm 0.0$ & $\mathbf{100.0} \pm 0.0$ \\
\midrule
COMPAS & 2 & Seen & $\mathbf{59.5} \pm 13.9$ & $37.5 \pm 15.8$ \\
 &  & Unseen & $30.5 \pm 14.6$ & $\mathbf{36.0} \pm 13.6$ \\
 & 5 & Seen & $\mathbf{94.0} \pm 4.4$ & $76.5 \pm 22.6$ \\
 &  & Unseen & $72.5 \pm 26.8$ & $\mathbf{89.5} \pm 7.2$ \\
\midrule
Polish & 2 & Seen & $\mathbf{91.5} \pm 4.5$ & $33.5 \pm 7.2$ \\
 &  & Unseen & $\mathbf{88.0} \pm 4.5$ & $39.5 \pm 8.3$ \\
 & 5 & Seen & $\mathbf{97.0} \pm 2.5$ & $73.5 \pm 8.4$ \\
 &  & Unseen & $\mathbf{96.0} \pm 2.8$ & $74.0 \pm 7.7$ \\
\midrule
German & 2 & Seen & $\mathbf{57.0} \pm 6.8$ & $5.0 \pm 2.4$ \\
 &  & Unseen & $\mathbf{35.5} \pm 10.6$ & $10.0 \pm 3.4$ \\
 & 5 & Seen & $\mathbf{100.0} \pm 0.0$ & $25.5 \pm 6.4$ \\
 &  & Unseen & $\mathbf{99.0} \pm 1.5$ & $27.0 \pm 6.4$ \\
\midrule
GMC & 50 & Seen & $\mathbf{100.0} \pm 0.0$ & $\mathbf{100.0} \pm 0.0$ \\
 &  & Unseen & $0.0 \pm 0.0$ & $\mathbf{99.0} \pm 1.5$ \\
 & 100 & Seen & $\mathbf{100.0} \pm 0.0$ & $\mathbf{100.0} \pm 0.0$ \\
 &  & Unseen & $0.0 \pm 0.0$ & $\mathbf{99.0} \pm 1.5$ \\
\midrule
ACSIncome & 3 & Seen & $\mathbf{74.0} \pm 8.2$ & $60.0 \pm 8.1$ \\
 &  & Unseen & $\mathbf{65.0} \pm 7.9$ & $58.5 \pm 7.9$ \\
 & 10 & Seen & $\mathbf{100.0} \pm 0.0$ & $\mathbf{100.0} \pm 0.0$ \\
 &  & Unseen & $\mathbf{100.0} \pm 0.0$ & $\mathbf{100.0} \pm 0.0$ \\
\bottomrule
\end{tabular}
\caption{Linear: ORPM vs. ElliCE.\label{tab:linear-orpm-ellice}}
\end{subtable}
\hfill
\begin{subtable}[t]{0.49\textwidth}
\centering
\captionsetup{font=footnotesize,skip=4pt}
\footnotesize
\setlength{\tabcolsep}{1.5pt}
\renewcommand{\arraystretch}{1.0}
\begin{tabular}{@{}lrlrrr@{}}
\toprule
Dataset & $\delta$ & Set & ORPM & MaxMargin & MinL2 \\
\midrule
Synthetic & 1 & Seen & $\mathbf{54.0} \pm 19.1$ & $\mathbf{54.0} \pm 19.1$ & $\mathbf{54.0} \pm 19.1$ \\
 &  & Unseen & $\mathbf{52.0} \pm 18.1$ & $\mathbf{52.0} \pm 18.1$ & $0.0 \pm 0.0$ \\
 & 3 & Seen & $\mathbf{100.0} \pm 0.0$ & $\mathbf{100.0} \pm 0.0$ & $\mathbf{100.0} \pm 0.0$ \\
 &  & Unseen & $\mathbf{100.0} \pm 0.0$ & $\mathbf{100.0} \pm 0.0$ & $0.0 \pm 0.0$ \\
\midrule
COMPAS & 1 & Seen & $\mathbf{42.0} \pm 12.5$ & $\mathbf{42.0} \pm 12.5$ & $\mathbf{42.0} \pm 12.5$ \\
 &  & Unseen & $\mathbf{36.0} \pm 13.1$ & $\mathbf{36.0} \pm 13.1$ & $0.0 \pm 0.0$ \\
 & 3 & Seen & $\mathbf{92.0} \pm 7.4$ & $\mathbf{92.0} \pm 7.4$ & $\mathbf{92.0} \pm 7.4$ \\
 &  & Unseen & $\mathbf{90.0} \pm 7.5$ & $\mathbf{90.0} \pm 7.5$ & $0.0 \pm 0.0$ \\
\midrule
Polish & 1 & Seen & $\mathbf{78.0} \pm 10.6$ & $60.0 \pm 6.7$ & $\mathbf{78.0} \pm 10.6$ \\
 &  & Unseen & $\mathbf{34.0} \pm 11.8$ & $26.0 \pm 11.8$ & $0.0 \pm 0.0$ \\
 & 3 & Seen & $\mathbf{94.0} \pm 6.9$ & $88.0 \pm 10.0$ & $\mathbf{94.0} \pm 6.9$ \\
 &  & Unseen & $\mathbf{88.0} \pm 7.4$ & $78.0 \pm 12.5$ & $0.0 \pm 0.0$ \\
\midrule
German & 1 & Seen & $\mathbf{20.0} \pm 9.5$ & $12.0 \pm 7.4$ & $\mathbf{20.0} \pm 9.5$ \\
 &  & Unseen & $\mathbf{0.0} \pm 0.0$ & $\mathbf{0.0} \pm 0.0$ & $\mathbf{0.0} \pm 0.0$ \\
 & 3 & Seen & $\mathbf{94.0} \pm 6.9$ & $76.0 \pm 14.8$ & $\mathbf{94.0} \pm 6.9$ \\
 &  & Unseen & $\mathbf{20.0} \pm 11.7$ & $14.0 \pm 11.8$ & $0.0 \pm 0.0$ \\
\midrule
GMC & 25 & Seen & $\mathbf{90.0} \pm 10.1$ & $\mathbf{90.0} \pm 10.1$ & $60.0 \pm 17.8$ \\
 &  & Unseen & $\mathbf{90.0} \pm 10.1$ & $82.0 \pm 18.4$ & $0.0 \pm 0.0$ \\
 & 75 & Seen & $92.0 \pm 10.0$ & $\mathbf{96.0} \pm 6.0$ & $66.0 \pm 19.1$ \\
 &  & Unseen & $\mathbf{92.0} \pm 10.0$ & $88.0 \pm 15.4$ & $0.0 \pm 0.0$ \\
\midrule
ACSIncome & 2 & Seen & $\mathbf{58.0} \pm 19.6$ & $\mathbf{58.0} \pm 19.6$ & $\mathbf{58.0} \pm 19.6$ \\
 &  & Unseen & $\mathbf{52.0} \pm 19.3$ & $\mathbf{52.0} \pm 19.3$ & $0.0 \pm 0.0$ \\
 & 7.5 & Seen & $\mathbf{100.0} \pm 0.0$ & $\mathbf{100.0} \pm 0.0$ & $\mathbf{100.0} \pm 0.0$ \\
 &  & Unseen & $\mathbf{100.0} \pm 0.0$ & $\mathbf{100.0} \pm 0.0$ & $2.0 \pm 4.5$ \\
\bottomrule
\end{tabular}
\caption{Neural: ORPM vs MIQCP approaches.\label{tab:neural-orpm-exact}}
\end{subtable}
\caption{Joint validity (\%) on seen and unseen models. The left panel compares ORPM and ElliCE on 24 seen linear models and 50 unseen models. The right panel compares ORPM with the MaxMargin and MinL2 MIQCP references on 12 seen networks with the same architecture and 50 unseen models. Entries are means $\pm$ pointwise 95\% Student-$t$ intervals. \label{tab:linear-comparisons}}
\end{table}

For linear competing sets, we created a competing set of 24 models and compared validity with ElliCE~\cite{TurbalVS25}, which is designed to provide robust recourse over an ellipsoidal approximation of the Rashomon set. Table~\ref{tab:linear-orpm-ellice} summarizes the results (the rows labeled as seen in Table~\ref{tab:linear-orpm-ellice}), where for each dataset and budget value, we report the joint validity of ORPM and ElliCE. We observe that ORPM achieves higher or equal validity on all seen models. However, evaluating ElliCE strictly on the finite subset may penalize its intended design, which aims to protect against any model in the ellipsoidal approximation of the Rashomon set. To highlight the tradeoff between true generalization and finite-set optimization, we drew 50 additional unseen models from the filtered ellipsoidal approximation of the Rashomon set and measured the validity of recourse provided by both ORPM and ElliCE on these unseen models. Results are summarized in Table~\ref{tab:linear-orpm-ellice} (the rows labeled as unseen in Table~\ref{tab:linear-orpm-ellice}). We see that ORPM's validity generally drops on unseen models, while ElliCE is more robust. Still, ORPM achieves higher or equal unseen joint validity on four datasets, while ElliCE performs better on COMPAS and GMC. Thus, ORPM shows broader applicability than its intended design. 

For the same-architecture neural experiment, we used 12 networks with one hidden layer of 20 ReLU units. We used 10 seeds and 5 instances per seed.
We compared ORPM with two joint neural-optimization baselines.
MinL2 adapts the minimum-distance formulation of~\citep{LeofanteBR23}, using $L_2$
rather than $L_1$ distance and requiring acceptance by every
competing network. MaxMargin 
maximizes the minimum network logit within budget $\delta$.
Both jointly encode the ReLU networks, and the quadratic distance
terms yield MIQCP formulations.
MaxMargin receives one 60-second solve per positive budget; MinL2 receives one 60-second solve at the largest budget, 
and its result is reused for smaller budgets. Appendix~\ref{sec:app-exp-details} specifies the formulations, numerical tolerances, and return rules. Table~\ref{tab:neural-orpm-exact} reports joint validity for ORPM and these two references.
On seen models (the rows labeled as seen in Table~\ref{tab:neural-orpm-exact}), ORPM matches
MaxMargin on Synthetic, COMPAS and ACSIncome at both
displayed budgets, and has higher means on Polish and
German. GMC at budget 75 is an exception: MaxMargin
achieves 96\% joint validity versus ORPM's 92\%. Similar to the linear case, to examine 
robustness, we drew 50 additional unseen models by using different initializations and feature masks. 
ORPM matches or exceeds
MaxMargin's mean unseen joint validity at every displayed
setting (the rows labeled as unseen in Table~\ref{tab:neural-orpm-exact}). MinL2 often matches ORPM on seen models but
achieves only 0--2\% unseen validity in these
settings. Thus, joint feasibility on the competing set
does not ensure transfer to unseen models.

\ifarxiv\paragraph{The Cost of Multiplicity.}
\else
\noindent\textbf{The Cost of Multiplicity.}
\fi
We next study how predictive multiplicity affects the difficulty of providing valid recourse by measuring budget-feasibility curves and the loss of recourse validity when separate model-specific actions are replaced by one common action. In particular, for each dataset, budget, and competing model set, we compared ORPM’s joint success rate with three feasibility benchmarks: (1) \emph{average singleton}, the fraction of models for which a budget-feasible successful action exists; (2) \emph{all separately}, the fraction of instances for which every model admits a budget-feasible successful action, allowing different actions for different models; and (3) \emph{one common action}, the fraction of instances for which a single budget-feasible action is accepted by every model simultaneously. These benchmarks impose increasingly stringent requirements, so their success rates weakly decrease in that order. We focused on homogeneous linear and neural competing sets, where we can better analyze the true success of one common action using convex quadratic optimization and MIQCP solvers. When numerical optimization could not resolve benchmark feasibility, we retained lower and upper bounds.

\begin{figure}[htbp]
\centering
\begin{subfigure}[t]{0.49\textwidth}
\centering\includegraphics[width=\linewidth]{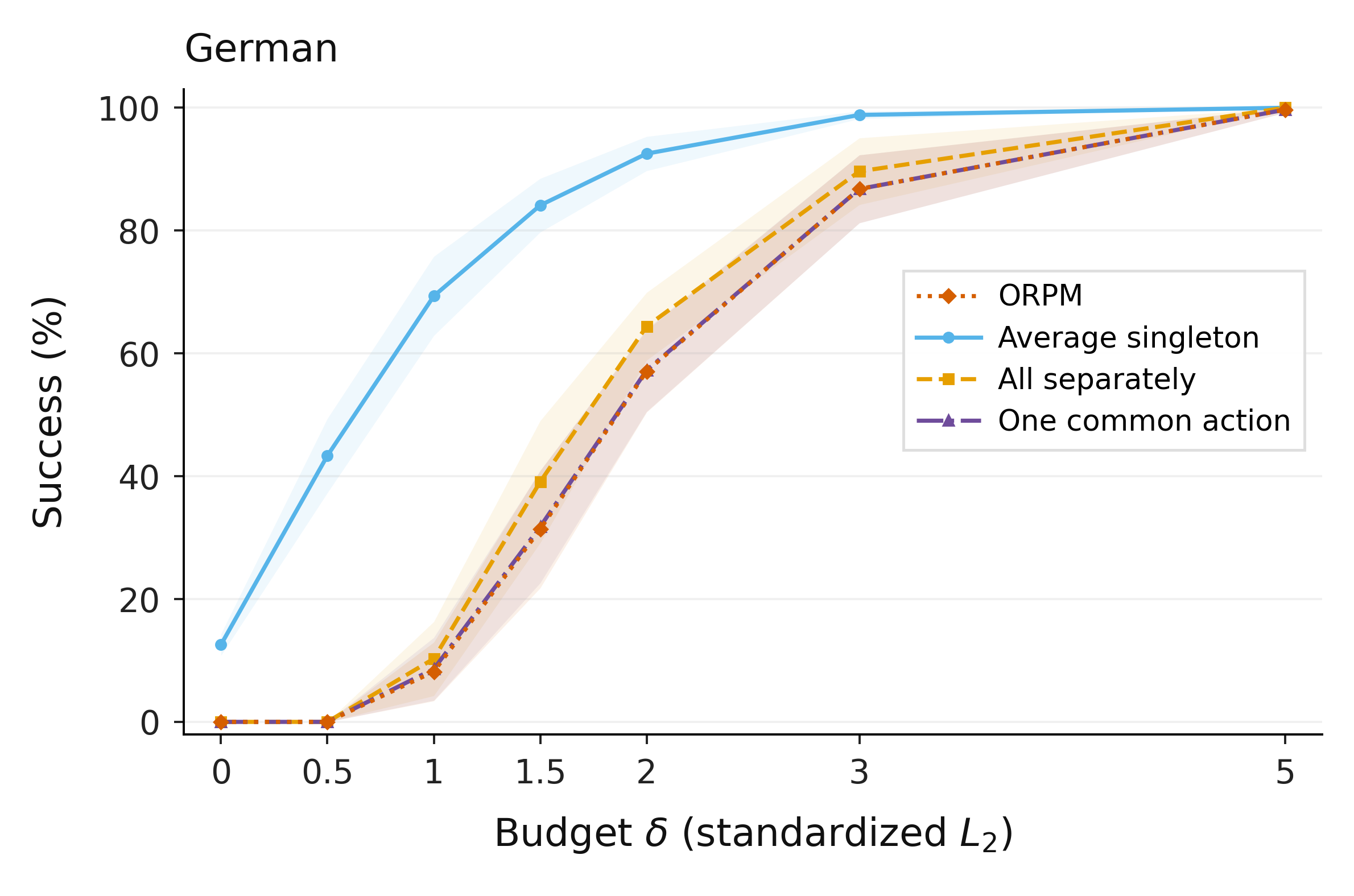}
\caption{German dataset, linear competing set of 24 models.}
\end{subfigure}\hfill
\begin{subfigure}[t]{0.49\textwidth}
\centering\includegraphics[width=\linewidth]{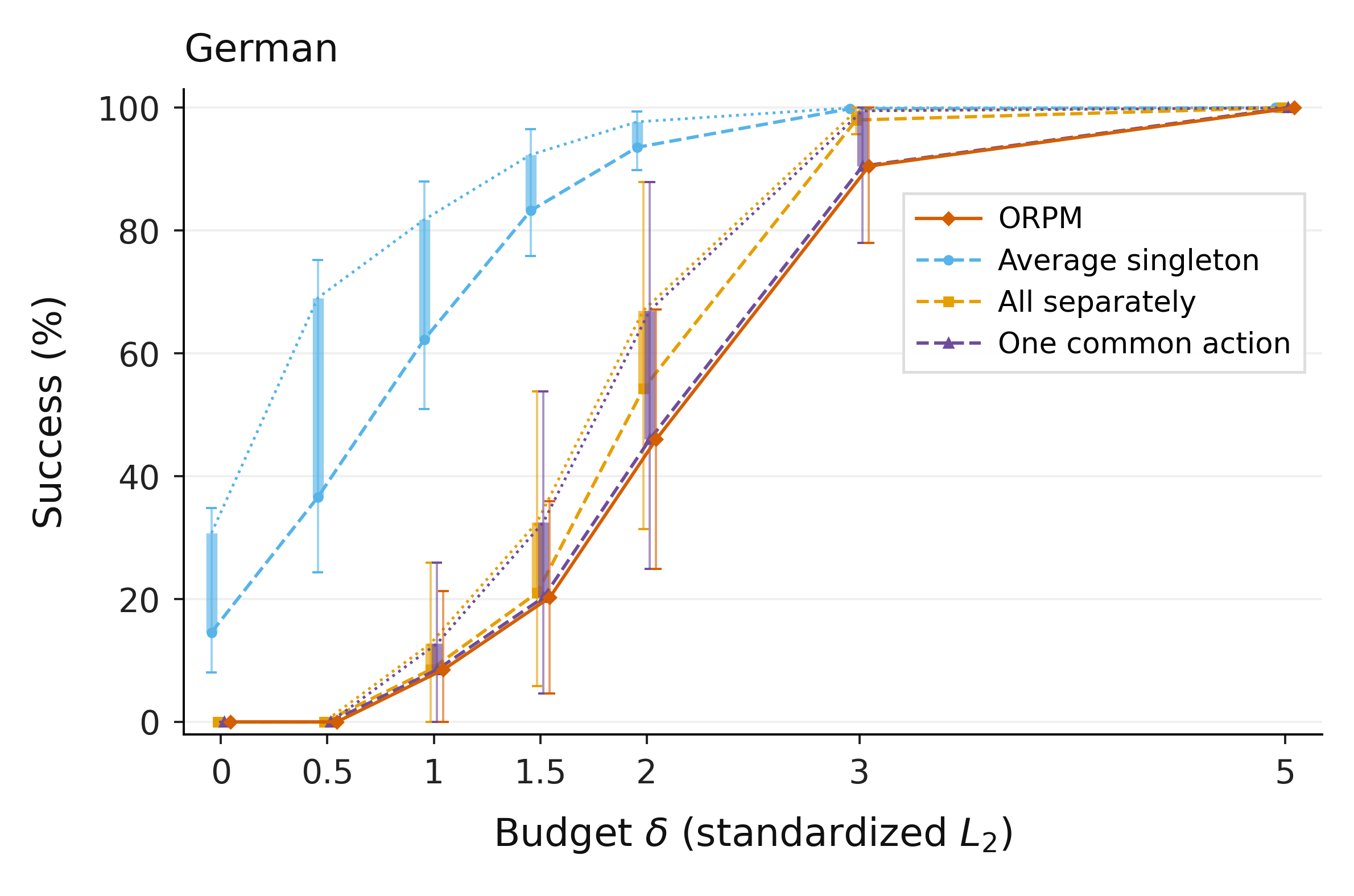}
\caption{German dataset, neural competing set of 24 models.}
\end{subfigure}
\caption{The validity degradation to guard against multiplicity. Each subfigure corresponds to a pair of datasets and competing sets as indicated in the caption. Each curve tracks the success rate (Y-axis) for ORPM and baselines as a function of allowable budget (X-axis).  \label{fig:cost-main}}
\end{figure}

For a fixed combination of dataset and competing set, we computed the success rate of ORPM and the baselines as a function of allowable budget. The results for the German dataset on competing sets of 24 models are summarized in Figure~\ref{fig:cost-main}. For linear models (left panel) at budget 2, average singleton success is $92.5\%$, whereas all-separate and common-action success are $64.4\%$ and $57.2\%$, respectively. The first gap reflects the difficulty of accommodating each model individually; the additional $ 7.2$-percentage-point drop reflects the added requirement that one action satisfy them simultaneously. ORPM achieves $57.1\%$, closely matching common-action feasibility and indicating little observed algorithmic shortfall at this budget. For neural models (right panel) at the same budget of 2, singleton success is bounded by $[93.6,97.7]\%$, while common-action success is bounded by $[46.0,67.0]\%$ and ORPM achieves a success rate of $46.0\%$, so the algorithmic shortfall is bounded by 0--21 percentage points. The unresolved range does not establish that ORPM attains the common-action optimum. Increasing the budget to 5 raises ORPM success to $99.7\%$ for linear models and $100\%$ for neural models, showing that additional budget largely overcomes the observed burden in the German dataset.  The appendix shows similar separations between singleton and joint feasibility across the remaining datasets, although their magnitudes vary. See Figures~\ref{fig:cost-app-1}~and~\ref{fig:cost-app-2} in Appendix~\ref{sec:app-exp-cost-of-multiplicity} for more details.

\ifarxiv\paragraph{The Disparity of Multiplicity.}
\else
\noindent\textbf{The Disparity of Multiplicity.}
\fi
We next examine demographic disparities in recourse success
under predictive multiplicity. We focus on four datasets:
COMPAS and ACSIncome, using the recorded female and male
groups, and GMC and German, using individuals younger than
40 and those aged 40 or older. Synthetic and Polish are
excluded because we do not designate a sensitive attribute
for these datasets in our analysis. We considered competing
sets of 24 linear models and compared ORPM with the three
cost-of-multiplicity benchmarks: average singleton, all
separately, and one common action. For each method or
benchmark, we computed the signed difference in success
rates between the two groups within each seed, average
these differences across ten seeds, and take the absolute
value. This measures the magnitude of the average group
success disparity, in percentage points.

\begin{figure}[htbp]
\centering
\begin{subfigure}[t]{0.49\textwidth}
\centering\includegraphics[width=\linewidth]{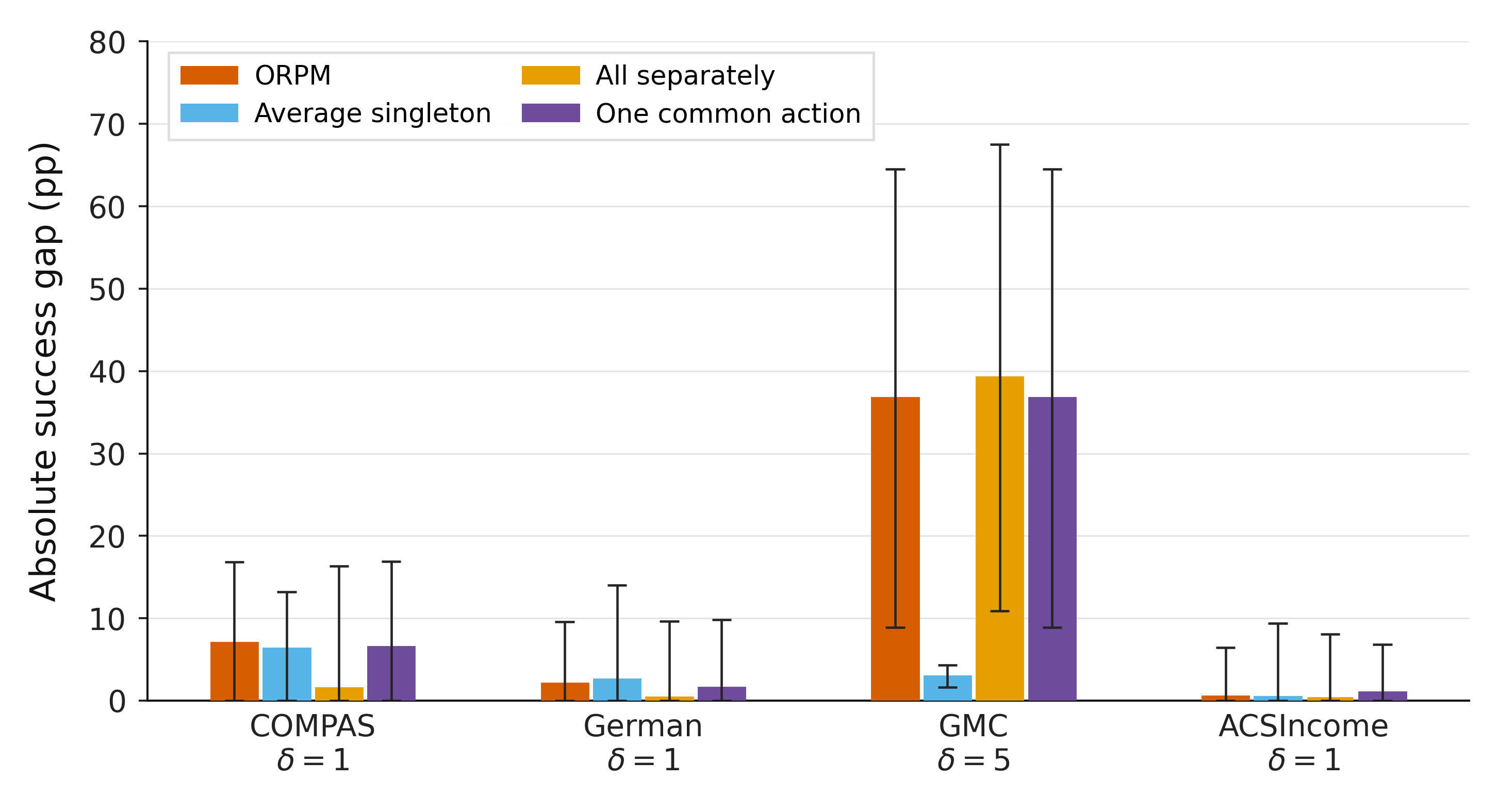}
\caption{Linear competing set, small budget regime.}
\end{subfigure}\hfill
\begin{subfigure}[t]{0.49\textwidth}
\centering\includegraphics[width=\linewidth]{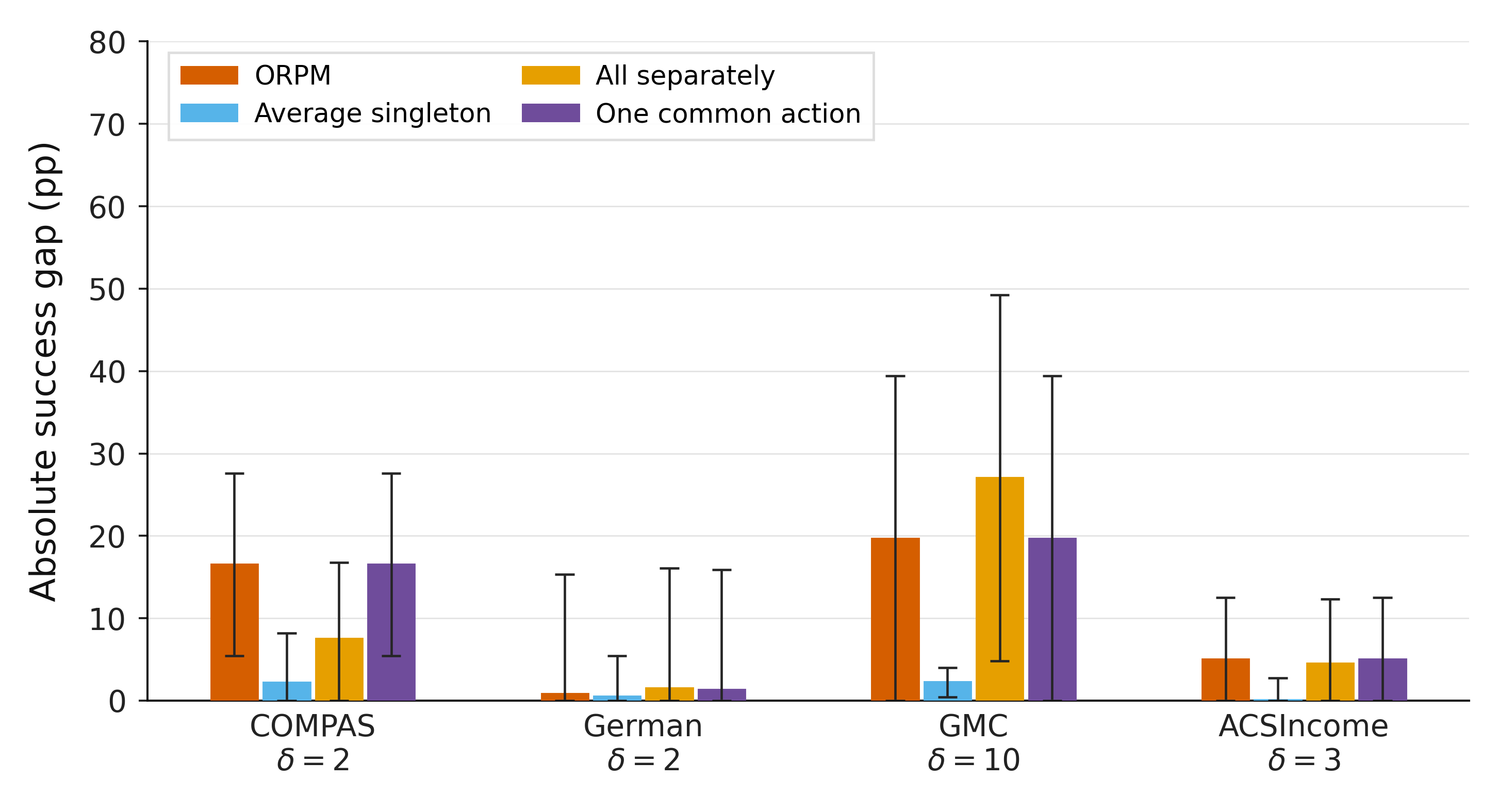}
\caption{Linear competing set, medium budget regime.}
\end{subfigure}
\caption{The disparity of recourse success across different groups. Each subfigure corresponds to different budget values, with small budgets on the left and medium budgets on the right. In each subfigure, the bars track the disparity in recourse success (Y-axis) for ORPM and baselines on each dataset.\label{fig:disparity-main}}
\end{figure}

Figure~\ref{fig:disparity-main} shows that ORPM closely
tracks the common-action benchmark: their estimated disparity gap
magnitudes differ by at most 0.5 percentage points across
the displayed linear settings. ORPM's disparity gaps are generally
larger than those of average singleton, demonstrating that multiplicity 
can further amplify the disparities. On the other hand,
all-separately can show larger or smaller gaps than ORPM.
These comparisons concern benchmarks with different
success requirements and do not imply an ordering of
disparity by stringency. Increasing the budget does not
uniformly reduce disparity: ORPM's disparity gap on GMC decreases from
36.7 to 19.6 percentage points as the budget increases
from 5 to 10, whereas its COMPAS gap increases from 7.0
to 16.5 points as the budget increases from 1 to 2.
Thus, additional budget can narrow or widen observed
group differences. In these experiments, confidence intervals are often wide,
limiting conclusions about method rankings. 
See Figure~\ref{fig:disparity-app} in
Appendix~\ref{sec:app-exp-disparity-of-multiplicity} for neural competing sets.

\begin{figure}[htbp]
\centering
\begin{subfigure}[t]{0.49\textwidth}
\centering\includegraphics[width=\linewidth]{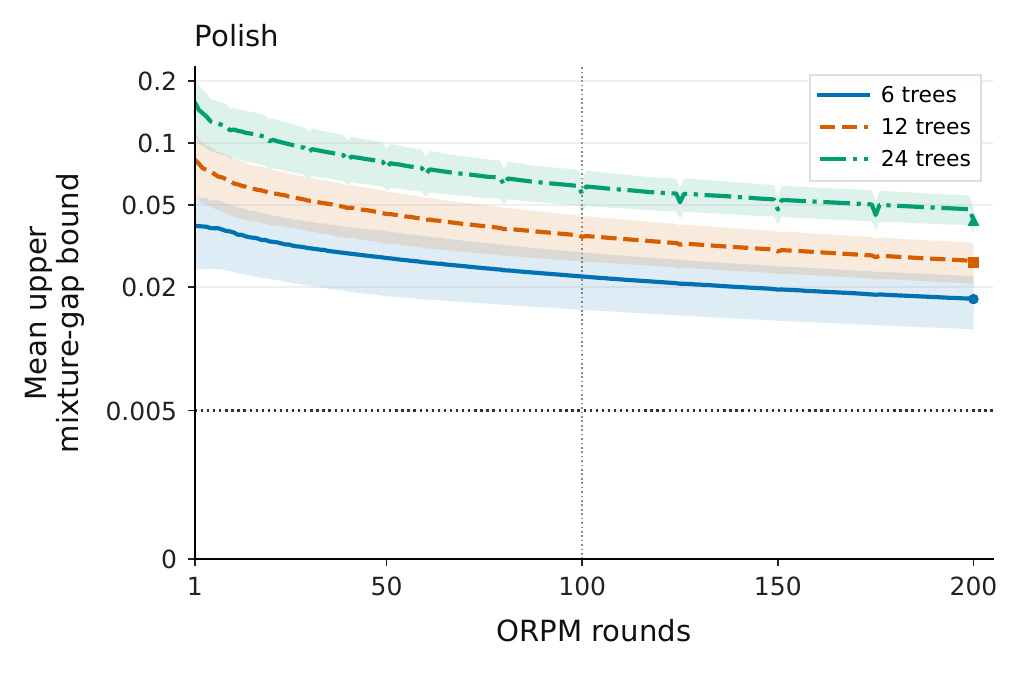}
\caption{Convergence for Polish on tree-based competing sets.\label{fig:tree-convergence-polish}}
\end{subfigure}\hfill
\begin{subfigure}[t]{0.49\textwidth}
\centering\includegraphics[width=\linewidth]{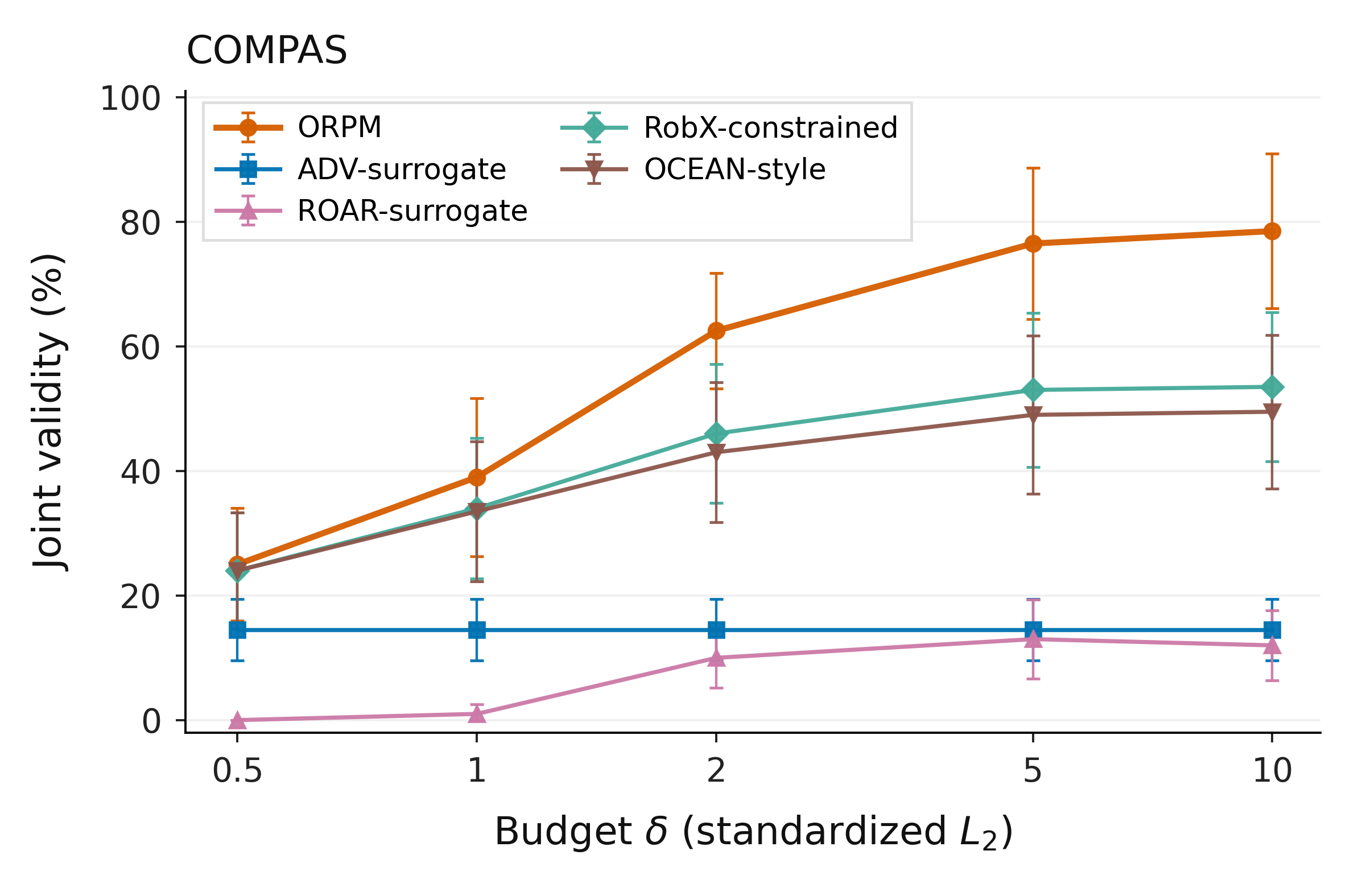}
\caption{Joint validity for COMPAS on tree-based competing sets.\label{fig:tree-trade-off-compas}}
\end{subfigure}
\caption{Experiments on tree-based competing sets: The left panel depicts the average upper bound on the mixture gap (Y-axis) as the number of rounds of ORPM (X-axis). The right panel tracks the joint validity (Y-axis) of ORPM and tree-based baselines as a function of recourse budget (X-axis).\label{fig:trees-result-1}}
\end{figure}

\ifarxiv\paragraph{Tree-Based Competing Sets.}
\else
\noindent\textbf{Tree-Based Competing Sets.}
\fi
So far we have focused on convex and nonconvex but differentiable competing sets. As our
last experiments, we revisit our prior empirical evaluations by focusing on non-convex tree-based
competing sets. For all experiments in this section, the competing sets contain trees of depth four.

We start with convergence experiments: we ran ORPM on different datasets and competing sets of different sizes for 200 rounds and measured the validity and the mixture gap of the returned recourse. Consistent with differentiable experiments,
we see that the mixture gap decreases when ORPM is run for more rounds. An example of such behavior is shown in Figure~\ref{fig:tree-convergence-polish}, 
where we track mean upper bounds on the mixture gap for competing sets of different sizes on the Polish dataset
using a budget of 2. We observe that the upper bound on the mixture gap is lower for competing sets with a smaller number of trees. Similar to differentiable competing sets, joint validity can be achieved before the mixture gap reaches our target of 0.005. For example, for the competing set with 12 trees, we observe 88.5\% joint validity after 100 rounds, 
despite no instances attaining the certified mixture-gap target. Increasing the number of rounds from 100 to 200 will increase the joint validity despite the smaller
progress rate in decreasing the mixture gap.

We also examine how the available recourse budget affects joint validity across five competing decision trees on COMPAS, using 10 seeds and 20 instances per seed. Figure~\ref{fig:tree-trade-off-compas} summarizes the results, tracking the joint validity of ORPM and baselines (Y-axis) across different budget values (X-axis). ORPM achieves the highest observed joint-validity rate at every tested budget among the tree-based baselines, increasing from 25.0\% at $\delta=0.5$ to 78.5\% at $\delta=10$. At $\delta=2$, ORPM achieves 62.5\% joint validity, compared with 46.0\% for RobX-constrained, 43.0\% for OCEAN-style, 14.5\% for ADV-surrogate, and 10.0\% for ROAR-surrogate. ORPM's advantage over the strongest baseline, RobX-constrained, is 16.5 percentage points. Increasing the budget also improves RobX-constrained and OCEAN-style, but their validity reaches only 53.5\% and 49.5\%, respectively, at $\delta=10$. ADV-surrogate remains at 14.5\% across the tested budgets, while ROAR-surrogate reaches at most 13.0\%. Compared to the baselines, ORPM benefits more from additional budget, although its improvement diminishes at larger budgets: increasing $\delta$ from 5 to 10 raises validity only from 76.5\% to 78.5\%. 

\begin{figure}[htbp]
\centering
\begin{subfigure}[t]{0.45\textwidth}
\centering\includegraphics[width=\linewidth]{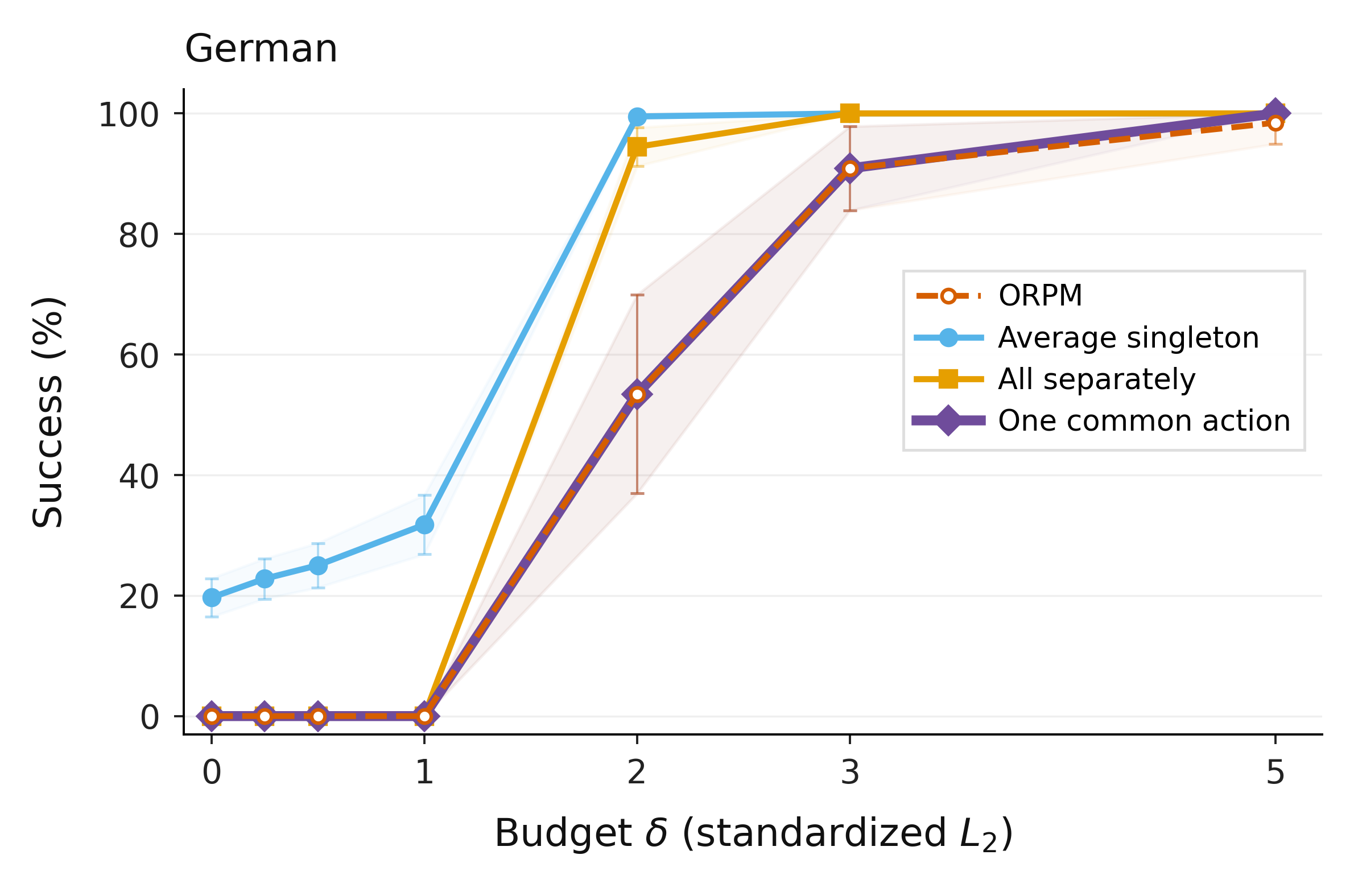}
\caption{Cost of multiplicity for the German dataset on tree-based competing sets.\label{fig:tree-cost-of-multiplicity-german}}
\end{subfigure}\hfill
\begin{subfigure}[t]{0.54\textwidth}
\centering\includegraphics[width=\linewidth]{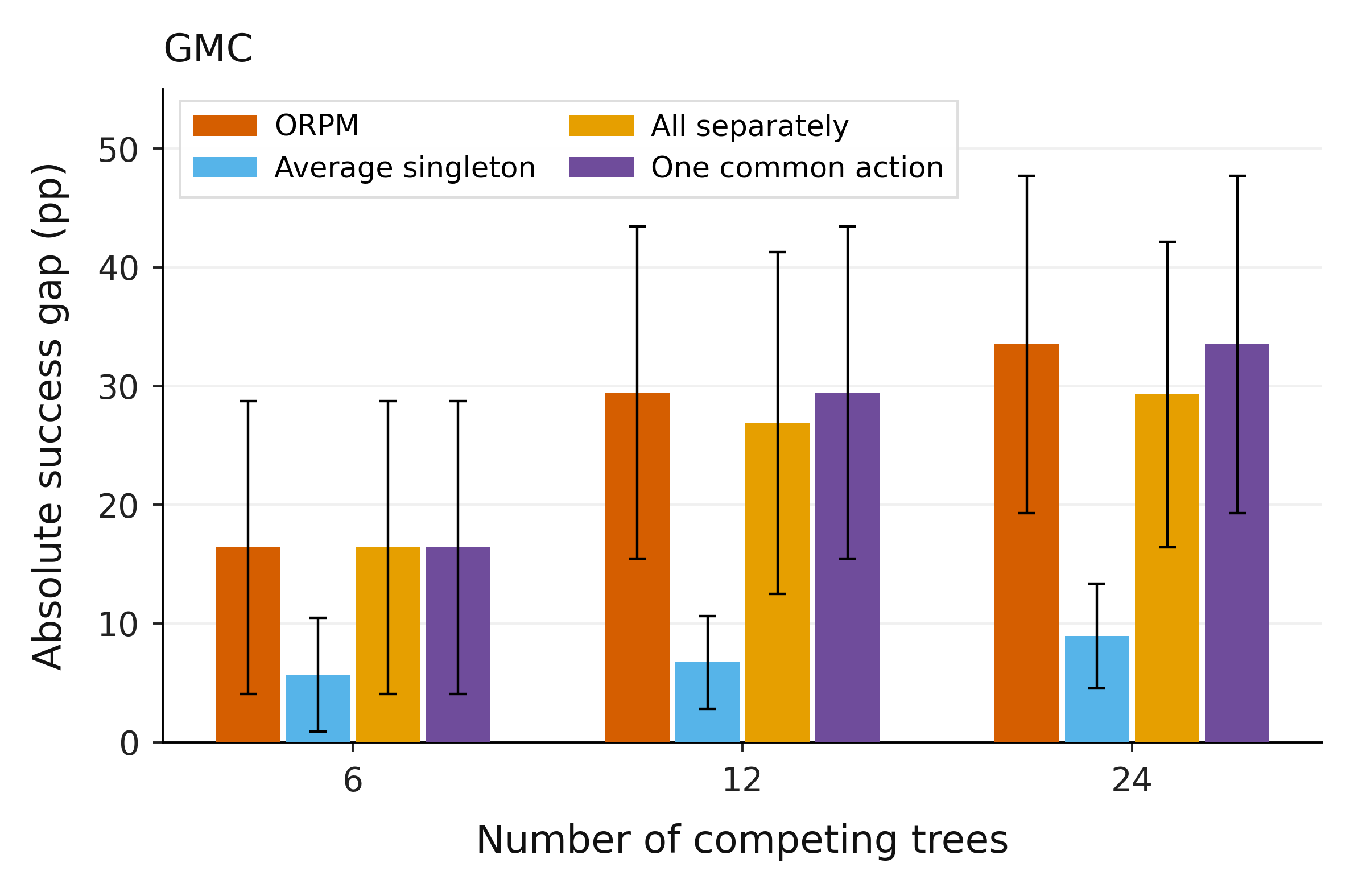}
\caption{Disparity of multiplicity for the GMC dataset on tree-based competing sets.\label{fig:tree-disparity-gmc}}
\end{subfigure}
\caption{Additional experiments on tree-based competing sets: The left panel depicts the joint validity (Y-axis) for each budget (X-axis) for ORPM and baselines that are affected by multiplicity to different degrees. In the right panel, the bars track the disparity in recourse success (Y-axis) for ORPM and baselines on the GMC dataset.\label{fig:trees-result-2}}
\end{figure}

We next examine how multiplicity affects the validity. In Figure~\ref{fig:tree-cost-of-multiplicity-german}, we track the joint validity of ORPM and baselines (Y-axis) against the recourse budget (X-axis) for the German dataset with 12 competing decision trees. These baselines include average singleton, all separately, and all common action, with the same description provided in the differentiable experiments. At $\delta=2$, average singleton success is 99.5\% and all-separate feasibility is 94.5\%, but common-action feasibility falls to 53.4\%, a reduction of 41.0 percentage points. ORPM attains the same 53.4\% joint-validity rate, indicating that the main difficulty at this budget is satisfying the competing trees with one action. Increasing the budget to $\delta=3$ raises both common-action feasibility and ORPM validity to 90.9\%. At $\delta=5$, a common action is feasible for every evaluated case, while ORPM achieves 98.4\% validity. Thus, additional budget substantially reduces the validity loss from requiring a shared action, and ORPM closely tracks common-action feasibility in this setting.

Finally, we examine whether the requirement to provide recourse across competing trees affects demographic groups differently. In Figure~\ref{fig:tree-disparity-gmc}, we depict the disparity in recourse success for ORPM and baselines on the GMC dataset for different sizes of competing sets. GMC gives a particularly clear example at budget $\delta=1$: the absolute success gap between the two subgroups (younger than 40 or aged 40 and older) is 6.7 percentage points for average-singleton success, 26.9 points for all-separately success, and 29.5 points for both common-action feasibility and ORPM success; the younger group has lower success in every case. Relative to average-singleton success, the common-action requirement therefore adds 22.7 points to the success gap. ORPM matches the common-action disparity in all three displayed settings: attaining the benchmark's feasibility does not remove the unequal burden associated with multiplicity. The ORPM/common-action gap is 16.4, 29.5, and 33.5 points for 6, 12, and 24 trees, respectively, illustrating a larger observed gap for larger sets. Additional results for tree-based competing sets are provided in Appendix~\ref{sec:app-exp-tree-based-competing-sets}.

\section{Discussion and Future Work}
\label{sec:disc}

We formulate recourse under predictive multiplicity as budget-constrained minimization of the worst-case validity loss over a specified competing model set. Requiring one action to satisfy every model restricts the set of successful actions relative to single-model recourse and can increase the minimum budget required for success, although a strict cost increase is not inevitable. This is consistent with prior work examining the relationship between robustness requirements and recourse cost~\citep{UpadhyayJL21, KyawKJ25}. Our formulation makes this tension explicit by optimizing worst-case loss within an allowable budget, rather than directly minimizing cost. Nevertheless, successful recourse may remain unavailable at the chosen budget, and good performance on the competing set does not always establish validity for unseen models.

Our framework accommodates competing models with different architectures or model classes without requiring proximity in parameter space. However, models with similar aggregate predictive performance can impose conflicting requirements on individual recourse. Defining a meaningful competing set therefore requires explicit choices about predictive quality, admissible models, and deployment conditions. Our disparity results additionally show that robustness across models does not ensure equal success across groups, and increasing the budget need not reduce disparities.

Future work could extend the framework to continuous or infinite Rashomon sets~\citep{CoupkovaB24}, combining model exploration methods~\citep{HsuBSLC24, HsuLH024} with procedures that search for models violating a proposed recourse. Establishing guarantees for such extensions would require controlling both recourse optimization error and approximation of the competing set. Another direction is to jointly learn competing models and recourse policies, balancing predictive accuracy, model diversity, and recourse feasibility. Incorporating equity objectives and action constraints is also left as future work. Finally, explicitly modeling how recourse changes individuals' behavior and subsequent training data can address robustness under endogenous model updates.

\ifarxiv\paragraph{Ethical Considerations.}
\else
\noindent\textbf{Ethical Considerations.}
\fi
Recourse helps individuals with actionable recommendations on how to improve outcomes. Predictive multiplicity can create opportunities to identify less discriminatory models~\cite{LauferRB25, HaysLB+26, DaiRYBN25}, though this is not always the case~\cite{JainWC+25}. Understanding how to provide recourse that is robust to predictive multiplicity is an important reliability concern, though by itself it does not ensure equitable outcomes, as shown in our experiments.
This is consistent with prior work on algorithmic recourse, which showed independently generating recourse for individuals might lead to disproportionate validity or burden across different subpopulations~\citep{UstunSL19}. This is also consistent with prior work on predictive multiplicity, which showed multiplicity exposes some groups to greater prediction instability~\citep{DaiRYBN25}. 

Robustness should therefore not be interpreted as an endorsement of the competing models or their decision criteria. Institutions remain responsible for assessing discriminatory model behavior and whether recommended changes are feasible, affordable, and appropriate for affected individuals. Our computational validity and cost measures do not establish these properties in practice, and observed group differences are descriptive rather than causal findings. Future work should extend the analysis to larger and intersectional groups, incorporate individual circumstances and causal constraints, and develop methods that jointly address robustness, validity, and equitable recourse burden and success.
\ifarxiv
\paragraph{Acknowledgment.} 
We thank Kshitij Kayastha for discussions during the early stages of this work.
\fi

\ifarxiv\paragraph{LLM Usage.}
\else
\noindent\textbf{LLM Usage.}
\fi
Gemini 3.1 Pro was used to write the initial version of the experiments from the problem formulation
and algorithm. We used GPT-6 Astra to verify the implementation. We additionally checked all the code. 
GPT-6 Astra was used to proofread the paper, especially in the description of the implementation details and
experimental results. LLMs were not used for problem formulation,
algorithmic solutions, design of experiments, and description of related work.

\bibliographystyle{plainnat}
\bibliography{bib}
%%%%%%%%%%%%%%%%%%%%%%%%%%%%%%%%%%%%%%%%%%%%%%%%%%%%%%%%%%%%
%%%%%%%%%%%%%%%%%%%%%%%%%%%%%%%%%%%%%%%%%%%%%%%%%%%%%%%%%%%%
\appendix
\section{Omitted Details from Section~\ref{sec:algorithm}}
\label{sec:appx-proof}
\begin{proof}[Proof of Theorem~\ref{thm:convergence}]
By the definition of the mixture gap, the definition of
$\bar{w}$, and linearity of $\mathcal{L}(x,\cdot)$,
\begin{align*}
\operatorname{Gap}_{\mathrm{mix}}(x_{1:T},\bar{w})
&=
\max_{w\in\Delta(H)}
\frac1T\sum_{t=1}^T\mathcal{L}(x_t,w)
-
\inf_{x\in K}
\frac1T\sum_{t=1}^T\mathcal{L}(x,w_t)\\
&=R_{\mathrm{adv}}+R_{\mathrm{rec}},
\end{align*}
where
\begin{align*}
R_{\mathrm{adv}}
&:=
\max_{w\in\Delta(H)}
\frac1T\sum_{t=1}^T\mathcal{L}(x_t,w)
-
\frac1T\sum_{t=1}^T\mathcal{L}(x_t,w_t),\\
R_{\mathrm{rec}}
&:=
\frac1T\sum_{t=1}^T\mathcal{L}(x_t,w_t)
-
\inf_{x\in K}
\frac1T\sum_{t=1}^T\mathcal{L}(x,w_t).
\end{align*}
We bound the two terms separately.

First, suppose $n\geq2$. We bound $R_{\mathrm{adv}}$ using
the standard exponential-weights analysis~\citep{AroraHK12}.
Recall that $v_{t,i}=\ell(h_i(x_t),1)\in[0,B]$ by
Assumption~\ref{assumpt:bounded-loss}.
Write the updates in unnormalized form:
\[
q_{1,i}=1,
\qquad
q_{t+1,i}=q_{t,i}\exp(\eta v_{t,i}),
\qquad
Z_t=\sum_{i=1}^n q_{t,i},
\qquad
w_{t,i}=\frac{q_{t,i}}{Z_t}.
\]
Hoeffding's lemma gives
\[
\ln\frac{Z_{t+1}}{Z_t}
=
\ln\left(\sum_{i=1}^n
w_{t,i}\exp(\eta v_{t,i})\right)
\leq
\eta\mathcal{L}(x_t,w_t)+\frac{\eta^2B^2}{8}.
\]
Summing over rounds and telescoping, using $Z_1=n$, yields
\[
\ln Z_{T+1}-\ln n
=
\sum_{t=1}^T\ln\frac{Z_{t+1}}{Z_t}
\leq
\eta\sum_{t=1}^T\mathcal{L}(x_t,w_t)
+\frac{T\eta^2B^2}{8}.
\]
On the other hand, since
$q_{T+1,i}=\exp(\eta\sum_{t=1}^T v_{t,i})$,
\begin{align*}
\ln Z_{T+1}
&=
\ln\left(\sum_{i=1}^n
\exp\left(\eta\sum_{t=1}^T v_{t,i}\right)\right)
\geq
\ln\left(
\exp\left(\eta\max_{1\leq i\leq n}
\sum_{t=1}^T v_{t,i}\right)\right)
=
\eta\max_{1\leq i\leq n}\sum_{t=1}^T v_{t,i},
\end{align*}
where the inequality holds because a sum of positive terms
is at least its largest term. Combining these bounds gives
\[
\eta\max_{1\leq i\leq n}\sum_{t=1}^T v_{t,i}
\leq
\ln n+\eta\sum_{t=1}^T\mathcal{L}(x_t,w_t)
+\frac{T\eta^2B^2}{8}.
\]
Rearranging and dividing by $\eta T$,
\[
\frac1T\max_{1\leq i\leq n}\sum_{t=1}^T v_{t,i}
-
\frac1T\sum_{t=1}^T\mathcal{L}(x_t,w_t)
\leq
\frac{\ln n}{\eta T}+\frac{\eta B^2}{8}.
\]
Maximizing a linear function over $\Delta(H)$ selects a
model with the largest cumulative loss, so
\[
\max_{w\in\Delta(H)}
\frac1T\sum_{t=1}^T\mathcal{L}(x_t,w)
=
\frac1T\max_{1\leq i\leq n}\sum_{t=1}^T v_{t,i}.
\]
Thus, the left-hand side of the preceding regret inequality
is $R_{\mathrm{adv}}$, and
\[
R_{\mathrm{adv}}
\leq
\frac{\ln n}{\eta T}+\frac{\eta B^2}{8}.
\]

Second, Assumption~\ref{assumpt:oracle} gives, for each $t$,
\[
\mathcal{L}(x_t,w_t)
\leq
\inf_{x\in K}\mathcal{L}(x,w_t)+\epsilon_t.
\]
Consequently,
\begin{align*}
R_{\mathrm{rec}}
&\leq
\frac1T\sum_{t=1}^T
\inf_{x\in K}\mathcal{L}(x,w_t)
-
\inf_{x\in K}
\frac1T\sum_{t=1}^T\mathcal{L}(x,w_t)
+
\frac1T\sum_{t=1}^T\epsilon_t
\leq\bar\epsilon_T,
\end{align*}
because the average of the infima is at most the infimum
of the average. Combining the two bounds proves
Equation~\ref{eq:general-regret-bound}.
Substituting $\eta=B^{-1}\sqrt{\ln n/T}$ gives
\[
\operatorname{Gap}_{\mathrm{mix}}(x_{1:T},\bar{w})
\leq
\frac98 B\sqrt{\frac{\ln n}{T}}+\bar\epsilon_T
\leq
2B\sqrt{\frac{\ln n}{T}}+\bar\epsilon_T.
\]
For $n=1$, the adversary has only one action and
$R_{\mathrm{adv}}=0$. The conclusion follows directly
from $R_{\mathrm{rec}}\leq\bar\epsilon_T$.
\end{proof}

\begin{proof}[Proof of Corollary~\ref{cor:convex-convergence}]
Since each $x_t\in K$ and $K$ is convex, $\bar{x}\in K$.
Also, $\bar{w}\in\Delta(H)$.
Convexity of $\mathcal{L}(\cdot,w)$ implies, for every
$w\in\Delta(H)$,
\[
\mathcal{L}(\bar{x},w)
\leq
\frac1T\sum_{t=1}^T\mathcal{L}(x_t,w).
\]
Maximizing over $w$ and subtracting the same term
$\inf_{x\in K}\mathcal{L}(x,\bar{w})$ from both sides yields
\[
\operatorname{Gap}(\bar{x},\bar{w})
\leq
\operatorname{Gap}_{\mathrm{mix}}(x_{1:T},\bar{w}).
\]
The result now follows from Theorem~\ref{thm:convergence}.
\end{proof}

\ifarxiv\paragraph{Deterministic Certificate for Tree-Based Models.}
\else
\noindent\textbf{Deterministic Certificate for Tree-Based Models.}
\fi
\begin{proposition}[Deterministic Recourse Certificate]
\label{prop:deterministic-certificate}
Let $C_T\subseteq K$ be a finite set of feasible candidates
and suppose
$b_t\leq\inf_{x\in K}\mathcal{L}(x,w_t)$ for each $t$.
Define
\[
\begin{aligned}
L_T&:=\max_{1\leq t\leq T}b_t,\quad 
U_T:=\min_{x\in C_T}
\max_{1\leq i\leq n}\ell(h_i(x),1), \quad 
V_{\mathrm{pure}}
:=\inf_{x\in K}
\max_{1\leq i\leq n}\ell(h_i(x),1).
\end{aligned}
\]
Then $L_T\leq V_{\mathrm{pure}}\leq U_T$, and a candidate
attaining $U_T$ has deterministic objective suboptimality
at most $U_T-L_T$.
\end{proposition}
\begin{proof}
For any $x\in K$ and $w_t\in\Delta(H)$,
\[
\mathcal{L}(x,w_t)
\leq
\max_{1\leq i\leq n}\ell(h_i(x),1).
\]
Taking infima over $x$ shows that
\[
b_t
\leq
\inf_{x\in K}\mathcal{L}(x,w_t)
\leq
V_{\mathrm{pure}}.
\]
Taking the maximum over $t$ gives $L_T\leq V_{\mathrm{pure}}$.
Since $C_T\subseteq K$, every candidate's worst-model loss
is at least $V_{\mathrm{pure}}$, and hence
$V_{\mathrm{pure}}\leq U_T$.
If $\widehat{x}_T$ attains $U_T$, then
\[
0
\leq
\max_{1\leq i\leq n}\ell(h_i(\widehat{x}_T),1)
-
V_{\mathrm{pure}}
=
U_T-V_{\mathrm{pure}}
\leq U_T-L_T.
\]
\end{proof}

This certificate is useful when the bounds are tight; the
mixture convergence theorem does not imply that $U_T-L_T$
vanishes. Accordingly, the experiments distinguish the convex
point gap, the tree response-mixture gap and deterministic
certificates, and empirical recourse quality for the
differentiable nonconvex oracles. A small optimization gap
does not by itself ensure validity across all competing
models, which may be impossible for the given budget and
competing set.

\ifarxiv\paragraph{Evaluating Gaps with an Approximate Optimization Solver.}
\else
\noindent\textbf{Evaluating Gaps with an Approximate Optimization Solver.}
\fi
Let $D(\bar{w}):=\inf_{x\in K}\mathcal{L}(x,\bar{w})$,
and suppose a separate solver provides bounds
$b\leq D(\bar{w})\leq u$.
For the mixture gap, set
$P=\max_i T^{-1}\sum_t\ell(h_i(x_t),1)$;
for a feasible deterministic point $\widehat{x}$, set
$P=\max_i\ell(h_i(\widehat{x}),1)$.
In either case, the corresponding gap belongs to
$\bigl[\max\{0,P-u\},\ P-b\bigr].$
A feasible local solution to the weighted minimization
problem provides an upper bound $u$, and therefore a lower
bound on the gap; it cannot by itself certify that the gap
is small. This optimization interval is distinct from a
confidence interval across experimental seeds.

\ifarxiv\paragraph{Convergence Bounds at Intermediate Checkpoints.}
\else
\noindent\textbf{Convergence Bounds at Intermediate Checkpoints.}
\fi
For a run using a fixed learning rate $\eta$, the proof of
Equation~\ref{eq:general-regret-bound} applies to every prefix
of $s\leq T$ rounds, giving the bound
$\ln n/(\eta s)+\eta B^2/8+s^{-1}\sum_{t=1}^s\epsilon_t$
when $n\geq2$.
A rate obtained by substituting
$\eta=B^{-1}\sqrt{\ln n/s}$ instead describes a learning rate
chosen for horizon $s$ and should not be substituted into a
fixed-horizon trajectory without changing its updates.

\ifarxiv\paragraph{Why the Nonconvex Distinction Matters.}
\else
\noindent\textbf{Why the Nonconvex Distinction Matters.}
\fi
Consider a feasible set with two recourses whose loss vectors
across two models are $(0,1)$ and $(1,0)$.
Every deterministic recourse has worst-model loss $1$, while
their uniform mixture has worst expected loss $1/2$.
Against the uniform model distribution, that mixture has zero
gap, but either deterministic recourse has point gap $1/2$
despite being optimal for the deterministic problem.
Thus, mixture convergence does not imply deterministic
convergence, and a positive nonconvex point gap need not
indicate a suboptimal deterministic recourse.

\section{Omitted Details from Section~\ref{sec:exp}}
\label{sec:app-exp}

\begin{table}[ht!]
\centering
\small\mdseries
\setlength{\tabcolsep}{3pt}
\renewcommand{\arraystretch}{1.15}
\begin{tabular}{@{}>{\raggedright\arraybackslash}p{.14\linewidth}>{\centering\arraybackslash}p{.09\linewidth}>{\centering\arraybackslash}p{.09\linewidth}>{\raggedleft\arraybackslash}p{.11\linewidth}>{\raggedright\arraybackslash\bfseries}p{.23\linewidth}>{\raggedright\arraybackslash\bfseries}p{\dimexpr.34\linewidth-10\tabcolsep\relax}@{}}
\toprule
\textbf{Dataset}
& \textbf{Features before encoding}
& \textbf{Model input features}
& \textbf{Observations}
& Favorable label in source data
& Immutable features \\
\midrule
Synthetic & 2 & 2 & 2,000$^{*}$
& $y=1$ (generated positive class)
& None \\
\addlinespace
COMPAS & 7 & 7 & 6,392
& \texttt{two\_year\_recid}$=0$ (no recorded two-year recidivism)
& Age (\texttt{age}); sex (\texttt{sex=female}) \\
\addlinespace
Polish & 64 & 64 & 864
& Recorded target $y=1$
& None \\
\addlinespace
German & 20 & 67 & 600
& Local $y=0$ (good credit)
& Age (\texttt{Attribute13}); dependents (\texttt{Attribute18}); personal status/sex (all \texttt{Attribute9} indicators); foreign-worker status (all \texttt{Attribute20} indicators) \\
\addlinespace
Give Me Some Credit & 10 & 10 & 120,269
& \texttt{SeriousDlqin2yrs}$=0$ (no serious delinquency within two years)
& Age (\texttt{age}); number of dependents (\texttt{NumberOfDependents}) \\
\addlinespace
Folktables (ACSIncome) & 10 & 10 & 195,665
& \texttt{target}$=1$ (annual personal income $>\$50{,}000$)
& Age (\texttt{AGEP}); marital status (\texttt{MAR}); place of birth (\texttt{POBP}); relationship to householder (\texttt{RELP}); sex (\texttt{SEX}); race (\texttt{RAC1P}) \\
\bottomrule
\end{tabular}
\caption{A summary of datasets. $^{*}$The synthetic nonlinear-baseline comparison uses 20,000 observations. The listed source labels are mapped to model class~1; immutable features remain fixed during recourse, including every indicator in a listed one-hot group.}
\label{tab:dataset_stats}
\end{table}

\subsection{Datasets}
\label{sec:app-exp-datasets}
For the synthetic dataset, we used 2,000 observations for linear baselines and 20,000 observations from the same distribution for non-linear baselines. The goal is to predict which of the two Gaussian distributions generated the observation.
For COMPAS, Polish, and German datasets (processed from the CSV files in the datasets folder), we removed non-feature index columns, converted entries to numeric values, and removed rows containing missing or non-numeric values. We then balanced the two labels by randomly undersampling the majority class without replacement, using a fixed seed of 42. The resulting datasets contain 6,392, 864, and 600 observations for each of the datasets, respectively. The German dataset originally contained 1,000 observations and 20 attributes. After balancing the labels and one-hot encoding, the observations decreased to 600 with 67 input coordinates. For the COMPAS dataset, the goal is to predict whether the defendant is arrested within two years. For the Polish dataset, the goal is to predict a company's bankruptcy status. The dataset contains 864 observations on 64 financial indicators. For the German dataset, the goal is to classify an applicant with good or bad credit. For the Give Me Some Credit dataset, we retained the 10 predictors and removed rows with missing values, leaving 120,269 labeled observations. We did not balance this dataset. The goal is to predict serious delinquencies (being at least 90 days past due) within the following two years. For Folktables, we use the standard ACSIncome task on California’s 2018 one-year ACS person-level data. Its eligibility filter retains individuals older than 16, with annual personal income above \$100, positive usual weekly working hours, and a person weight of at least one. The goal is to predict whether annual personal income exceeds \$50,000. This produces 195,665 observations with 10 predictors. We retain the numeric feature codes without additional one-hot encoding and do not balance the classes. See Table~\ref{tab:dataset_stats} for a summary of the statistics of each dataset. 

For each experimental seed, we construct stratified training, validation, and test splits containing approximately 64\%, 16\%, and 20\% of the observations, respectively. We fit the feature standardization using only the training split and apply the same transformation to validation and test observations. 

\subsection{Implementation Details}
\label{sec:app-exp-details}

\ifarxiv\paragraph{Models and Training.}
\else
\noindent\textbf{Models and Training.}
\fi
We implemented logistic regression and neural networks in PyTorch.
Logistic regression consists of an affine logit followed by a sigmoid.
The first neural architecture has one fully connected hidden layer
with 20 units; the second has two fully connected hidden layers with
50 and 100 units, respectively. Both architectures use ReLU hidden
activations and a sigmoid output. The main training procedure uses
full-batch Adam for 100 epochs with learning rate $0.01$, minimizing
binary cross-entropy through its numerically stable logit formulation.

Decision trees were trained using Scikit-Learn.
The larger tree experiments use maximum depth 4, a minimum of 5
training observations per leaf, and square-root feature subsampling.

\ifarxiv\paragraph{Data Splitting and Instance Selection.}
\else
\noindent\textbf{Data Splitting and Instance Selection.}
\fi
For each seed, we first reserved 20\% of the data for testing and
then reserved 20\% of the remaining data for validation, producing
approximately 64\% training, 16\% validation, and 20\% testing data.
Both splits were stratified by label. Feature standardization was
fitted exclusively on the training split. Models were trained on the
training data and admitted to the competing set using validation
accuracy.

The first model in the competing set serves as the reference model in the reference baseline.
Recourse is generated only for test instances for which at least half of the models in the 
competing set assign a probability below $0.5$. The same models and factual instances
are used across methods and budgets within each experimental
configuration. Unless a numerical acceptance convention is explicitly specified, 
a returned point is counted as jointly valid only if it satisfies the budget and feature 
constraints and every competing model assigns it probability at least $0.5$. 
Table~\ref{tab:neural-orpm-exact} uses the logit tolerance described below.

\ifarxiv\paragraph{Construction of Differentiable Competing Sets.}
\else
\noindent\textbf{Construction of Differentiable Competing Sets.}
\fi
In the linear baseline comparison, we trained a central logistic
regression model and sampled additional parameter vectors from its
local Hessian ellipsoid, following the geometric construction
of~\citet{TurbalVS25}. Including the intercept, the sampling region is
$(\theta-\theta_0)^\top A(\theta-\theta_0)
    \leq 2\kappa,$
where $A$ is the Hessian of the average training binary cross-entropy,
regularized by $10^{-5}I$, and $\kappa=0.05$.

The mixed-model experiments train logistic regressions and neural networks with different initializations. 
Feature-diverse replacements retain approximately 50\% or 75\% of predictor groups, including possible omission of 
actionable groups. Candidates must pass the validation-accuracy floor and lie within 0.02 accuracy of their slot's 
control model. Greedy selection favors validation disagreement, with control-model fallback when no eligible 
replacement exists. 

For both construction procedures, a candidate is admitted when its
validation accuracy is no more than $0.05$ below that of the central
logistic regression model. This is a one-sided admission criterion;
candidates with higher accuracy are also admitted. These are
empirically filtered competing sets, rather than certified
enumerations of a Rashomon set. The floor is relative to central-logistic accuracy, 
not to the best available model.
The mixed-model trade-off experiments
use $(2,2,2)$, $(4,4,4)$, and $(8,8,8)$ logistic,
single-hidden-layer, and double-hidden-layer models, respectively,
giving competing sets of 6, 12, and 24 models.

\ifarxiv\paragraph{Construction of Tree Competing Sets.}
\else
\noindent\textbf{Construction of Tree Competing Sets.}
\fi
Tree models are trained on the same training split using different
random seeds. Trained models must have validation accuracy
within $0.05$ of the best model in the generated optimization pool.
Separately generated evaluation models must lie within a two-sided
$0.05$ accuracy band around that reference and are never queried by
the recourse optimizers. 

\ifarxiv\paragraph{Recourse Domain and Loss.}
\else
\noindent\textbf{Recourse Domain and Loss.}
\fi
We use standardized $L_2$ distance as the cost, fixing the listed immutable coordinates. 
The differentiable trade-off and same-architecture neural experiments use raw logistic 
loss $\ell_j(z)=\log(1+\exp[-g_j(z)])$, where $g_j$ is model $j$'s logit. Tree experiments use 
squared probability loss. The convex duality-gap experiment uses the separately specified normalized 
logistic loss.

\ifarxiv\paragraph{Differentiable ORPM Oracle.}
\else
\noindent\textbf{Differentiable ORPM Oracle.}
\fi
Neural ORPM starts with uniform weights, updates $w_{t+1,j}\propto w_{t,j}\exp(\ell_j(z_t))$, and 
returns the arithmetic mean of its responses. Each response uses SLSQP with at most 100 iterations 
and tolerance $10^{-8}$, initialized at the previous response (the factual point initially). 
The best weighted-loss point among the factual point, initialization, and projected proposal is 
retained. We use 200 rounds for every dataset except GMC, where we use 1,000 rounds. 
These are local neural optimization runs with raw outer rate 1; 
they do not establish global-oracle assumptions.

\ifarxiv\paragraph{Tree-based ORPM Oracle.}
\else
\noindent\textbf{Tree-based ORPM Oracle.}
\fi
Following the mixed-integer approach to tree recourse
of~\citet{ParmentierV21}, the oracle jointly encodes the competing
models' leaf selections, predictions, immutable features, and
recourse budget. With squared probability loss and an $L^2$ budget,
the implementation is a mixed-integer quadratically constrained
optimization problem. It is solved using Gurobi, reusing the
formulation across rounds and updating the model weights and
warm start. The expanded tree experiments allow one second per ORPM response solve, 
use one solver thread, and request an absolute optimality gap of $10^{-5}$. 
ORPM uses 100 rounds in the expanded tree budget-validity 
comparison and 200 in the tree convergence experiments. 
These are computational limits and stopping targets,
not guarantees that every call attains that accuracy. Returned
candidates are checked against the original models and feasibility
constraints. Solver status, objective bounds, and verified incumbent
values are retained. Any resulting optimality certificate is subject
to the solver's numerical tolerances. Tree ORPM returns the candidate with the smallest worst-model loss
among the factual point and oracle responses, breaking ties by cost.
It does not average feature vectors. Consequently, the convex theorem
for an averaged recourse point does not directly justify this return
rule.

\ifarxiv\paragraph{Differentiable Baselines.}
\else
\noindent\textbf{Differentiable Baselines.}
\fi
We use the following approaches as baselines.
\begin{itemize}
\item \textbf{ADV:} ADV applies 500 projected Adam steps with learning rate $0.01\delta$, 
where $\delta$ is the recourse budget, directly to the largest loss among the competing models. 
Each step differentiates through a currently worst-loss model. It returns its best observed worst-loss point, 
without tolerance-based early stopping.
\item \textbf{ROAR:} Our ROAR implementation~\citep{UpadhyayJL21} uses a fixed
$L^\infty$ uncertainty box around the deployed model's affine
parameters, including its intercept. The radius is the largest
coordinatewise parameter difference between the deployed model and
the competing models. For nonlinear models, these parameters come
from factual-point LIME approximations of the
logits~\citep{RibeiroSG16}, anchored to match each original model
at that point. Recourse is computed by maximizing the box's
worst-case logit using SLSQP, subject to the common budget and
feature constraints. ROAR uses SLSQP with at most 500 iterations and tolerance $10^{-4}$. 
For neural models, local logit surrogates use 5,000 samples.
\item \textbf{ElliCE:} Our implementation of ElliCE~\citep{TurbalVS25} uses the training-loss Hessian ellipsoid
centered on the deployed model. For neural networks, the feature
extractor is fixed, and uncertainty is applied to the final affine
layer. So unlike ROAR, LIME is not used for non-linear models. SLSQP first searches for a favorable
worst-case ellipsoid score and, when successful and iterations
remain, reduces $L^2$ cost while preserving that score.
ElliCE uses SLSQP tolerance $10^{-4}$ and
experiment-specific iteration caps.
\end{itemize}

\ifarxiv
\paragraph{Neural Optimization Baseline in Table~\ref{tab:neural-orpm-exact}.}
\else
\noindent\textbf{Neural Optimization in Table~\ref{tab:neural-orpm-exact}.}
\fi
The neural references in Table 1b use the continuous action region
\[
\mathcal K_{\delta}(x)=\{z:\|z-x\|_2\leq \delta,\ z_j=x_j\text{ for immutable }j\}.
\]
Writing $g_j(z)$ for network $j$'s logit, MaxMargin maximizes the minimum logit, while MinL2 minimizes the distance required for acceptance by every seen network. ReLU encodings and the quadratic budget make these MIQCPs. They are our implementation-specific optimization references, related to prior work on joint neural verification~\citep{LeofanteBR23}.

For the raw logistic loss used here,
\[
\max_j\log\bigl(1+\exp[-g_j(z)]\bigr)
=\log\bigl(1+\exp[-\min_j g_j(z)]\bigr).
\]
Thus MaxMargin and worst-logistic-loss minimization have the same global optimizers on $\mathcal K_{\delta}(x)$. Comparisons concern the returned solutions under the stated computational limits: outperforming a time-limited MaxMargin output does not imply outperforming the global minimax optimum.

\begin{tabularx}{\linewidth}{@{}>{\raggedright\arraybackslash}p{.15\linewidth}>{\raggedright\arraybackslash}X>{\raggedright\arraybackslash}X@{}}\toprule
 & MaxMargin & MinL2\\\midrule
Objective & $\max_{z,t}t$, with $g_j(z)\geq t$ for every seen network & $\min_z\tfrac12\|z-x\|_2^2$, with $g_j(z)\geq0$ for every seen network\\
Domain & $z\in\mathcal K_{\delta}(x)$ & $z\in\mathcal K_{{\delta}_{\max}}(x)$\\
Solver/cap & Gurobi, one thread, 60 seconds per positive-budget solve & Gurobi, one thread, 60 seconds once per person/model set at the largest budget; reuse for smaller budgets\\
Return & Verified incumbent or factual point, whichever has the larger minimum logit & Verified incumbent when its distance fits the queried budget; otherwise mark the result unavailable/unresolved as appropriate\\
Unavailable output & Factual fallback, evaluated explicitly & Factual placeholder; primary validity is zero, and cost is missing, not zero\\\bottomrule
\end{tabularx}
The native relative and absolute gap targets are $10^{-5}$ and $10^{-6}$; feasibility and integrality tolerances are $10^{-9}$. 
For Table~\ref{tab:neural-orpm-exact}, primary acceptance uses $g_j(z)\geq-10^{-7}$ for every evaluated network, with the same convention for ORPM, MaxMargin and MinL2 on both seen and unseen panels. Strict zero-threshold results are retained separately, and budget checks allow $10^{-8}$.
Timeouts retain their native status and do not establish infeasibility. Solver bounds are numerical certificates, not exact-arithmetic proofs. No ORPM warm starts or replacement actions are used. The 60-second limit excludes formulation and shared-preparation time.

\ifarxiv\paragraph{Tree-based Baselines.}
\else
\noindent\textbf{Tree-based Baselines.}
\fi
We include the following explicitly labeled adaptations:
\begin{itemize}
    \item \textbf{ADV-surrogate:}
    projected optimization using smooth sigmoid approximations to
    tree splits, inspired by FOCUS~\citep{LucicOH+22}, with
    temperature $0.1$. Candidate quality is checked using the
    original tree models.

    \item \textbf{ROAR-surrogate:}
    the parameter-box method applied to anchored local logit
    surrogates.

    \item \textbf{RobX-constrained:}
    an adaptation of RobX~\citep{DuttaLMTM22} that refines a
    deployed-model counterfactual using a local stability score
    and projected training-data anchors. Defaults are 100 Gaussian
    probes with standard deviation $0.1$, stability threshold
    $0.7$, interpolation coefficient $0.1$, and up to three
    anchors selected from 200 candidates.

    \item \textbf{OCEAN-style:}
    minimum-$L^2$-cost recourse for the deployed tree 
    using our mixed-integer formulation, following
    \citet{ParmentierV21}.

    \item \textbf{Direct minimax}: a direct mixed-integer optimization reference for the finite competing-set minimax objective. Its objective and computational allowance are specified in Appendix~\ref{sec:app-exp-tree-based-competing-sets}.
\end{itemize}

The adapted baselines retain their stated budget and search limits;
theoretical guarantees from their original formulations are not
automatically transferred to these adaptations.

\ifarxiv\paragraph{Evaluation of Duality Gap.}
\else
\noindent\textbf{Evaluation of Duality Gap.}
\fi
The convex convergence experiment uses affine-logit models with
logistic loss $
    \widetilde{\ell}_i(x)
    = \log\!\left(1+\exp(-a_i^\top x-b_i)\right),$
divided by a common upper bound over the feasible region so that
all losses lie in $[0,1]$. Its oracle uses float64 SLSQP with at most
200 iterations and tolerance $10^{-10}$, together with a convex
supporting-plane lower bound. At each diagnostic checkpoint, let $P$ denote the primal objective
and let $
    q(\bar w)
    := \min_{x\in K}\mathcal{L}(x,\bar w)$
denote the dual objective. If the diagnostic solver supplies bounds
$L\leq q(\bar w)\leq U$, we report$
    \operatorname{Gap}
    \in
    \left[
        \max\{0,P-U\},
        \;P-L
    \right].$
These optimization-bound intervals are distinct from statistical
confidence intervals. Linear convergence diagnostics use normalized logistic loss; neural/mixed diagnostics use raw logistic loss. The numerical target $0.005$ therefore does not represent a common loss scale across these settings, and attainment rates are interpreted within each stated diagnostic.

For neural sets, we evaluate the saved responses after the trajectory has finished. At each checkpoint, we use the average pre-update weights $\bar w$ and compute both $P_{\rm point}=\max_i\ell_i(\bar z)$ and $P_{\rm mix}=\max_i T^{-1}\sum_{t=1}^T\ell_i(z_t)$. A feasible SLSQP diagnostic response supplies an upper bound on $q(\bar w)=\inf_{z\in\mathcal K_\delta(x)}\sum_i\bar w_i\ell_i(z)$. Lower bounds combine analytic component bounds with a mixed-integer encoding of the saved ReLU networks and piecewise-linear tangent minorants of logistic loss, using a one-second diagnostic allowance per checkpoint and five seconds at the final checkpoint. All quantities use the same loss scale. We report $[\max\{0,P-U\},P-L]$ separately for each primal quantity. Numerical lower bounds contradicted by saved feasible witnesses, including later responses, are discarded together with their derived bounds. Note that these are numerical optimization intervals, not sampling intervals or exact-arithmetic proofs.

Tree diagnostics evaluate both the returned candidate and the
uniform distribution over oracle responses. For the latter, the
primal objective is
$    P_{\mathrm{mix}}
    =
    \max_{i\in\{1,\ldots,n\}}
    \frac{1}{T}\sum_{t=1}^{T}\ell(h_i(x_t),1),
$
rather than the loss at an averaged feature vector.
Diagnostic solves are separate from the ORPM trajectory and do
not influence its updates.

\ifarxiv\paragraph{Cohorts and Aggregation for the Cost-of-Multiplicity Success Curves.}
\else
\noindent\textbf{Cohorts and Aggregation for the Cost-of-Multiplicity Success Curves.}
\fi
For a fixed dataset, model count, and budget, let $n_{sg}$ be the evaluated count and $\widehat p_{sg}$ the mean instance-level success measure in group $g$ of seed $s$. Our cost of multiplicity results use 
\[
 \widehat p=\frac{1}{10}\sum_{s=401}^{410}
 \sum_{g:\,n_{sg}>0}\frac{W_{sg}}{\sum_h W_{sh}}\widehat p_{sg},
\]
as the estimated overall recourse success rate. For linear and tree models, $W_{sg}$ counts \emph{all eligible held-out test instances} in that group's saved model bank: at least half the models reject the factual point in each of the nested 6/12/24-model sets. For neural models, $W_{sg}$ counts group members in the original twenty-instance source cohort, whereas only four instances per seed are evaluated in the cost comparison. Synthetic and Polish have no designated demographic groups and use ordinary within-seed means: twenty instances per seed for linear/trees and four for neural models.

Linear selection retains the original twenty instances and appends instances in the saved eligible order until twenty per group or exhaustion; tree selection retains the earlier cost-study cases and fills to twenty; neural selection retains pilot cases, targets two per group within the original twenty, and fills shortages in saved order. No selected case is dropped from these success curves because recourse fails; unresolved benchmark feasibility remains a lower/upper success range, with the same aggregation applied to each endpoint. The only absent group is recorded women in neural COMPAS seed 409 ($n=0$, $W=0$), which contributes zero weight to its overall rate. Disparity uses $D=|S|^{-1}\sum_{s\in S}(\widehat p_{s1}-\widehat p_{s2})$, without population weights, where $S$ contains seeds with both groups: nine for neural COMPAS (excluding 409), ten otherwise. Plots of gap magnitude use $|D|$. Pointwise 95\% Student-$t$ intervals use the contributing seed means (nine degrees of freedom for overall rates, eight for neural COMPAS disparity). These conventions concern the cost/multiplicity studies and their group contrasts, not the separate trade-off or unseen-model experiments.

For example, German tree seed 401 evaluates 14 instances under 40 and 6 aged 40+, but uses eligible-population counts 41 and 17, hence weights $41/58$ and $17/58$, rather than $14/20$ and $6/20$. Across ten seeds, German ORPM success is 53.4272\% at twelve trees and $\delta=2$ (unweighted: $107/200=53.5\%$); the linear 24-model values are 57.0909\% and 99.6822\% at $\delta=2,5$. Thus, the displayed 53.4\%, 57.1\% and 99.7\% are weighted estimates, not raw percentages out of 200.

\subsection{Additional Experimental Results}
\label{sec:app-exp-additional}

\subsubsection{Duality-Gap}
\label{sec:app-exp-duality}
\begin{figure}[ht!]
\centering
\begin{subfigure}[t]{0.49\textwidth}
\centering\includegraphics[width=\linewidth]{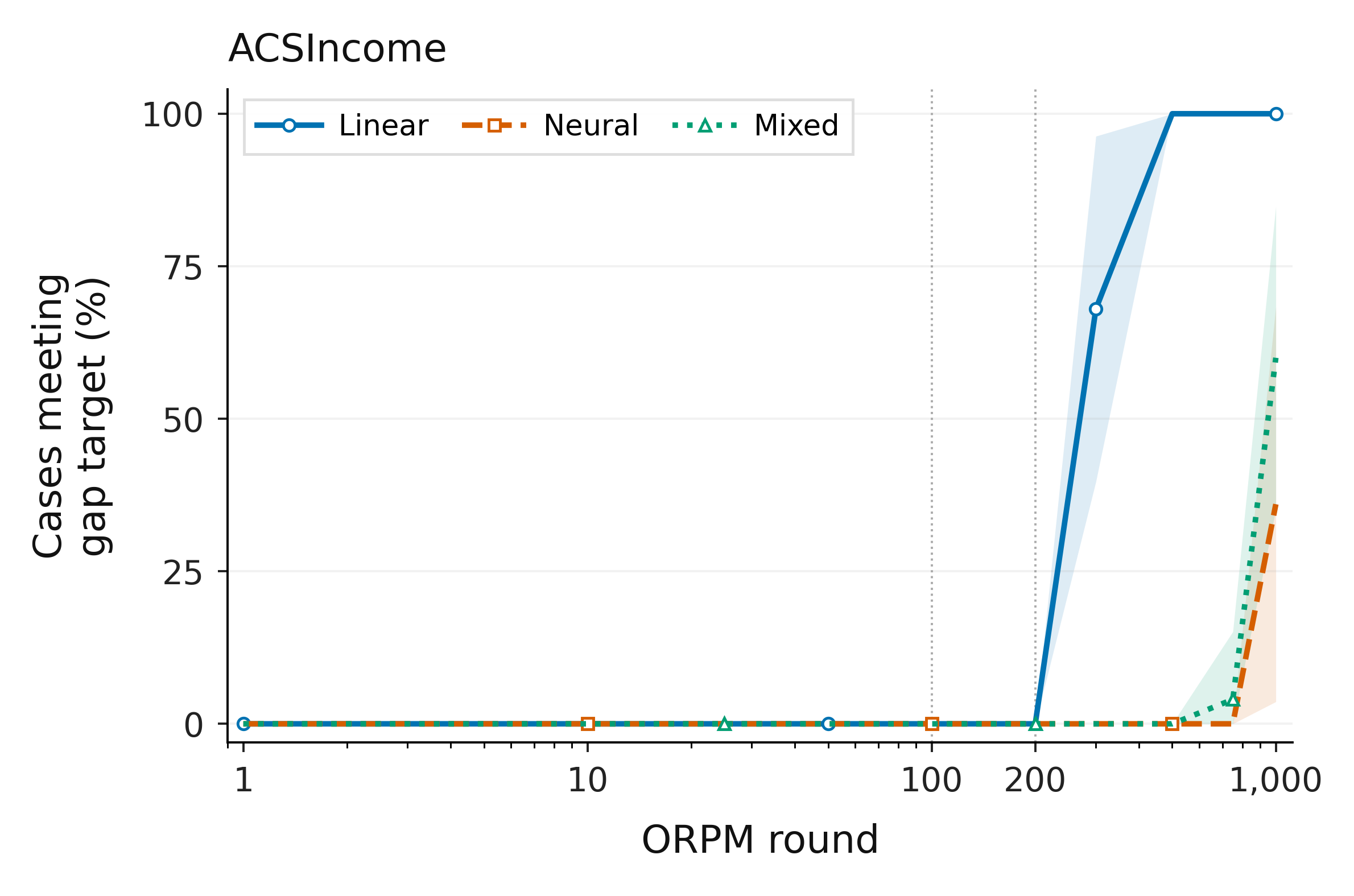}
\caption{ASCIncome dataset, 6 competing models, $\delta=2$.}
\end{subfigure}\hfill
\begin{subfigure}[t]{0.49\textwidth}
\centering\includegraphics[width=\linewidth]{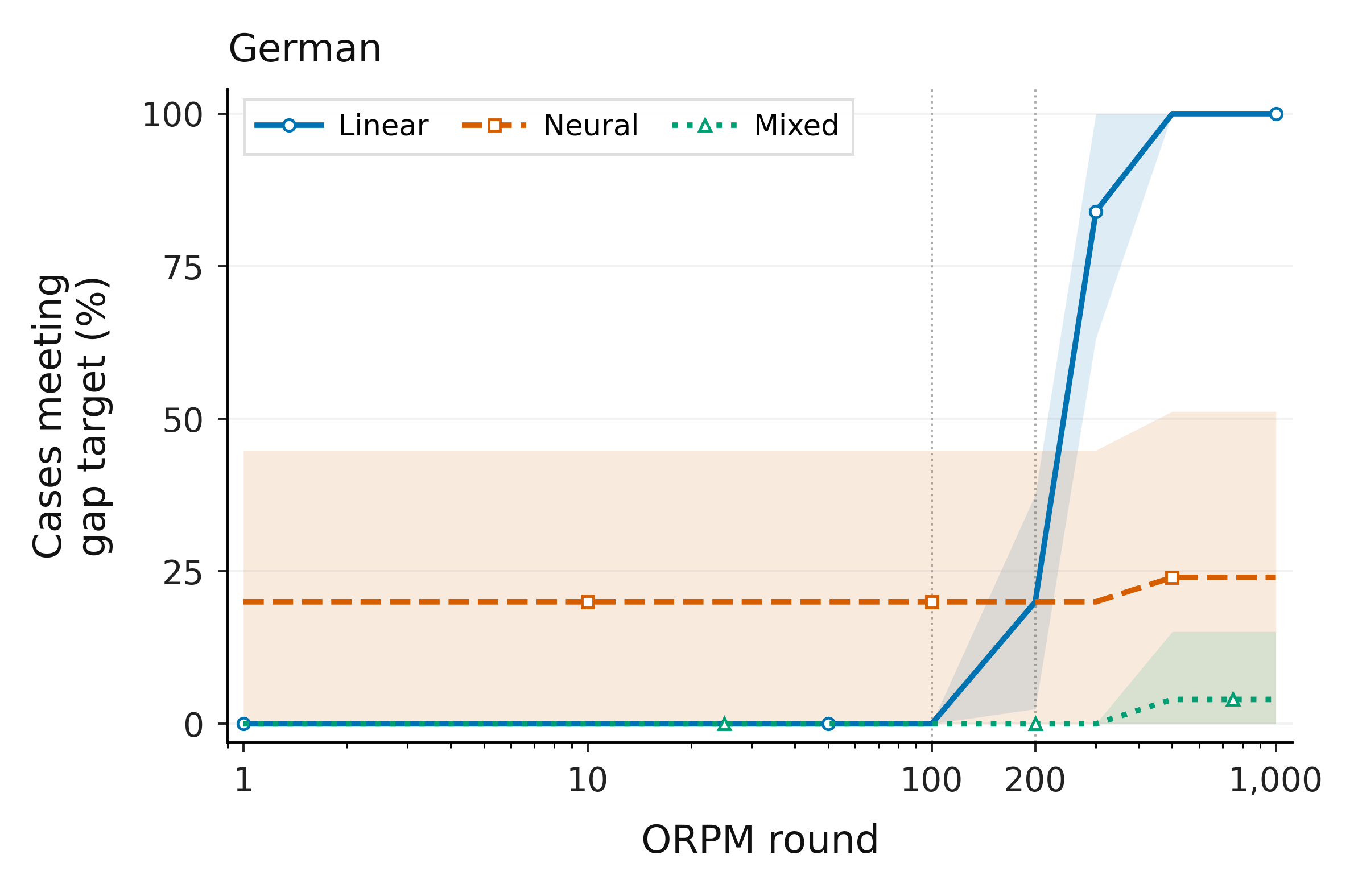}
\caption{German dataset, 6 competing models, $\delta=2$.}
\end{subfigure}
\begin{subfigure}[t]{0.49\textwidth}
\centering\includegraphics[width=\linewidth]{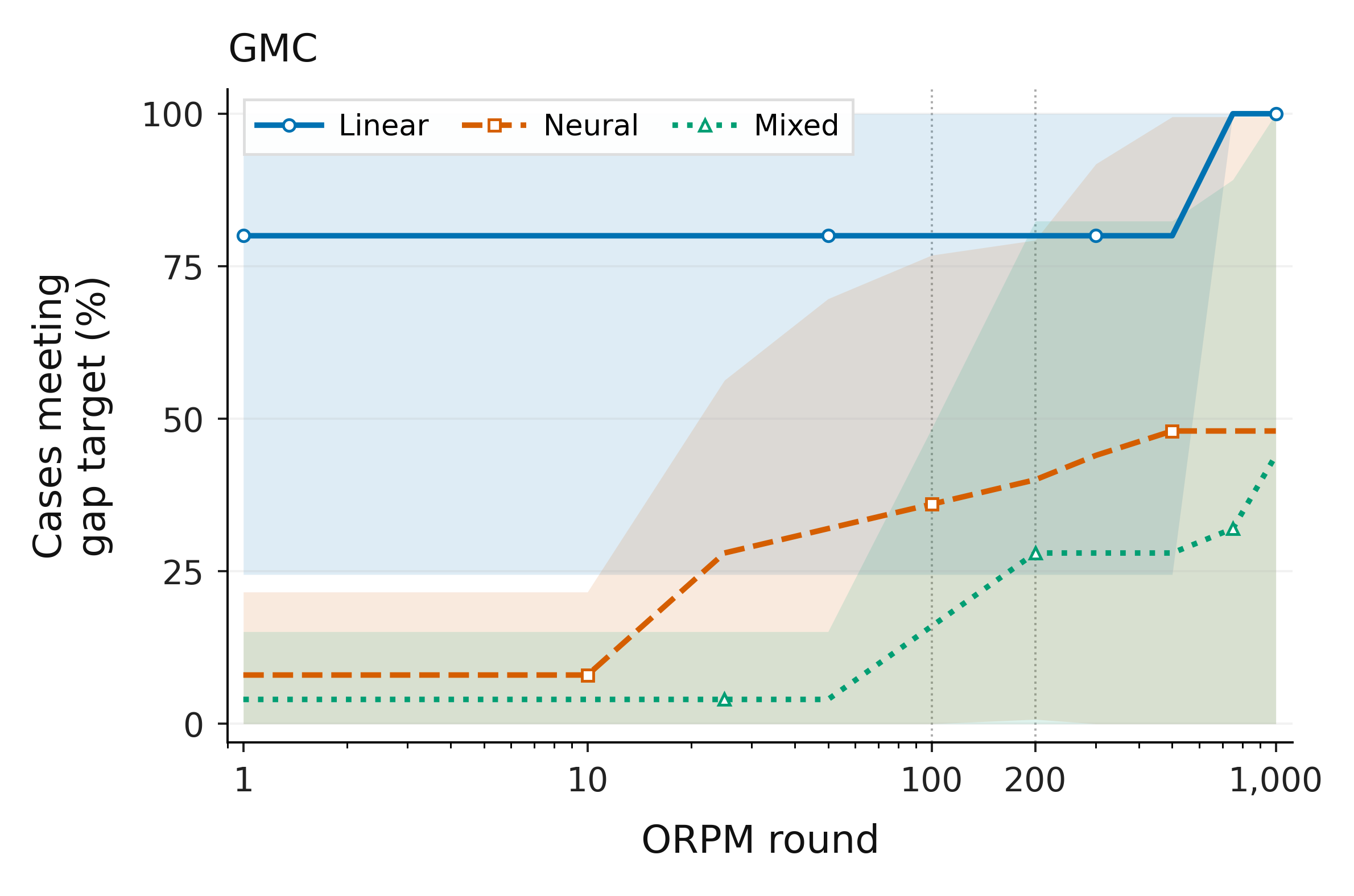}
\caption{GMC dataset, 24 competing models, $\delta=75$.}
\end{subfigure}\hfill
\begin{subfigure}[t]{0.49\textwidth}
\centering\includegraphics[width=\linewidth]{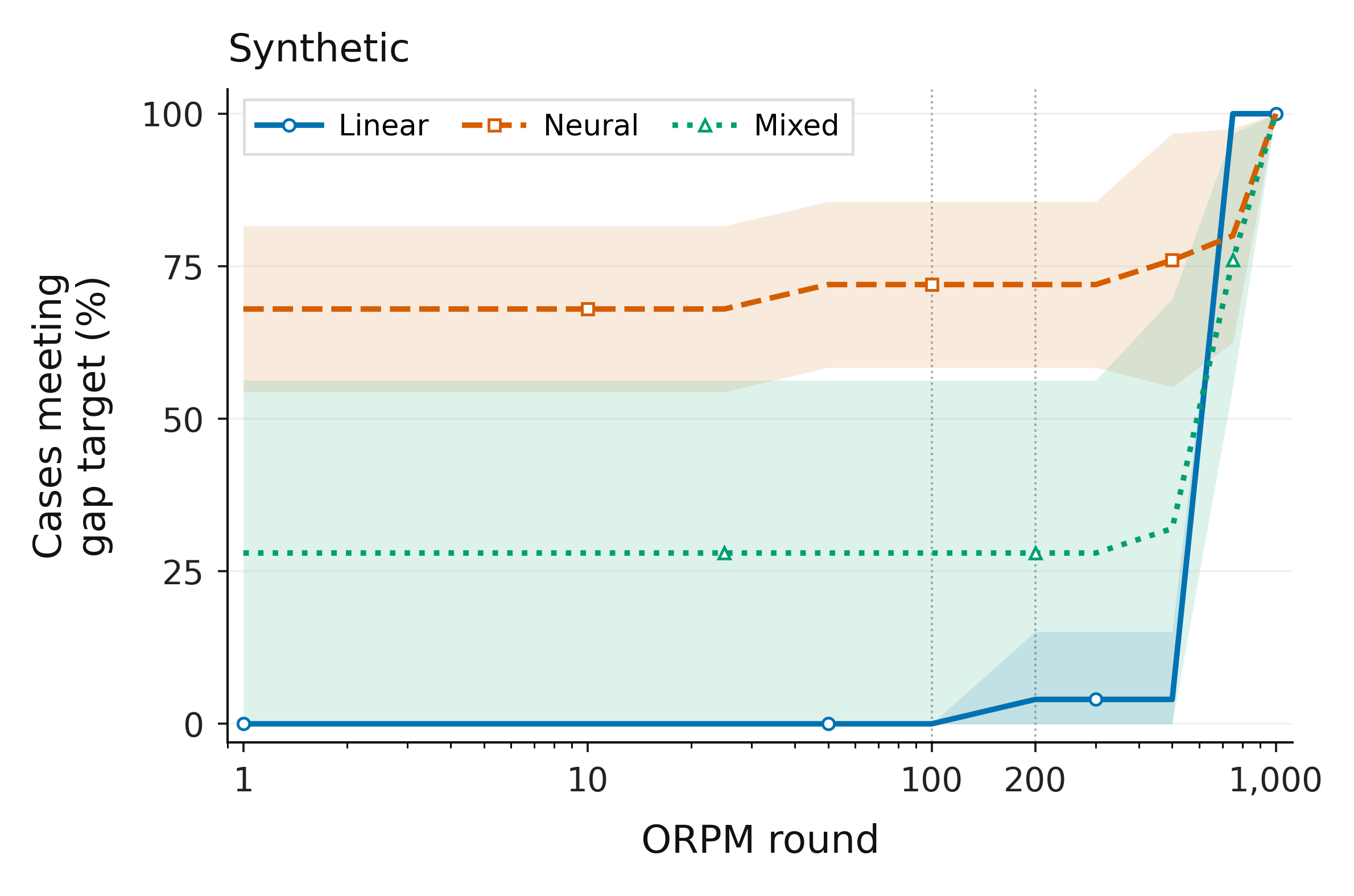}
\caption{Synthetic dataset, 24 competing models, $\delta=2$.}
\end{subfigure}
\caption{Fraction of instances whose certified upper bound on the averaged-point gap is at most 0.005 at each checkpoint. Each subfigure corresponds to a different dataset. The number of models and budget are mentioned in the legend.\label{fig:conv-app}}
\end{figure}

For Synthetic, COMPAS, Polish, and German datasets, we used a budget of $\{1, 2, 5\}$. For GMC, we used a budget of $\{50,60,75\}$. The set of budgets used for ACSIncome is $\{2, 5, 10\}$. Additional convergence results for other datasets not covered in the main body are provided in Figure~\ref{fig:conv-app}.

We also compare three potential return rules for ORPM: the mean of the first $T$ responses (as the theory suggests), the last response, and the recourse with the smallest worst loss among the factual point and the first $T$ responses, breaking exact loss ties by cost. For this analysis, we use a competing set of twelve neural networks (six with one 20-ReLU hidden layer and six with 50/100-ReLU hidden layers), five seeds, and five instances per seed on every dataset, at $\delta=2$ for Synthetic, COMPAS, Polish and German, $60$ for GMC and $5$ for ACSIncome. We observe that joint validity is equal across the three rules on five datasets. On GMC and using $T=200$ rounds, the joint validity is 72\% for averaging and 76\% for last and best, becoming 76\% for all three at $T=1{,}000$.

\subsubsection{Trade-off Between Validity and Budget}
\label{sec:app-exp-cost-validity-tradeoff}
\begin{figure}[ht!]
\centering
\begin{subfigure}[t]{0.49\textwidth}
\centering\includegraphics[width=\linewidth]{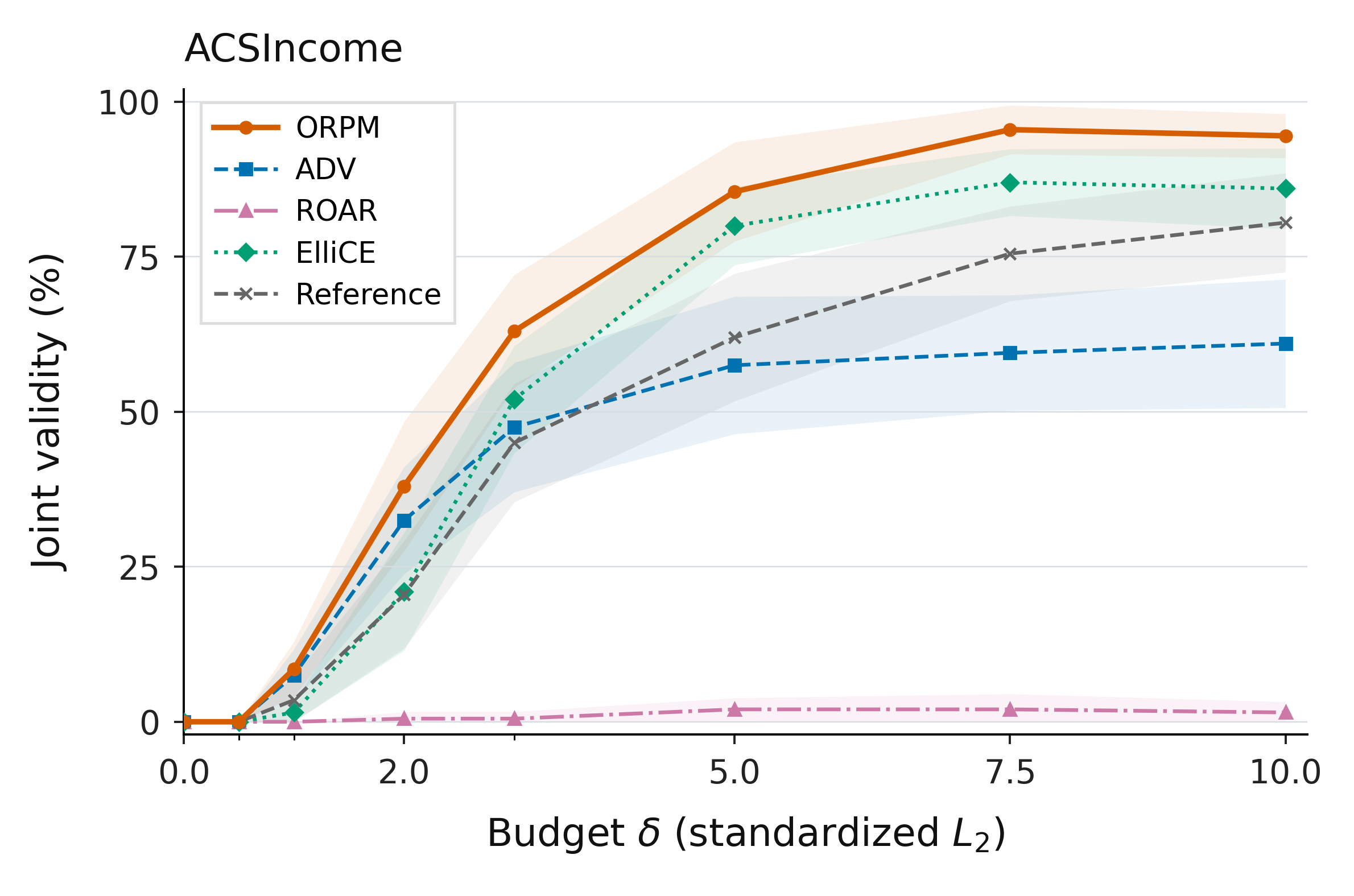}
\caption{ACSIncome dataset, mixed competing set with 12 models}
\end{subfigure}\hfill
\begin{subfigure}[t]{0.49\textwidth}
\centering\includegraphics[width=\linewidth]{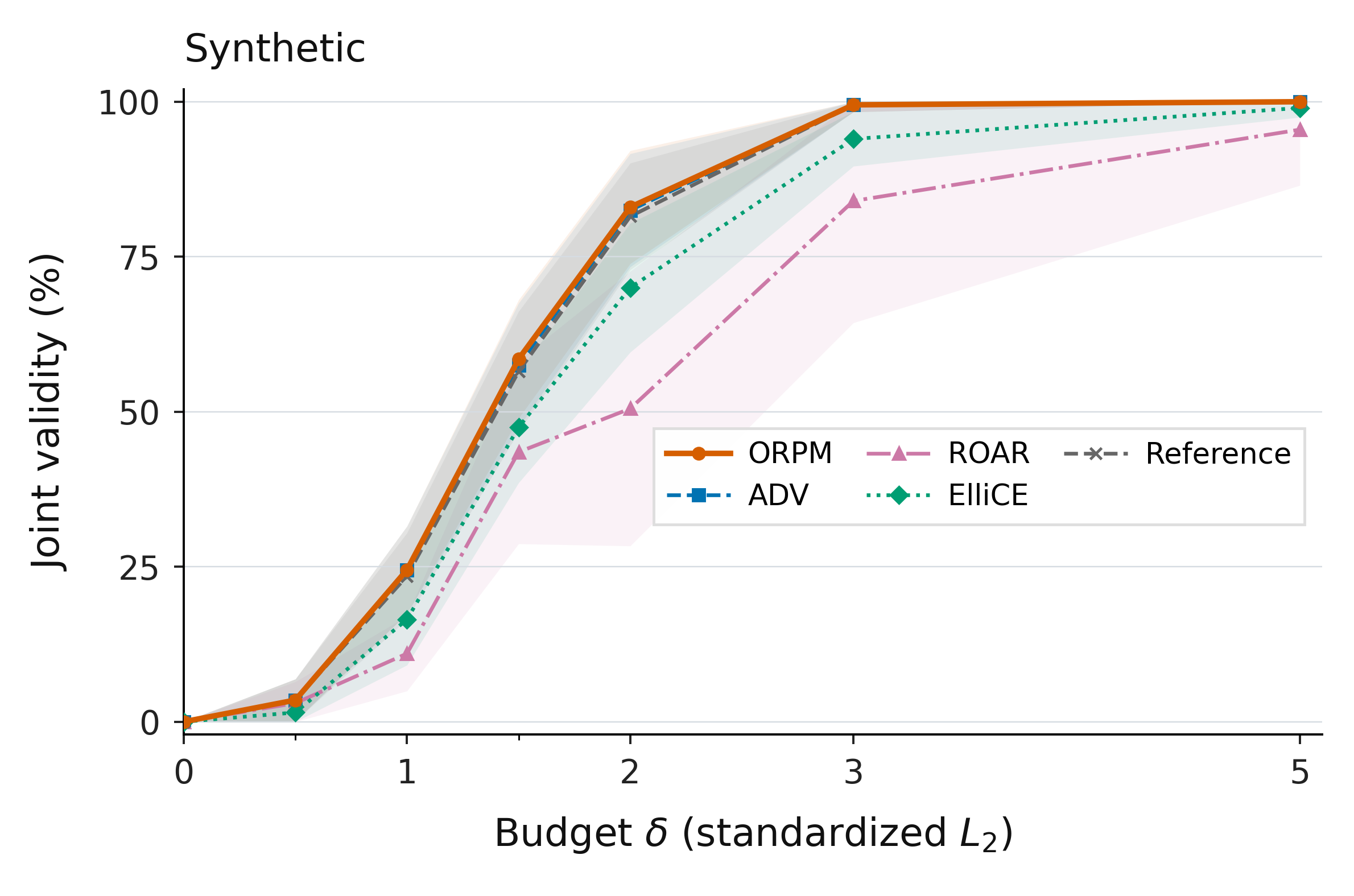}
\caption{Synthetic dataset, linear competing set with 12 models}
\end{subfigure}
\begin{subfigure}[t]{0.49\textwidth}
\centering\includegraphics[width=\linewidth]{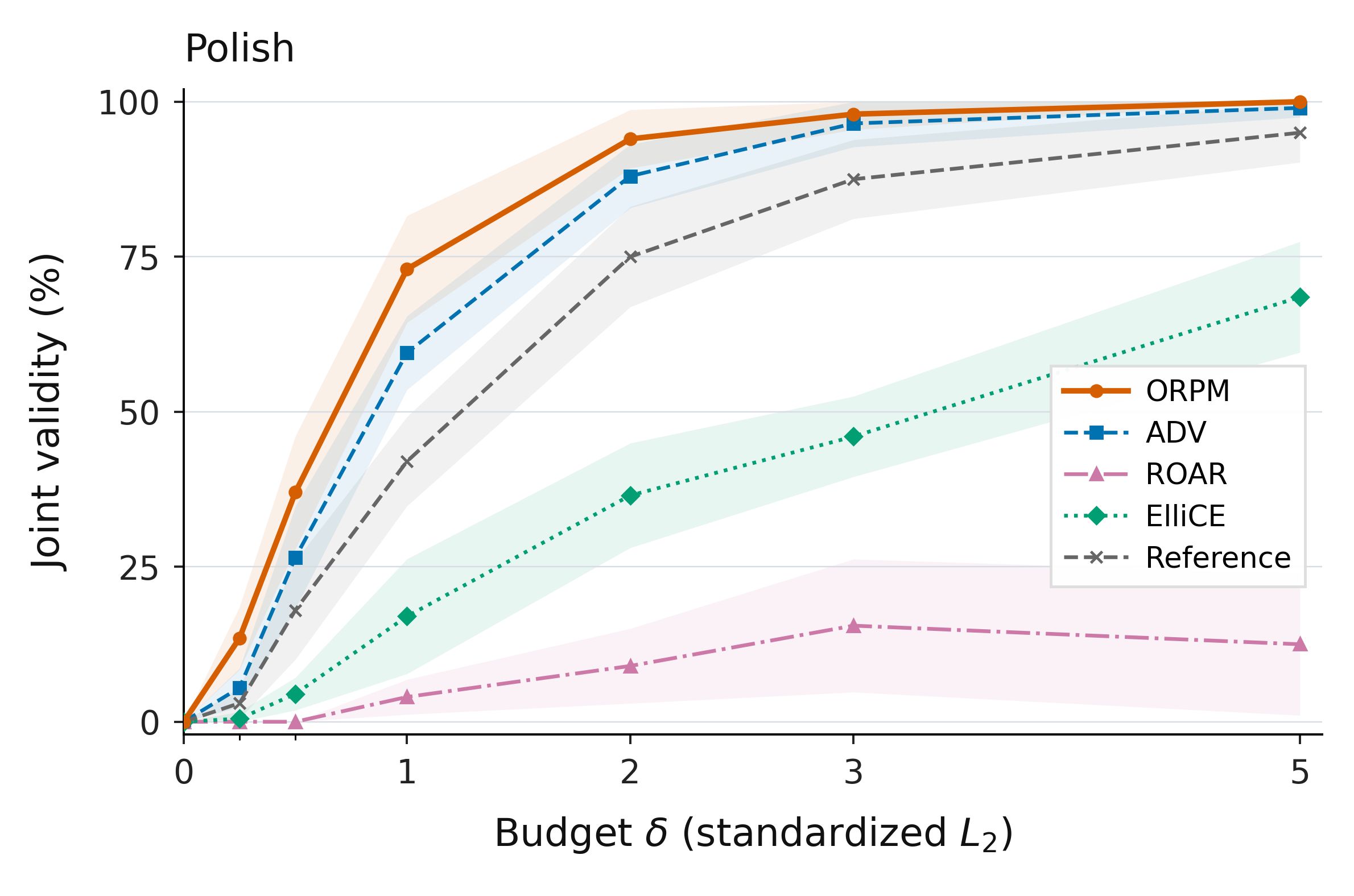}
\caption{Polish dataset, linear competing set with 6 models}
\end{subfigure}\hfill
\begin{subfigure}[t]{0.49\textwidth}
\centering\includegraphics[width=\linewidth]{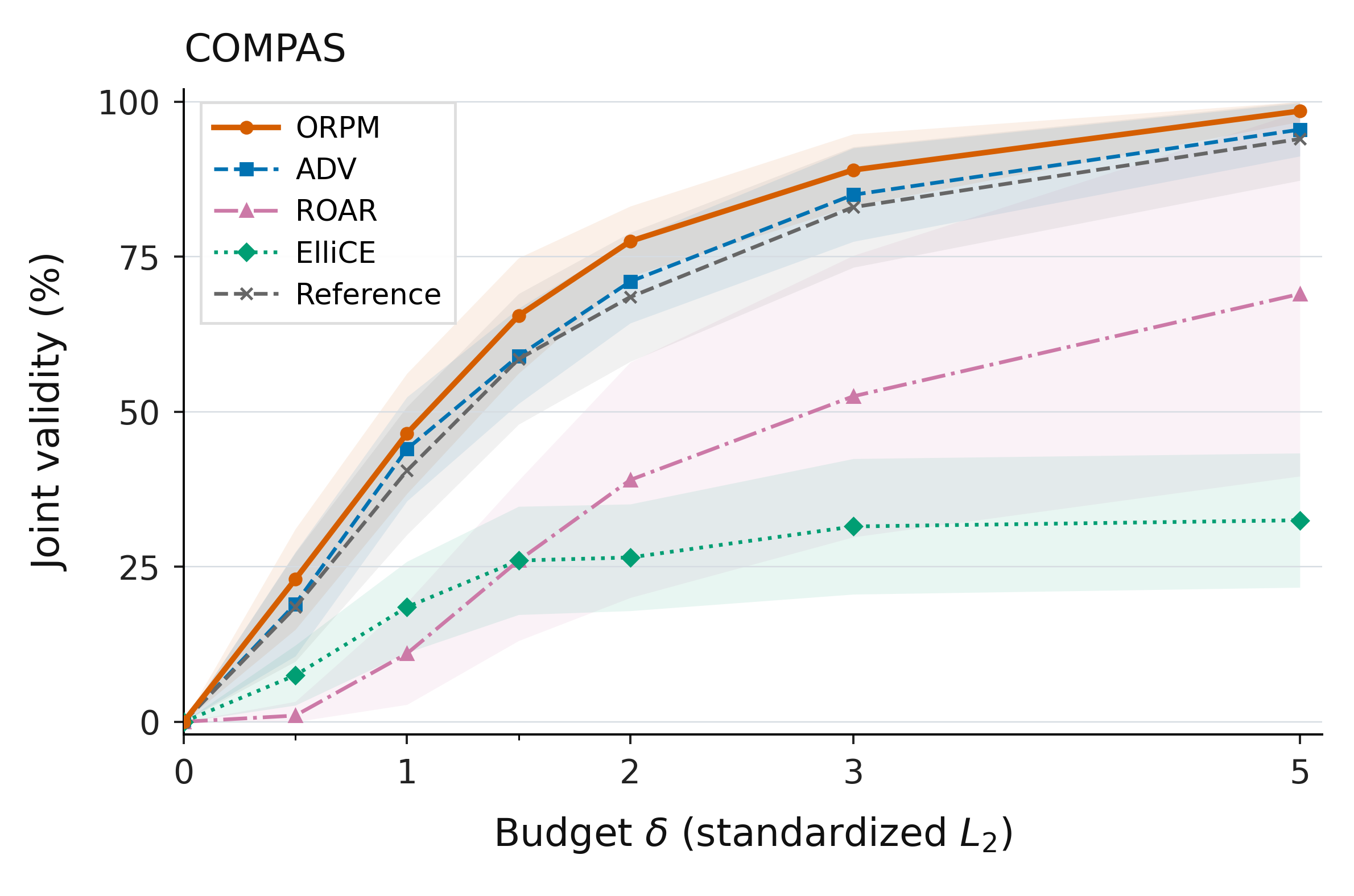}
\caption{COMPAS dataset, neural competing set with 6 models}
\end{subfigure}
\begin{subfigure}[t]{0.49\textwidth}
\centering\includegraphics[width=\linewidth]{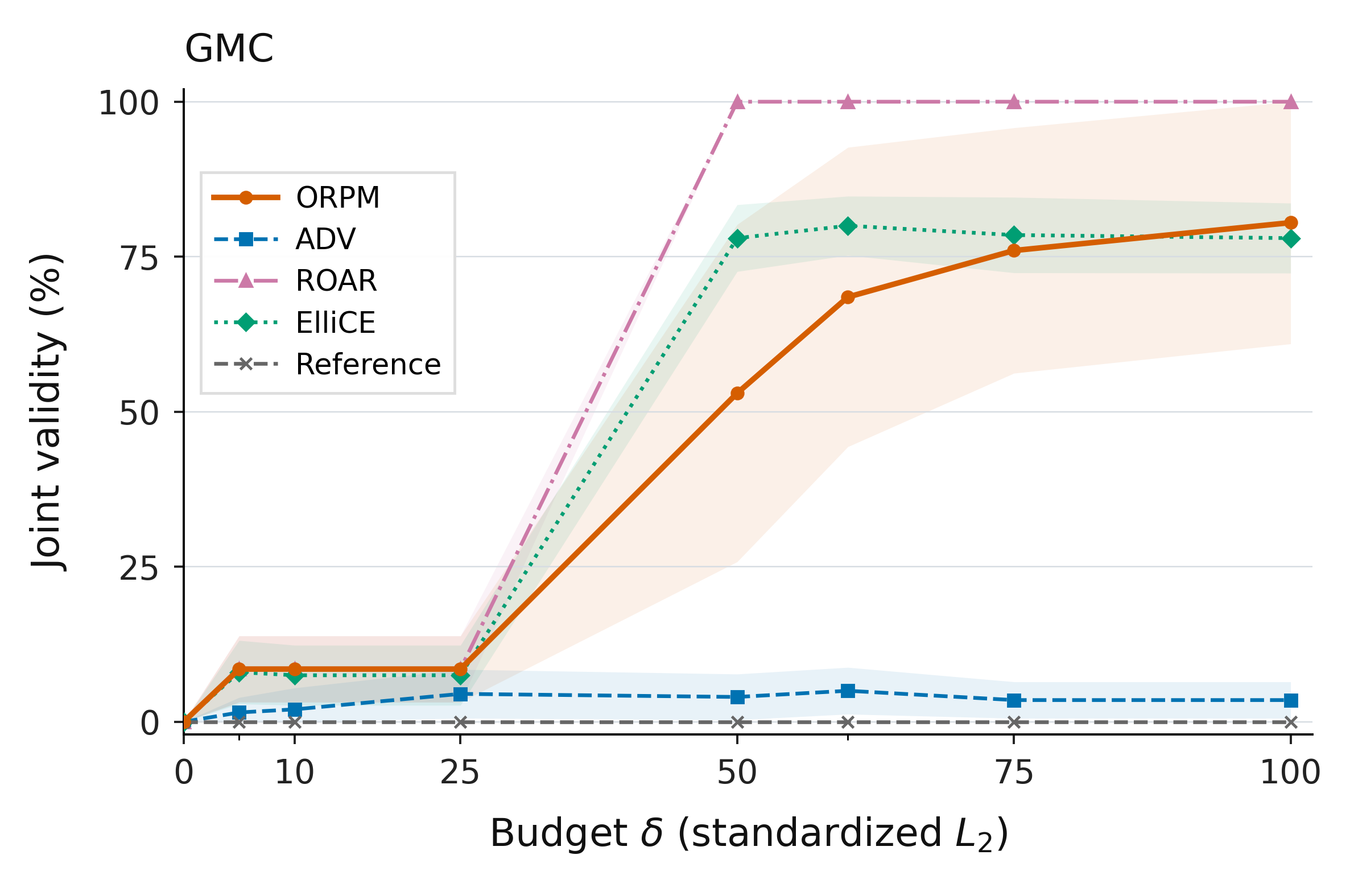}
\caption{GMC dataset, mixed competing set with 24 models}
\end{subfigure}\hfill
\begin{subfigure}[t]{0.49\textwidth}
\centering\includegraphics[width=\linewidth]{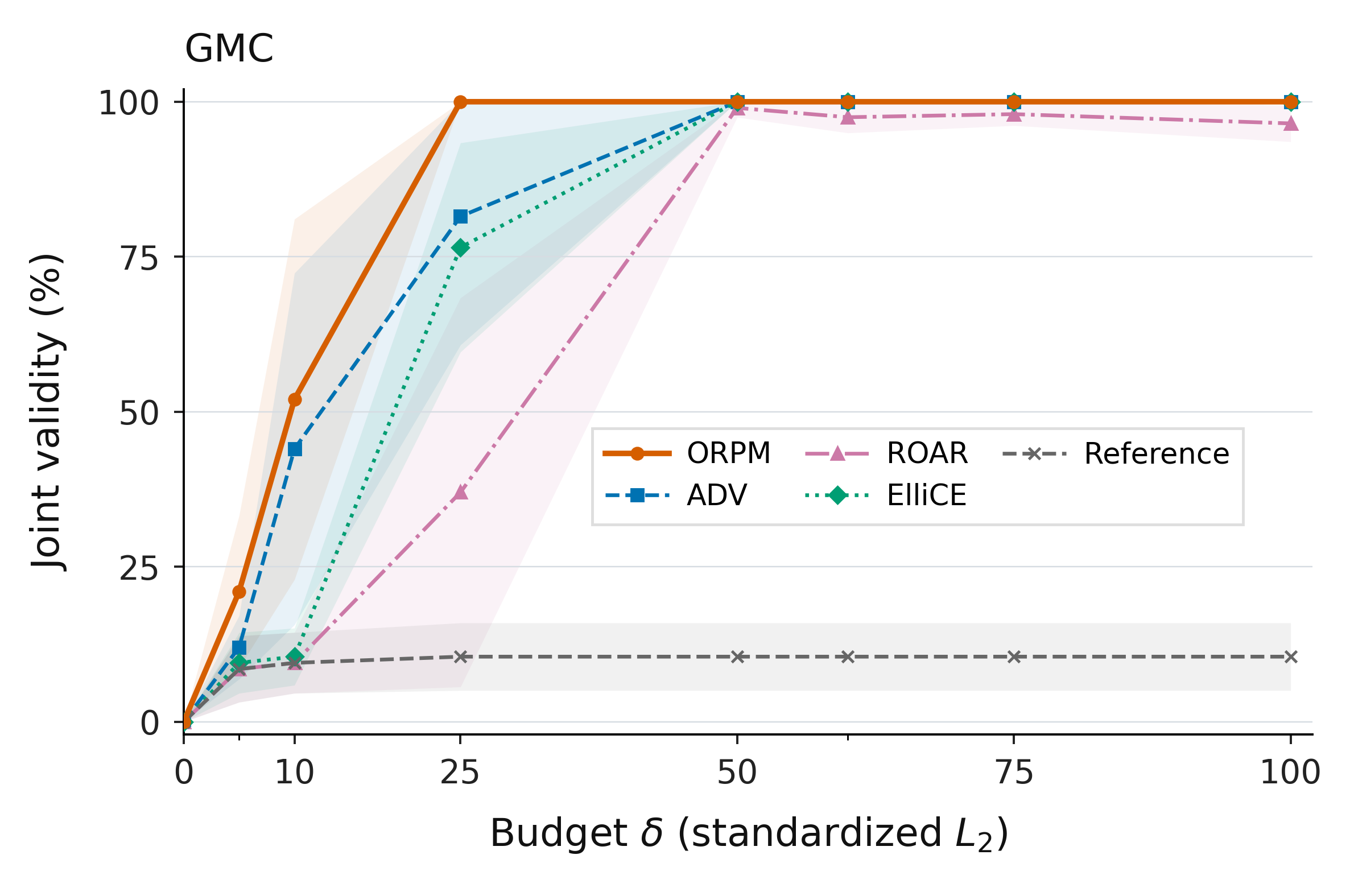}
\caption{GMC dataset, linear competing set with 24 models}
\end{subfigure}
\caption{The trade-off between validity and budget. Each subfigure corresponds to a dataset and competing set as described in the legend. Each curve corresponds to a different approach to provide recourse.\label{fig:trade-off-app}}
\end{figure}

\begin{figure}[ht!]
\centering
\begin{subfigure}[t]{0.49\textwidth}
\centering\includegraphics[width=\linewidth]{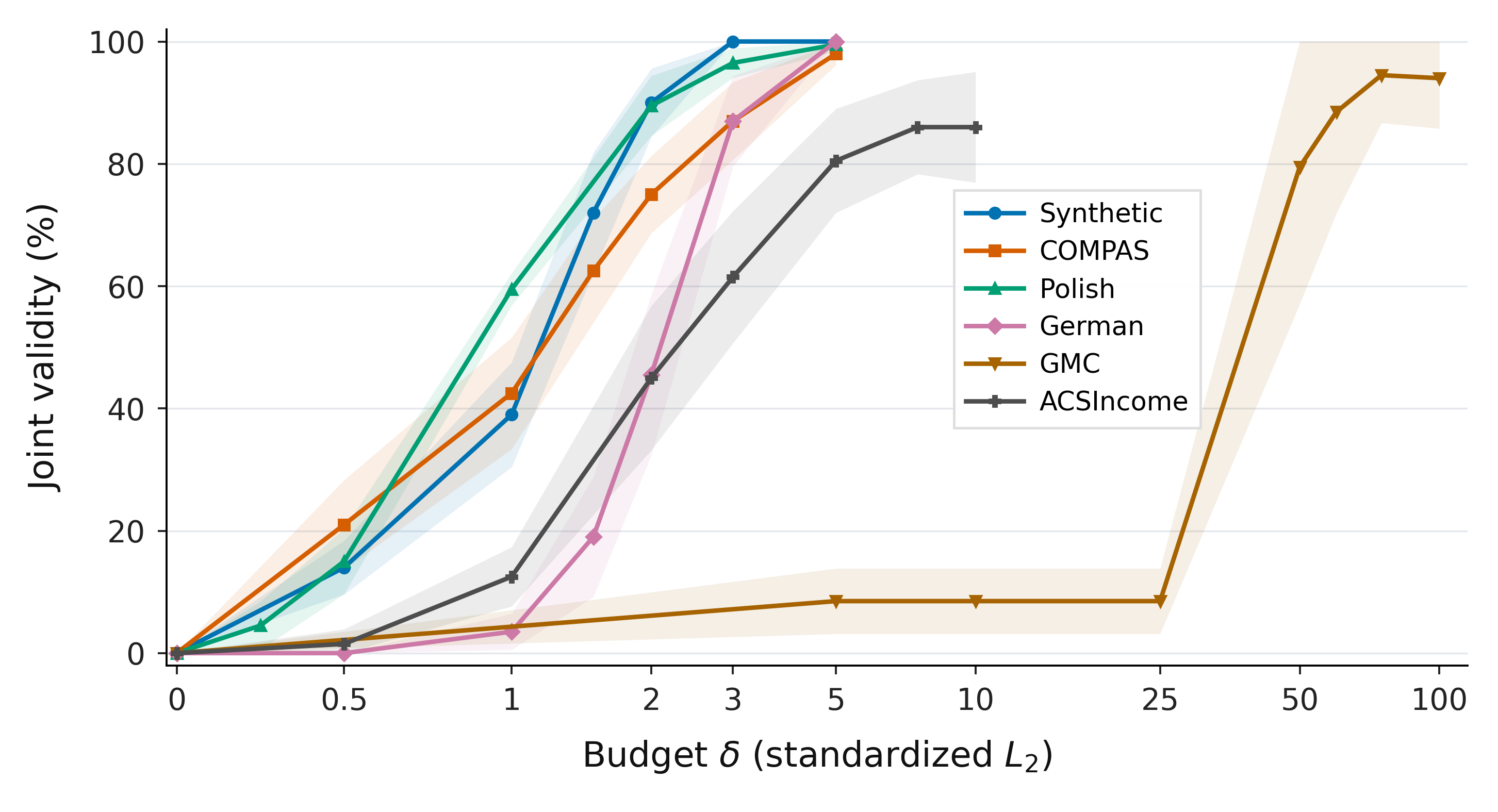}
\caption{The effect of budget on ORPM's validity for a neural competing set of 24 models.}
\end{subfigure}\hfill
\begin{subfigure}[t]{0.49\textwidth}
\centering\includegraphics[width=\linewidth]{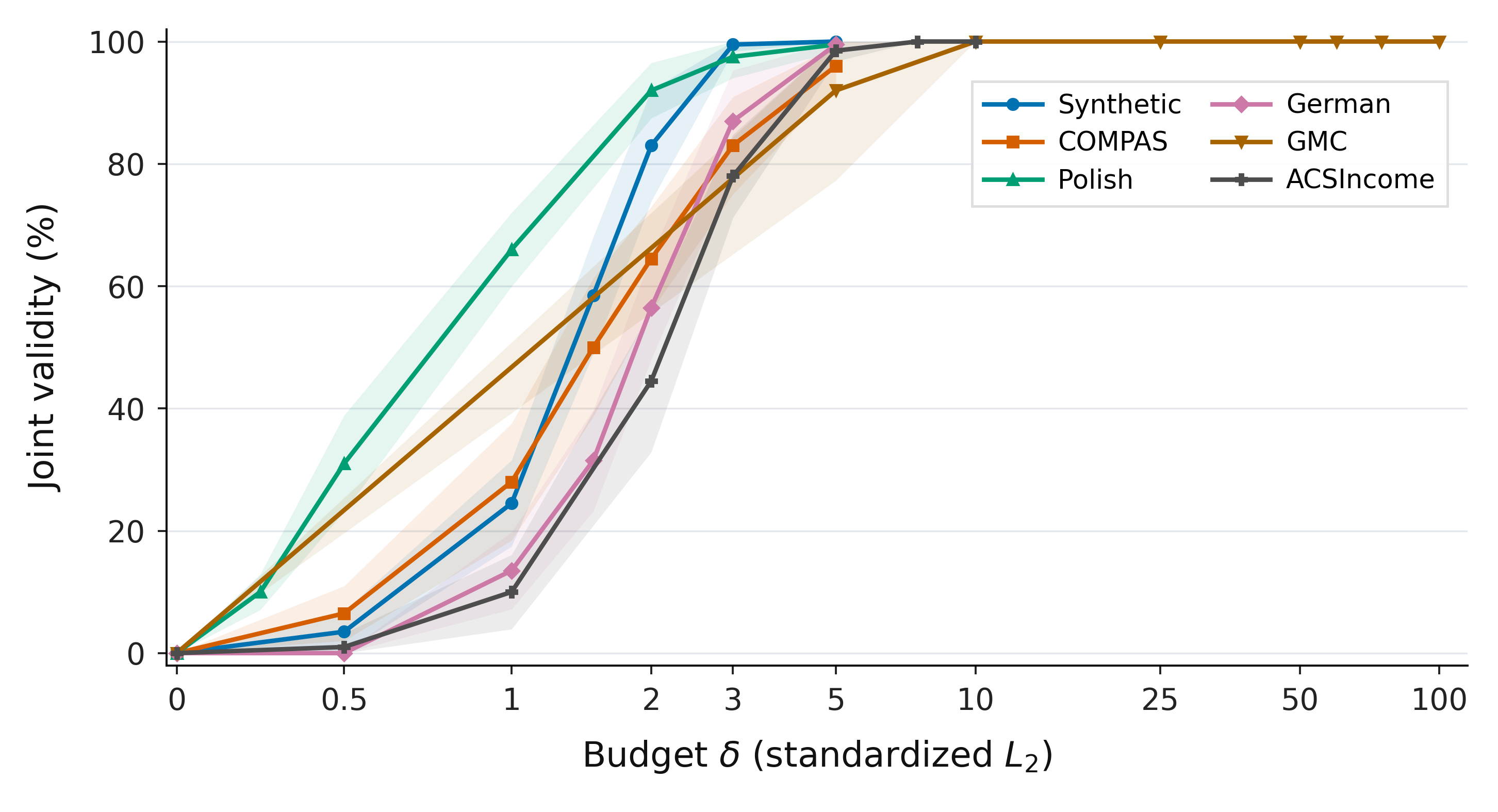}
\caption{The effect of budget on ORPM's validity for a linear competing set of 12 models.}
\end{subfigure}
\caption{The effect of budget (X-axis) on the validity of the recourse returned by ORPM (Y-axis). Each curve corresponds to a different dataset.\label{fig:parameters-app-budget}}
\end{figure}

\begin{figure}[ht!]
\centering
\begin{subfigure}[t]{0.49\textwidth}
\centering\includegraphics[width=\linewidth]{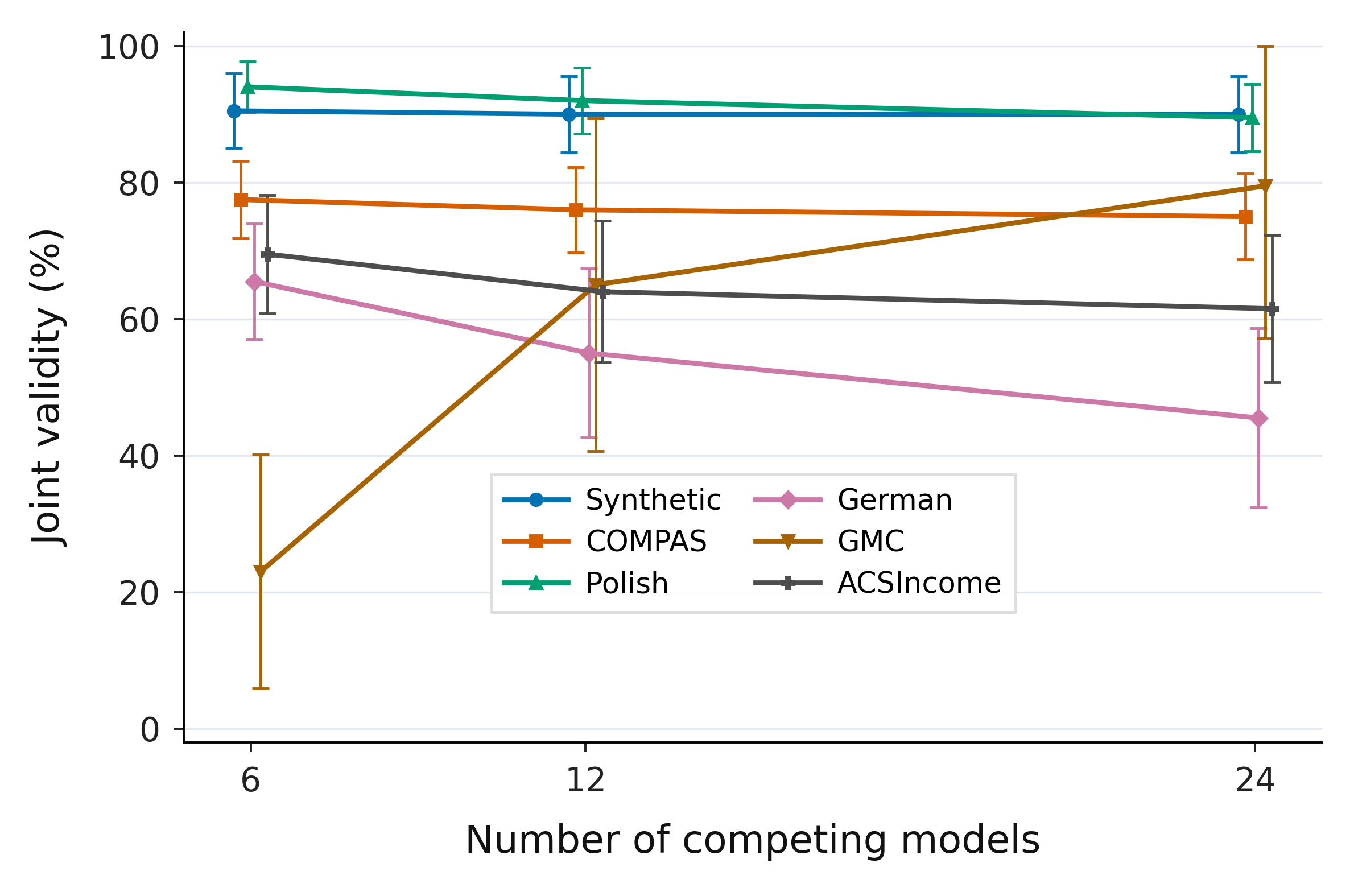}
\caption{The effect of the number of models in a neural competing set on ORPM's validity.}
\end{subfigure}\hfill
\begin{subfigure}[t]{0.49\textwidth}
\centering\includegraphics[width=\linewidth]{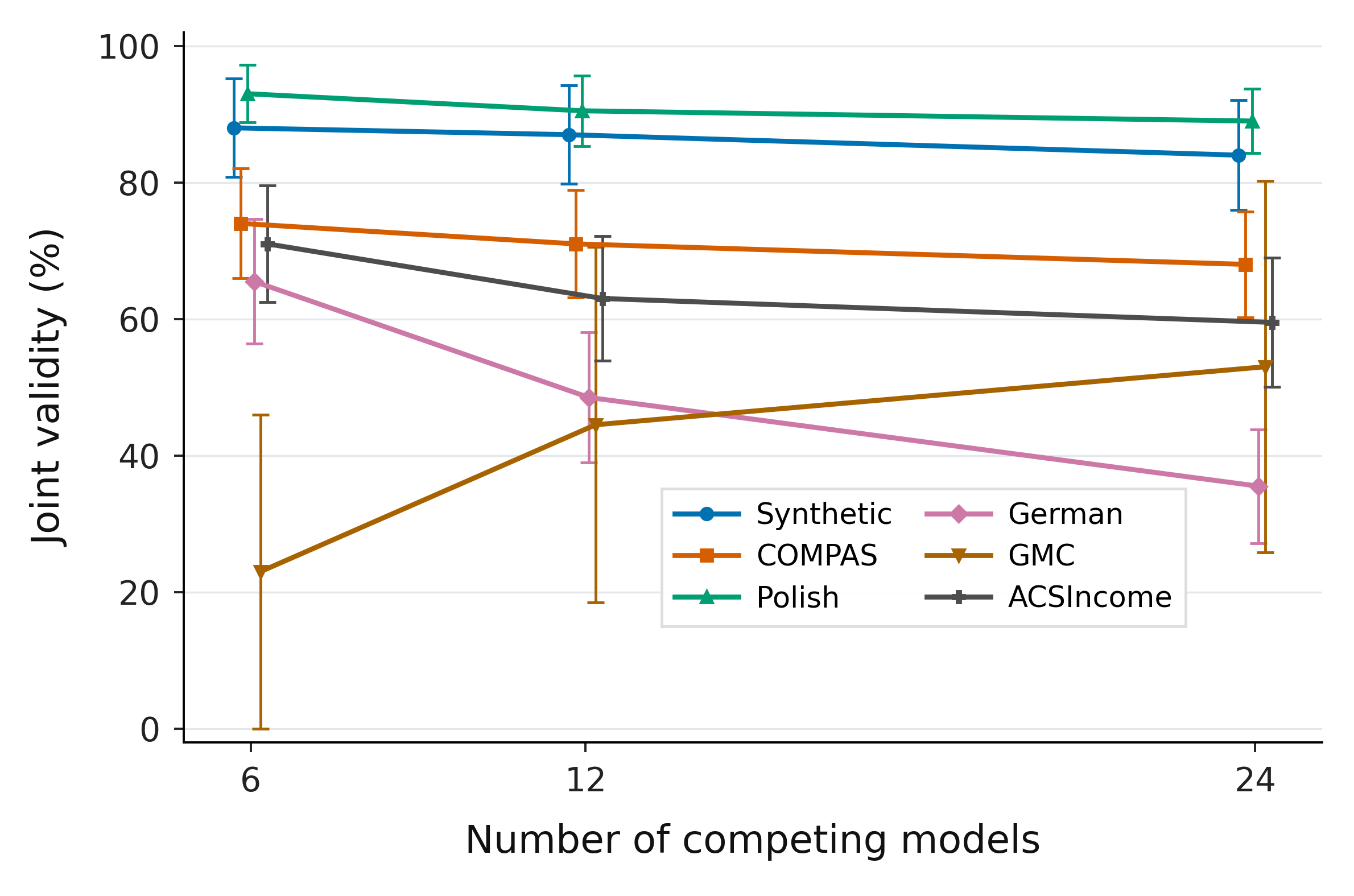}
\caption{The effect of the number of models in a mixed competing set on ORPM's validity.}
\end{subfigure}
\begin{subfigure}[t]{0.49\textwidth}
\centering\includegraphics[width=\linewidth]{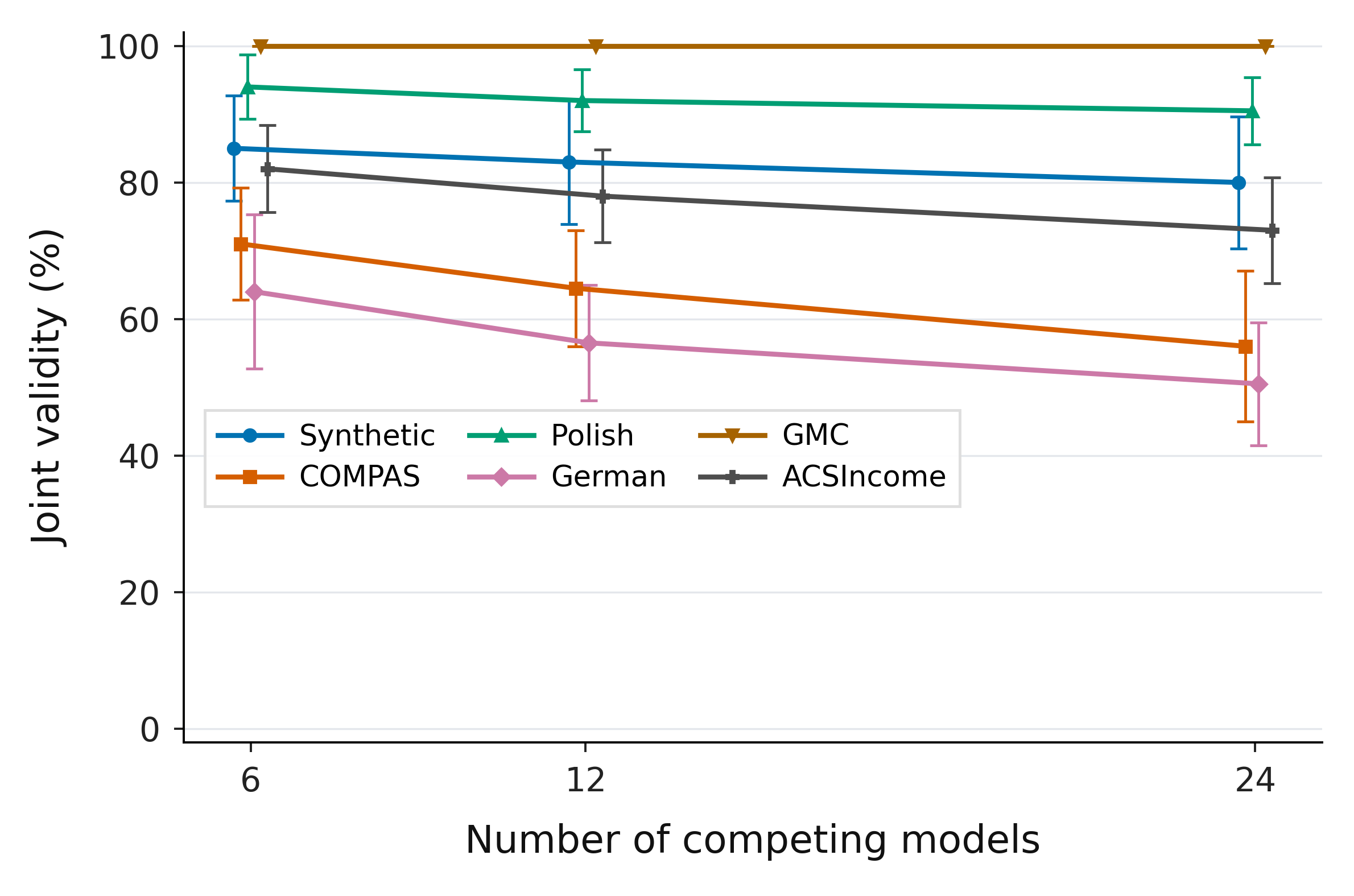}
\caption{The effect of the number of models in a linear competing set on ORPM's validity.}
\end{subfigure}\hfill
\begin{subfigure}[t]{0.49\textwidth}
\centering\includegraphics[width=\linewidth]{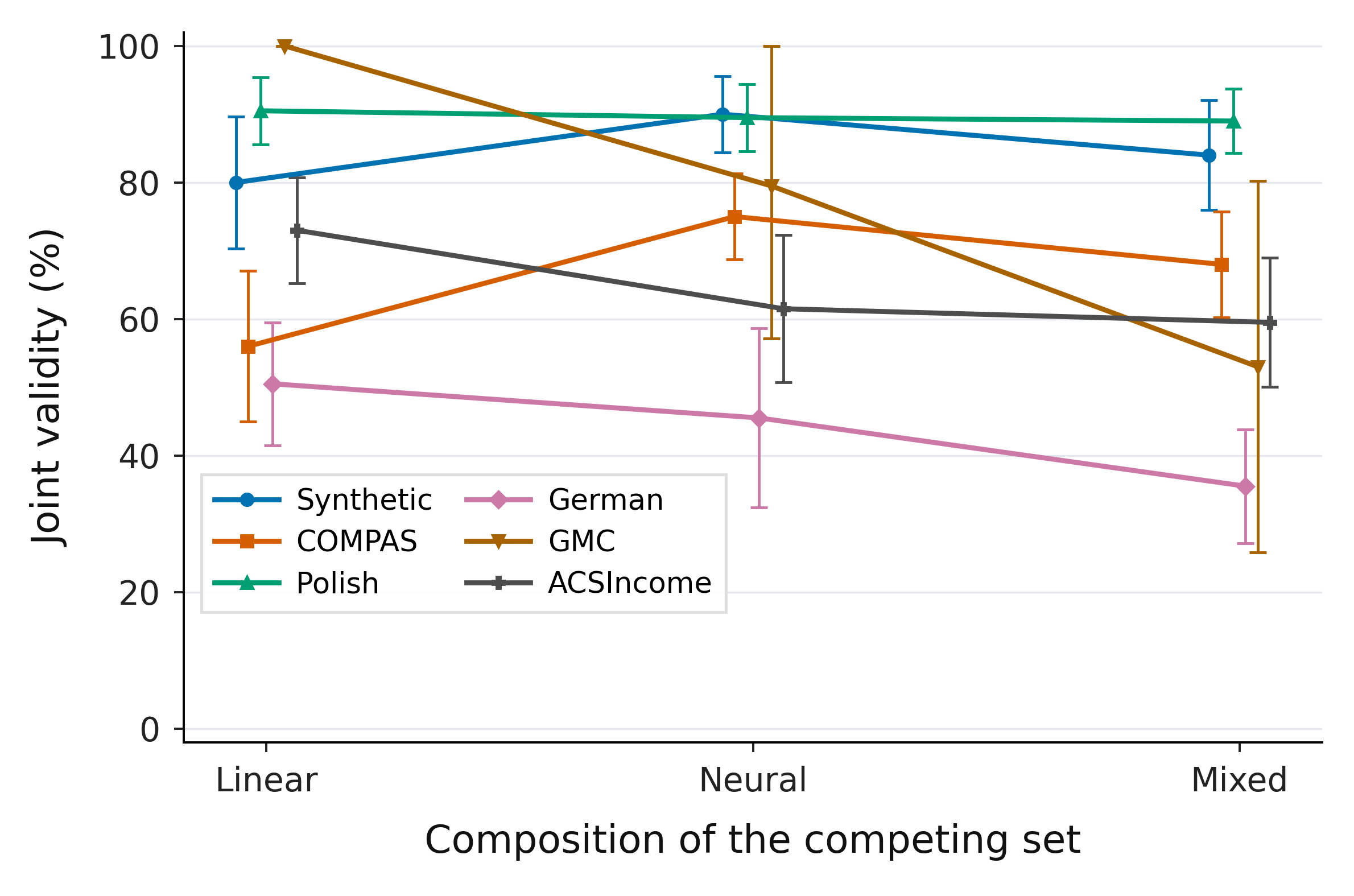}
\caption{The effect of composition of the competing set of 24 models on ORPM's validity.}
\end{subfigure}
\caption{The effect of composition or size of the competing set (X-axis) on validity of the recourse returned by ORPM (Y-axis). Each curve corresponds to a different dataset.\label{fig:parameters-app-size}}
\end{figure}

For Synthetic, COMPAS,  and German datasets, we used a budget of $\{0.5, 1, 2, 3, 5\}$. For the Polish dataset, we used a budget of $\{0.25, 0.5, 1, 2, 3, 5\}$. For GMC, we used a budget of $\{5, 10, 25, 50, 60, 75, 100\}$. The budget used for ACSIncome is $\{0.5, 1, 2, 3, 5, 7.5, 10\}$. Additional trade-off results are provided in Figure~\ref{fig:trade-off-app}.  Additional results on how budget and the size or complexity of the competing set affect the validity are provided in Figures~\ref{fig:parameters-app-budget}~and~\ref{fig:parameters-app-size}. 

\subsubsection{The Cost of Multiplicity}
\label{sec:app-exp-cost-of-multiplicity}
\begin{figure}[ht!]
\centering
\begin{subfigure}[t]{0.49\textwidth}
\centering\includegraphics[width=\linewidth]{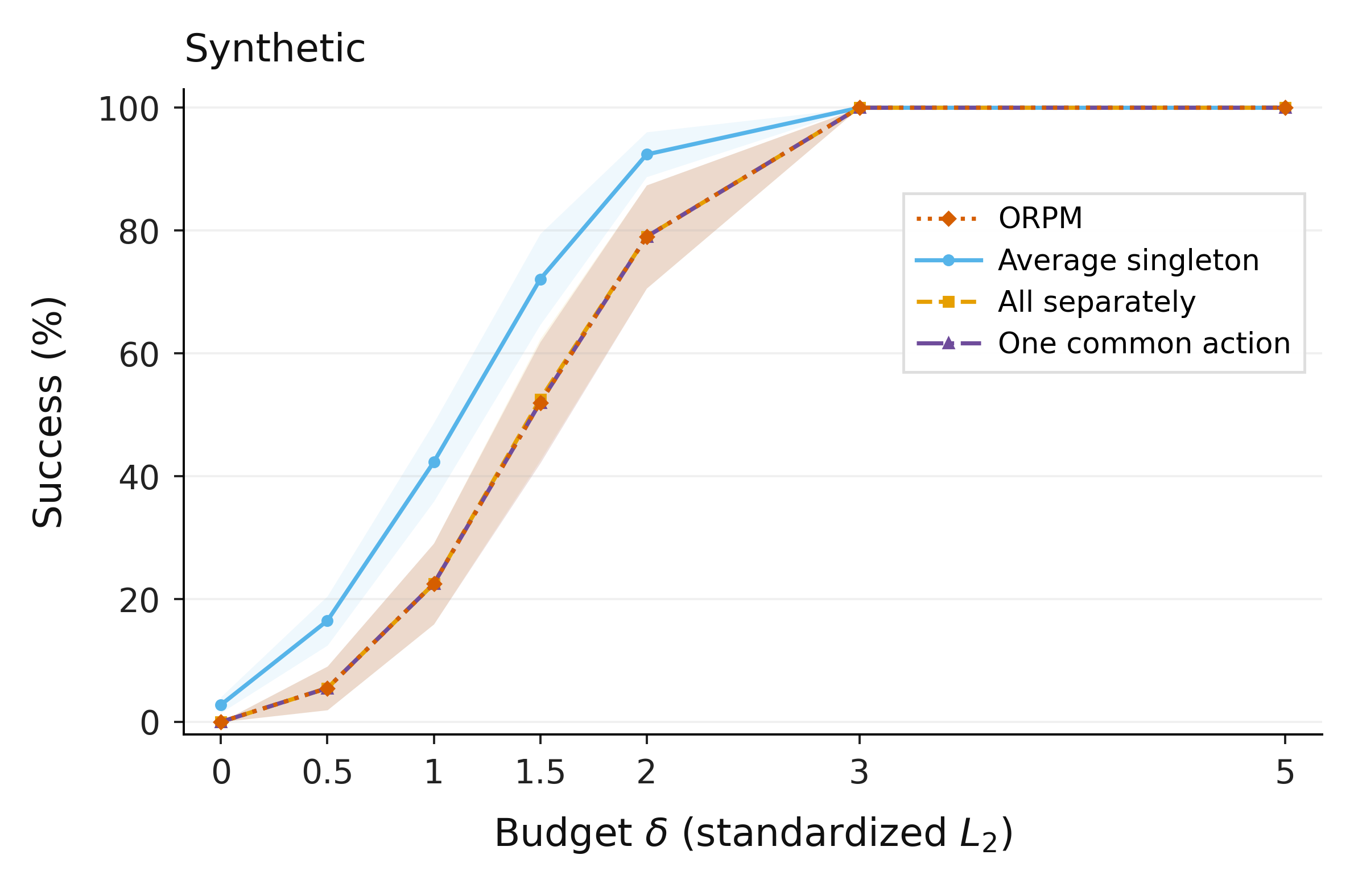}
\caption{Synthetic dataset, linear competing set of 24 models.}
\end{subfigure}\hfill
\begin{subfigure}[t]{0.49\textwidth}
\centering\includegraphics[width=\linewidth]{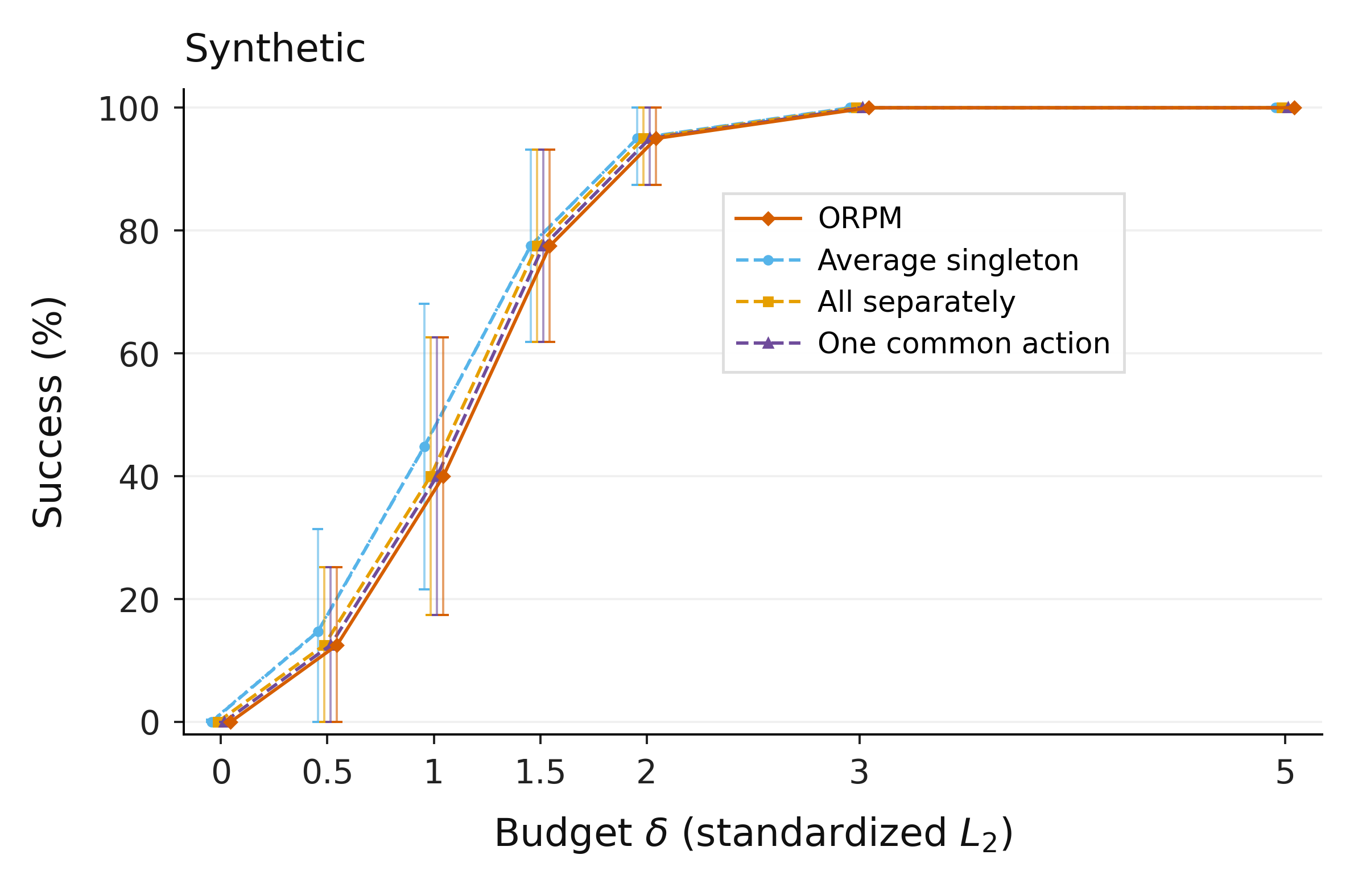}
\caption{Synthetic dataset, neural competing set of 24 models.}
\end{subfigure}
\begin{subfigure}[t]{0.49\textwidth}
\centering\includegraphics[width=\linewidth]{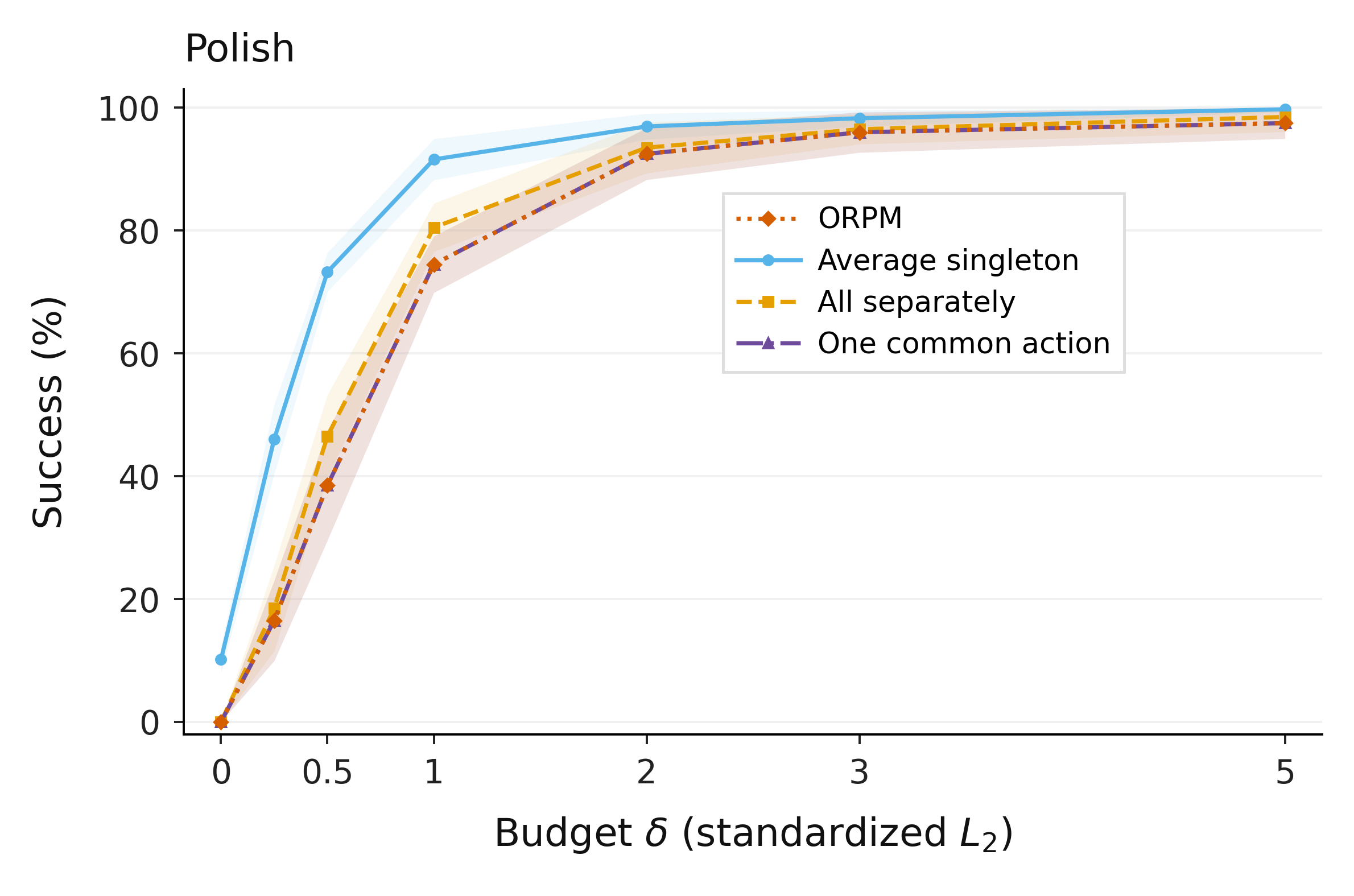}
\caption{Polish dataset, linear competing set of 12 models.}
\end{subfigure}\hfill
\begin{subfigure}[t]{0.49\textwidth}
\centering\includegraphics[width=\linewidth]{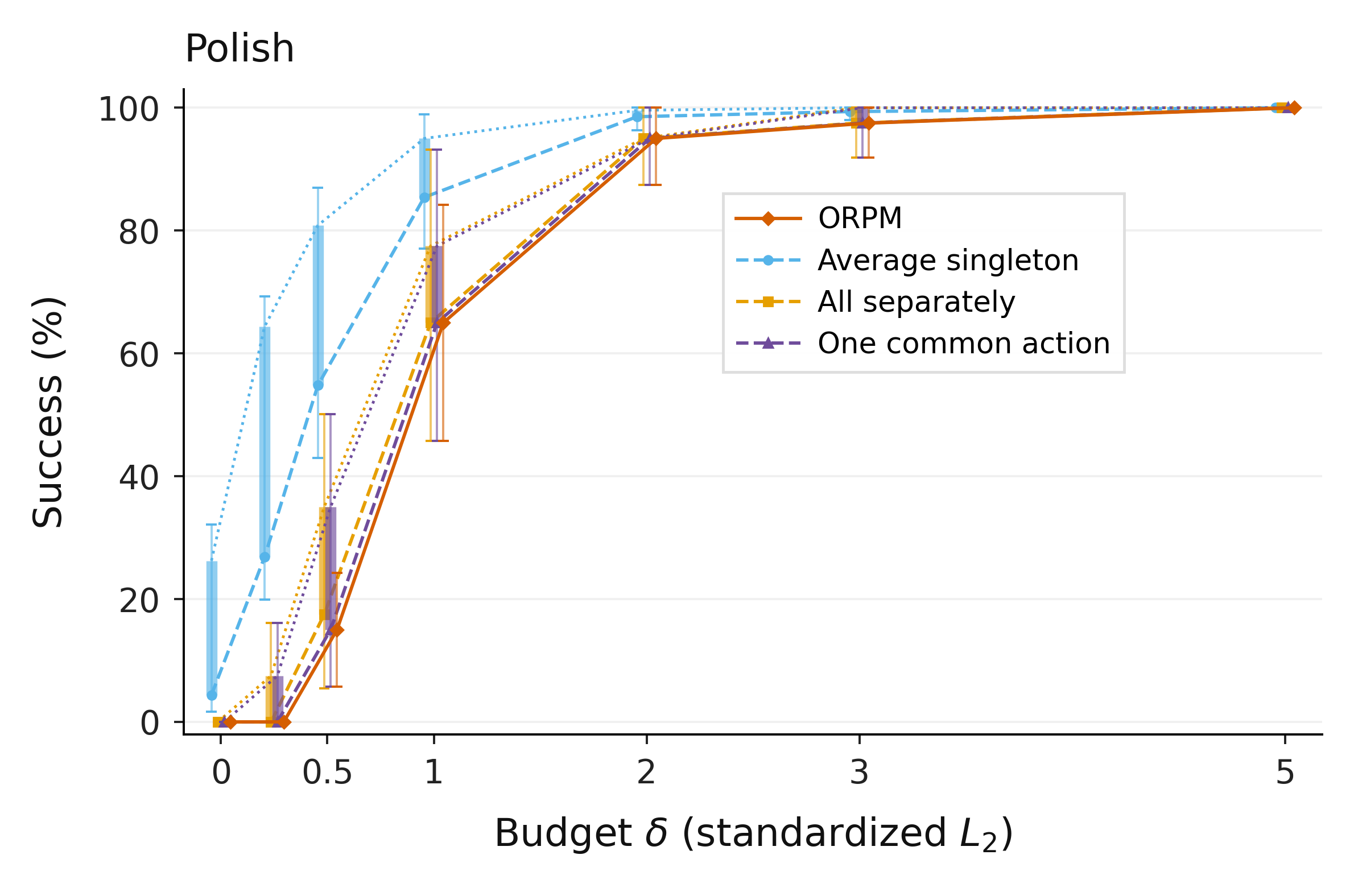}
\caption{Polish dataset, neural competing set of 12 models.}
\end{subfigure}
\begin{subfigure}[t]{0.49\textwidth}
\centering\includegraphics[width=\linewidth]{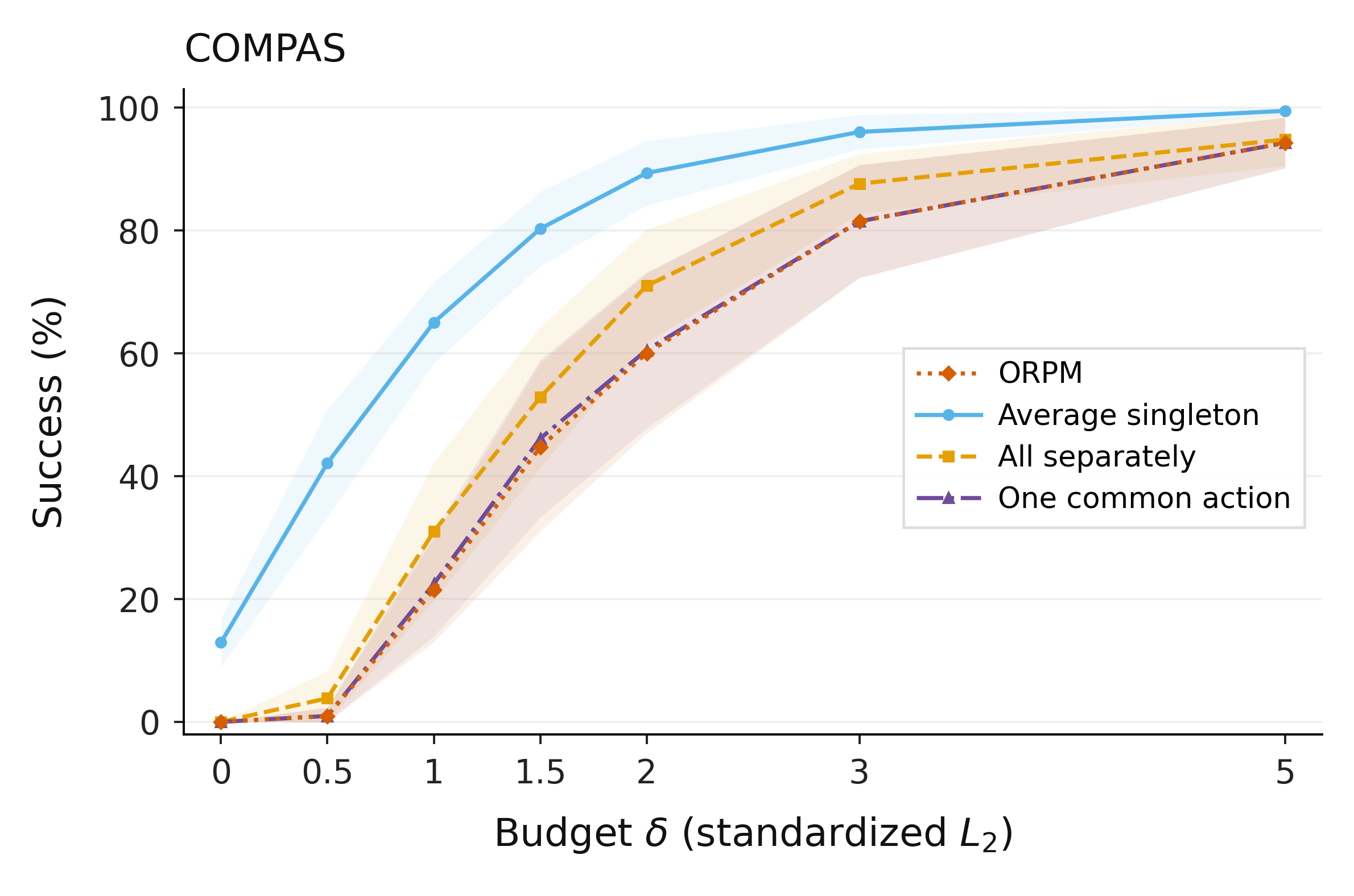}
\caption{COMPAS dataset, linear competing set of 24 models.}
\end{subfigure}\hfill
\begin{subfigure}[t]{0.49\textwidth}
\centering\includegraphics[width=\linewidth]{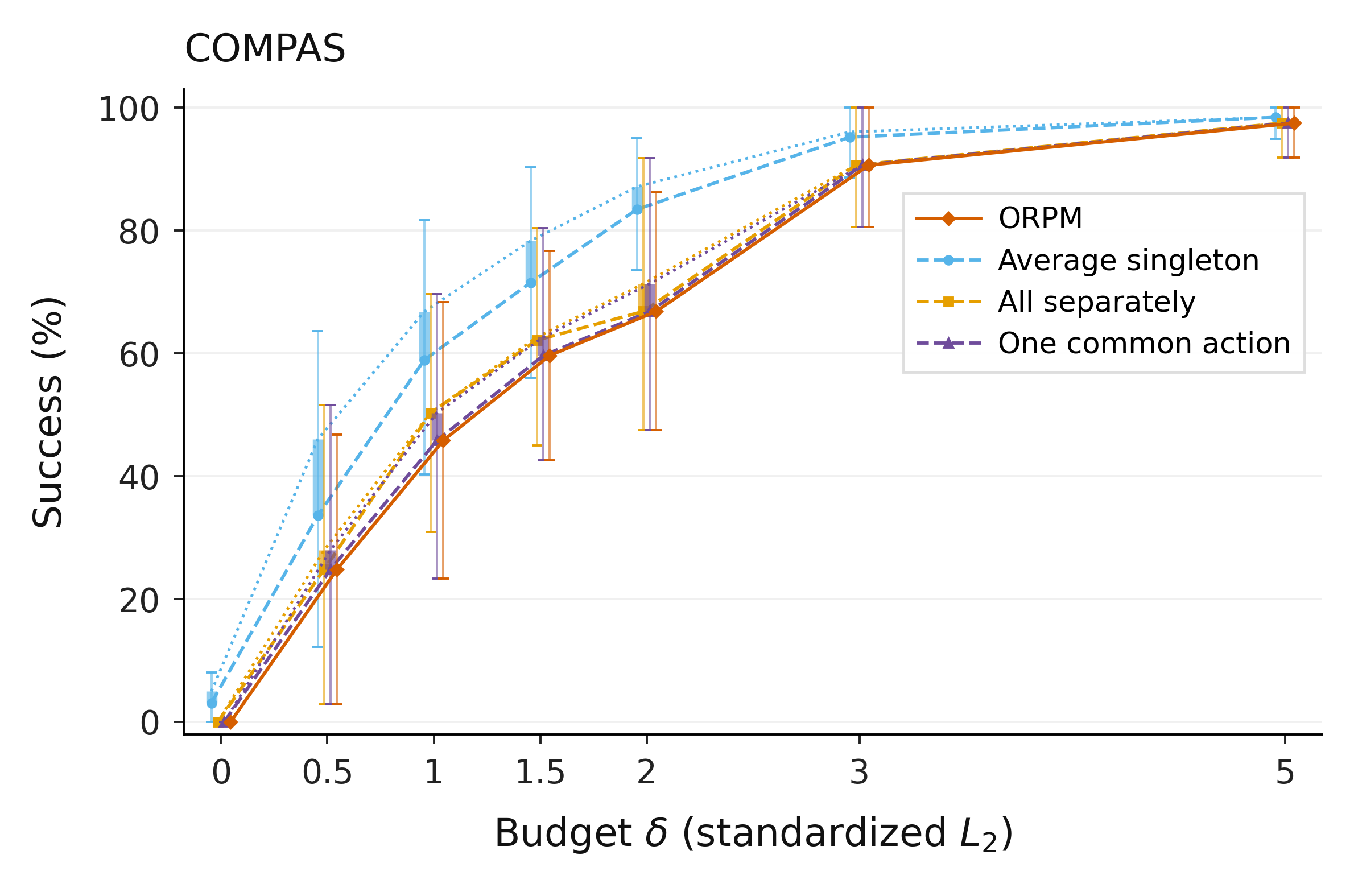}
\caption{COMPAS dataset, neural competing set of 24 models.}
\end{subfigure}
\caption{The validity degradation to guard against multiplicity. Each subfigure corresponds to a pair of datasets and competing sets as indicated in the legend. Each curve tracks the success rate (Y-axis) for ORPM and baselines as a function of allowable budget (X-axis).\label{fig:cost-app-1}}
\end{figure}

\begin{figure}[htbp]
\centering
\begin{subfigure}[t]{0.49\textwidth}
\centering\includegraphics[width=\linewidth]{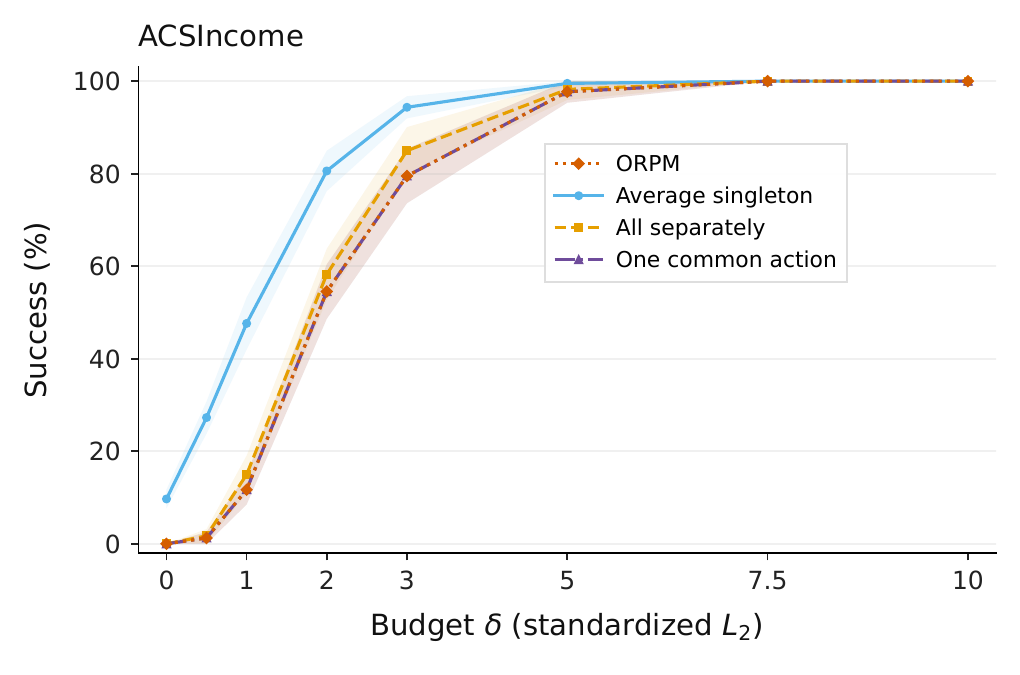}
\caption{ACSIncome dataset, linear competing set of 12 models.}
\end{subfigure}\hfill
\begin{subfigure}[t]{0.49\textwidth}
\centering\includegraphics[width=\linewidth]{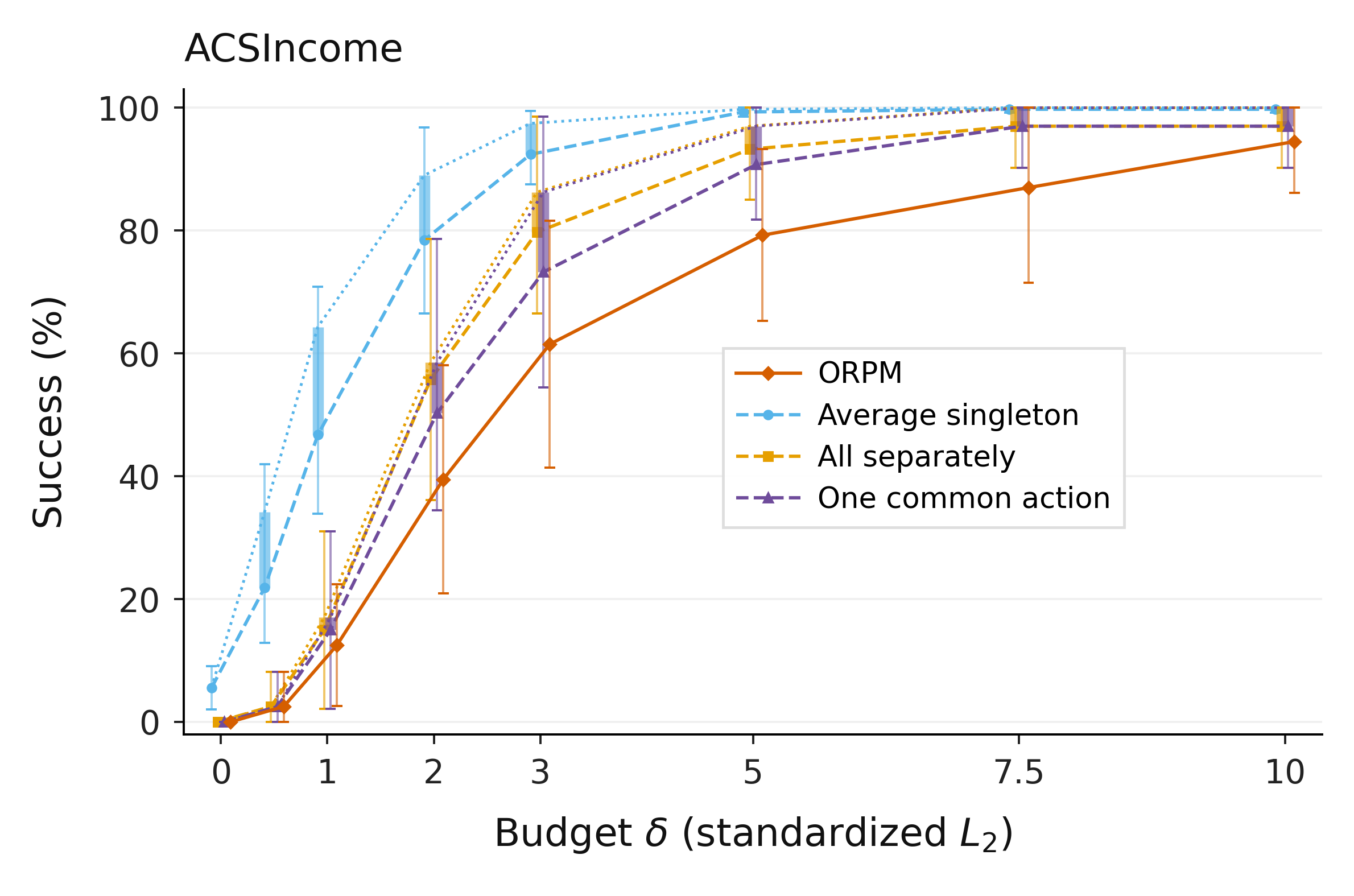}
\caption{ACSIncome dataset, neural competing set of 12 models.}
\end{subfigure}
\begin{subfigure}[t]{0.49\textwidth}
\centering\includegraphics[width=\linewidth]{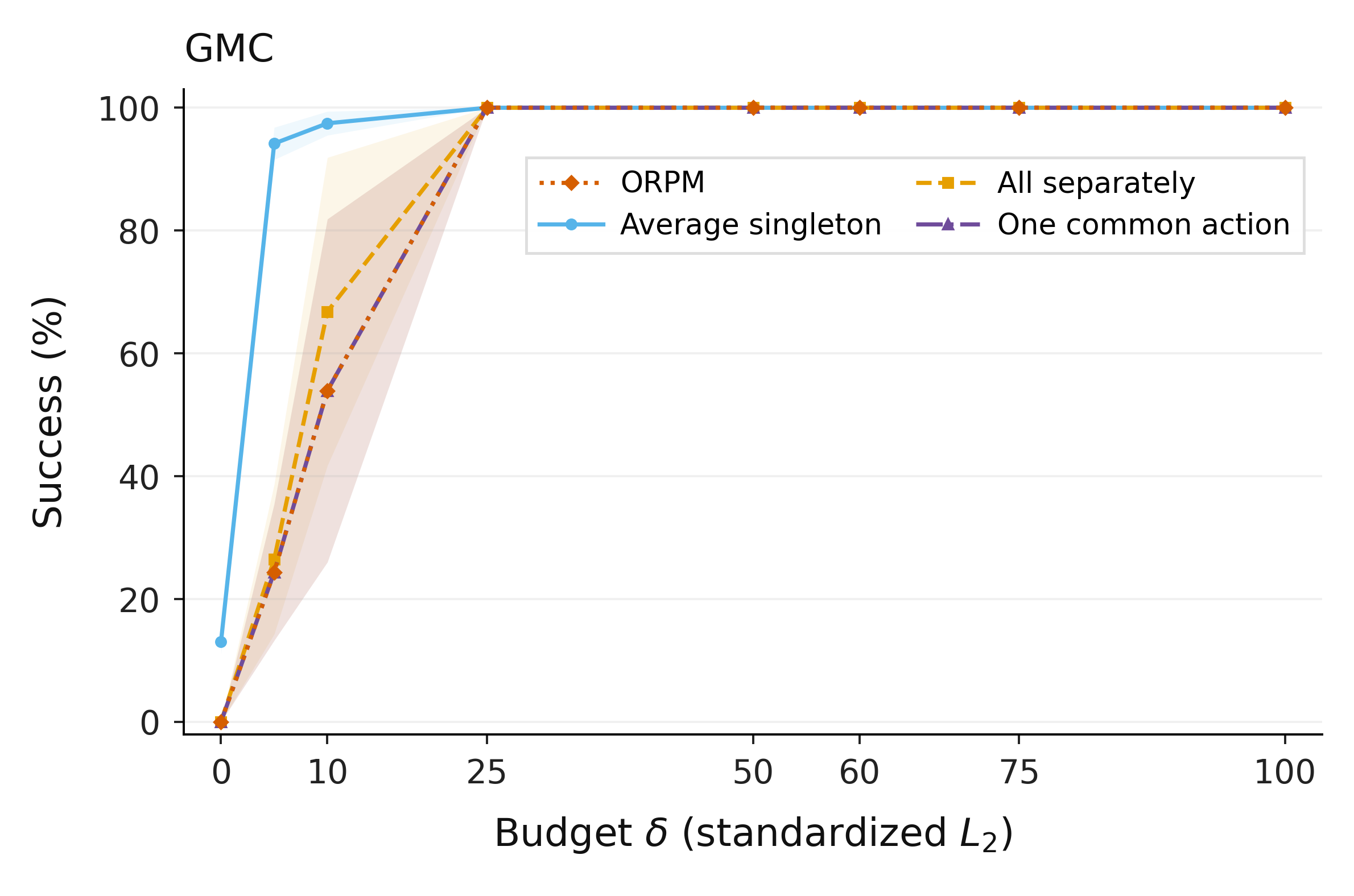}
\caption{GMC dataset, linear competing set of 24 models. \label{fig:cost-main-GMC-linear}}
\end{subfigure}\hfill
\begin{subfigure}[t]{0.49\textwidth}
\centering\includegraphics[width=\linewidth]{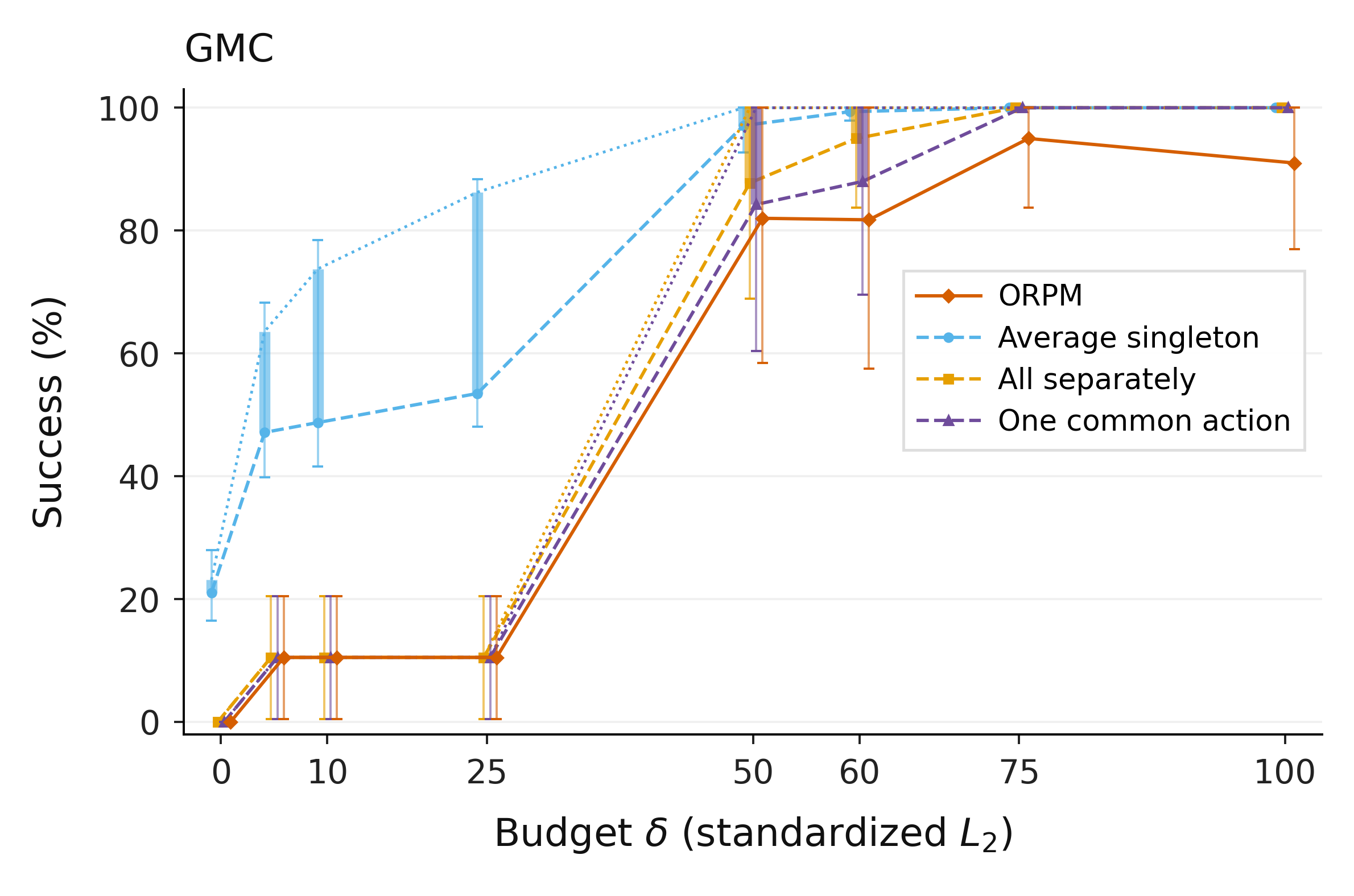}
\caption{GMC dataset, neural competing set of 24 models. \label{fig:cost-main-GMC-neural}}
\end{subfigure}
\caption{The validity degradation to guard against multiplicity. Each subfigure corresponds to a pair of datasets and competing sets as indicated in the legend. Each curve tracks the success rate (Y-axis) for ORPM and baselines as a function of allowable budget (X-axis).\label{fig:cost-app-2}}
\end{figure}

Additional cost of multiplicity results for the remaining datasets are provided in Figures~\ref{fig:cost-app-1}~and~\ref{fig:cost-app-2}. These are the analogous results to Figure~\ref{fig:cost-main} in the main body, which focused on the German dataset.

\subsubsection{The Disparity of Multiplicity}
\label{sec:app-exp-disparity-of-multiplicity}
\begin{figure}[ht!]
\centering
\begin{subfigure}[t]{0.49\textwidth}
\centering\includegraphics[width=\linewidth]{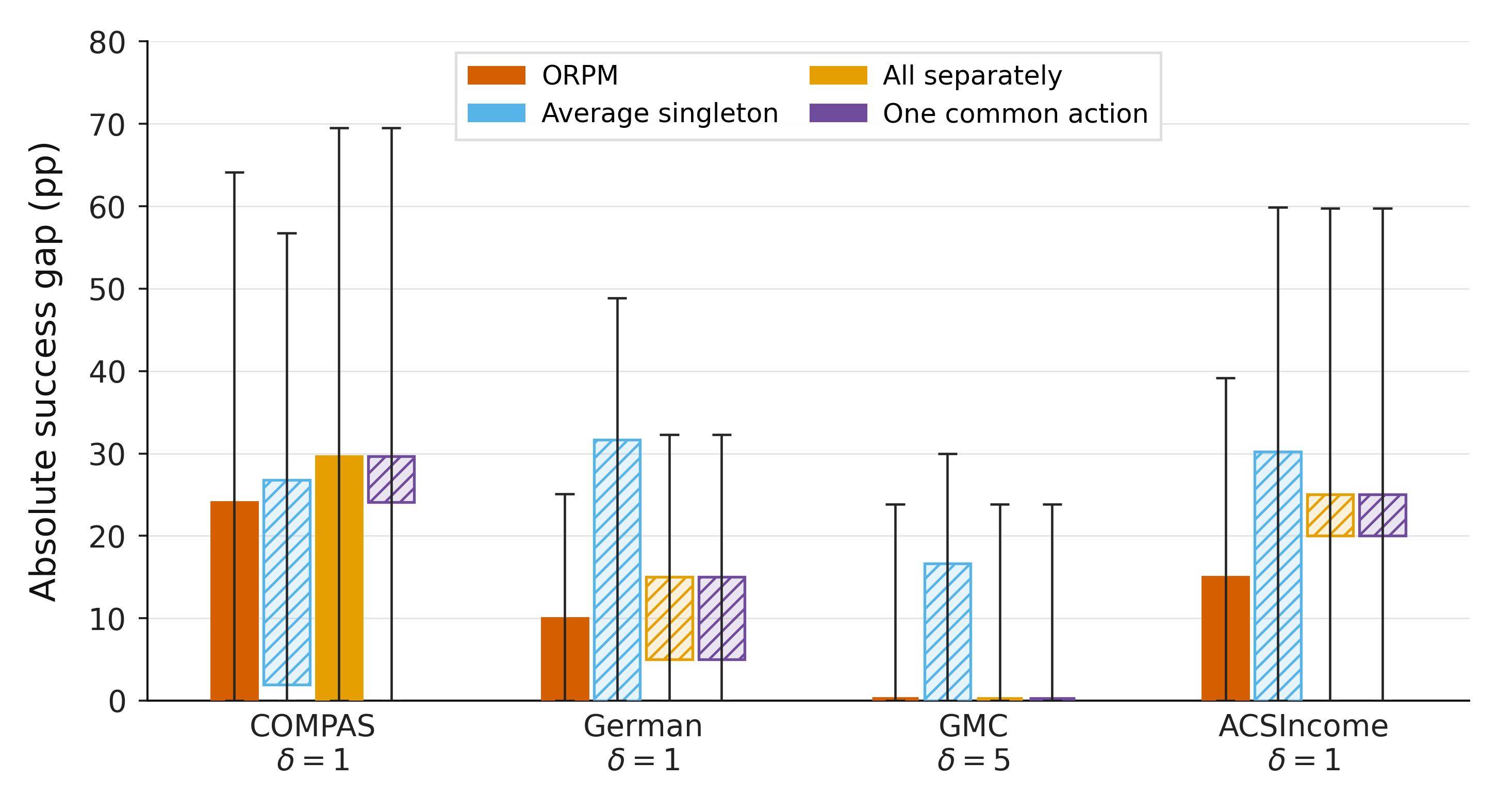}
\caption{Neural competing set, small budget regime.}
\end{subfigure}\hfill
\begin{subfigure}[t]{0.49\textwidth}
\centering\includegraphics[width=\linewidth]{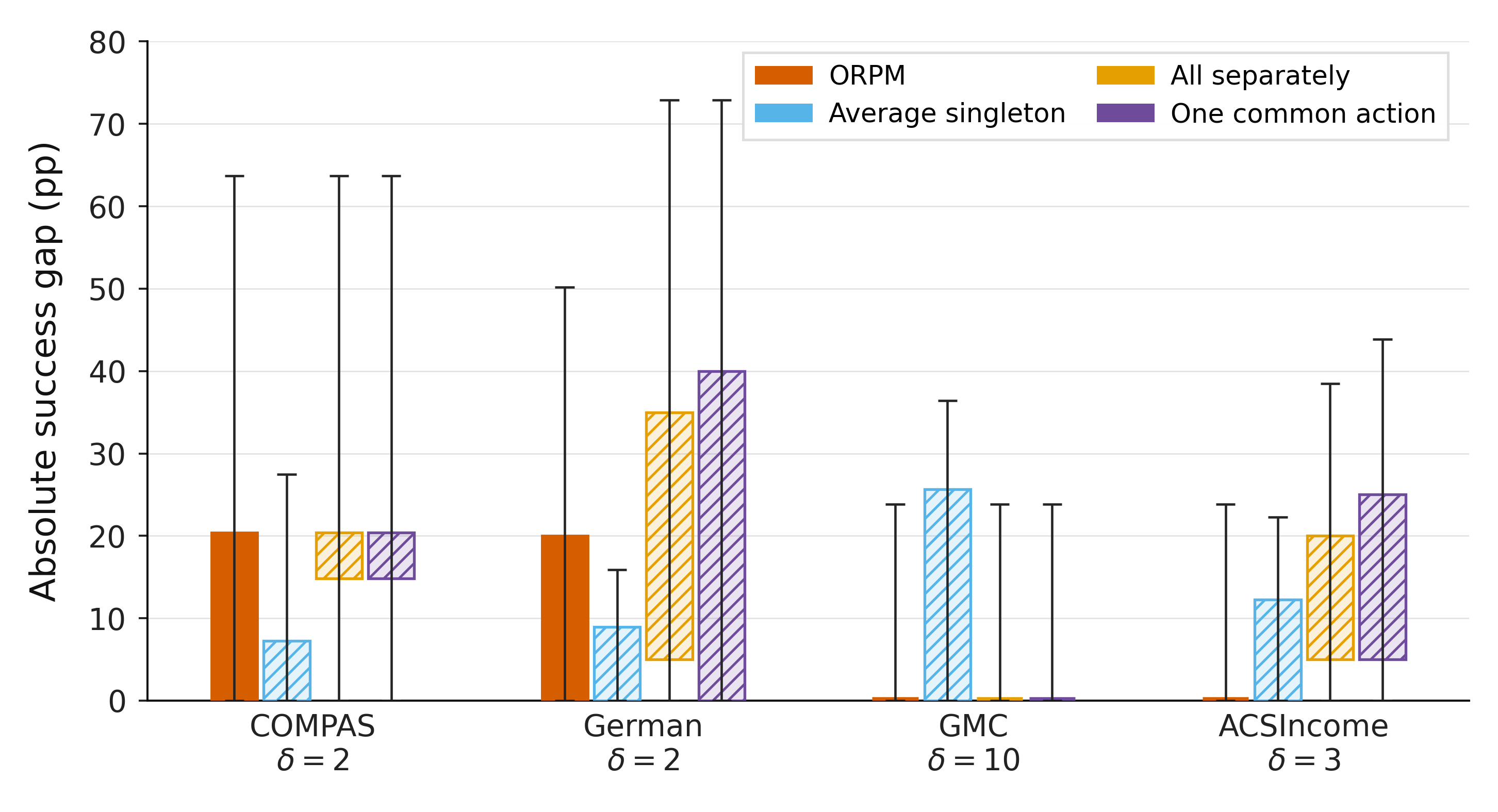}
\caption{Neural competing set, medium budget regime.}
\end{subfigure}
\caption{The disparity of recourse success across different groups. Each subfigure corresponds to different budget values, with small budgets on the left and medium budgets on the right. In each subfigure, the bars track the disparity in recourse success for ORPM and baselines on each dataset.\label{fig:disparity-app}}
\end{figure}

Figure~\ref{fig:disparity-app} depicts the disparity gap of ORPM
and baselines for the neural settings. 
On ACSIncome, ORPM's gap decreases from 15 percentage points at budget 1 to zero at budget 3, below the optimization ranges identified for the all-separately and common-action benchmarks at both budgets. The COMPAS gap decreases from 24.1 to 20.4 points as the budget increases from 1 to 2, whereas the German gap increases from 10 to 20 points. At the moderate budget, ORPM has larger gaps than the average singleton on both COMPAS and German. On GMC, ORPM, all separately, and common action have zero estimated disparity at both displayed budgets, but this accompanies only 10\% success in each group. Small gaps therefore need not indicate effective recourse. 

\subsubsection{Tree-Based Competing Sets}
\label{sec:app-exp-tree-based-competing-sets}
\begin{figure}[ht!]
\centering
\includegraphics[width=\linewidth]{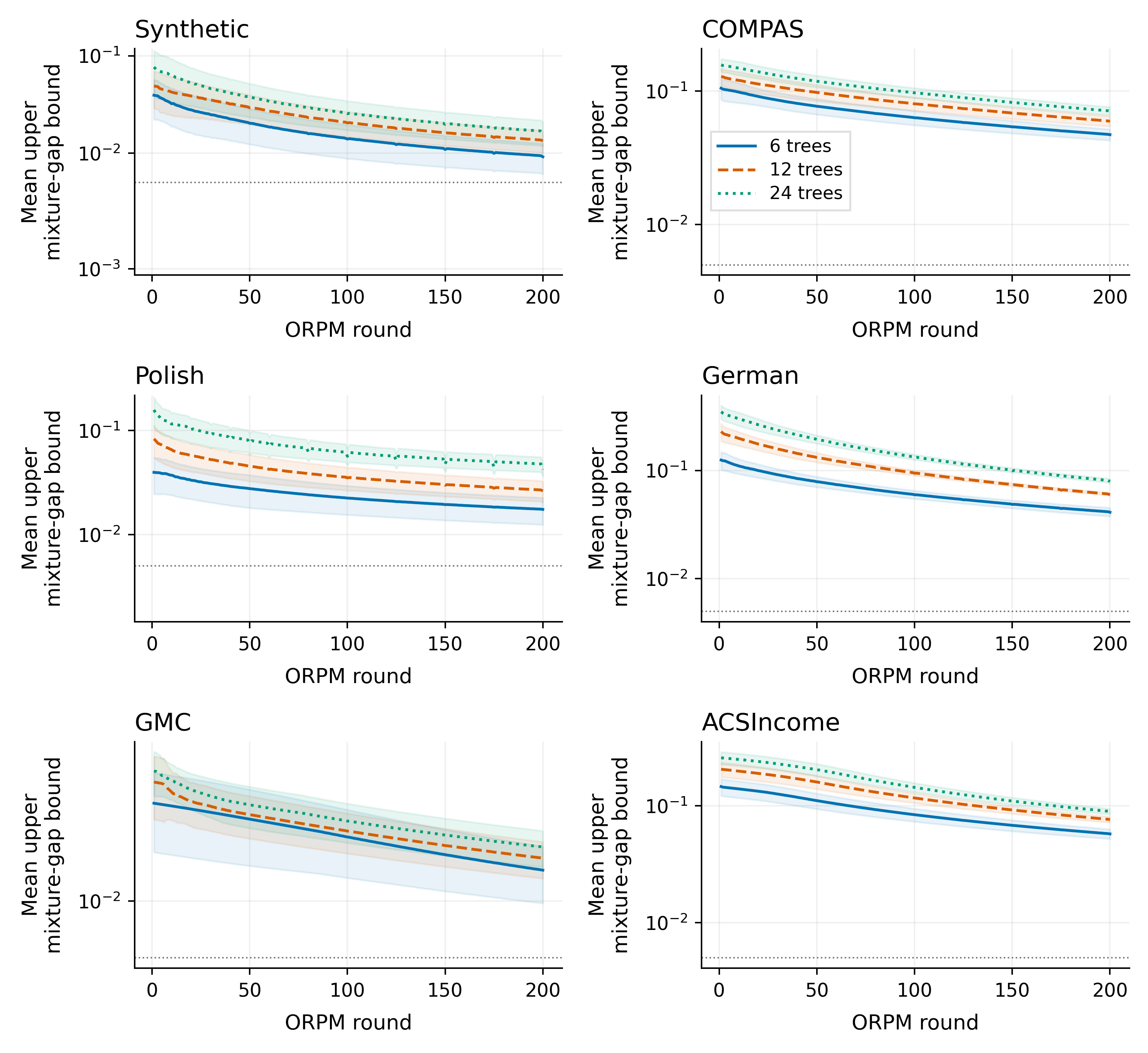}
\caption{Decision-tree convergence at $\delta=2$ for Synthetic, COMPAS, Polish and German, $\delta=10$ for GMC, and $\delta=5$ for ACSIncome. Curves show the mean upper bound on the empirical-response-mixture gap for 6, 12, and 24 trees (lower is better), across 10 seeds and 20 instances per seed. The vertical scale is linear near zero and logarithmic above 0.005; the dotted horizontal line marks 0.005. Shaded bands are pointwise 95\% Student-$t$ intervals over seed means. The mixture objective averages response losses, not feature vectors, and is distinct from deterministic joint validity.}
\label{fig:tree-gap-app}
\end{figure}

\begin{figure}[ht!]
\centering
\includegraphics[width=\linewidth]{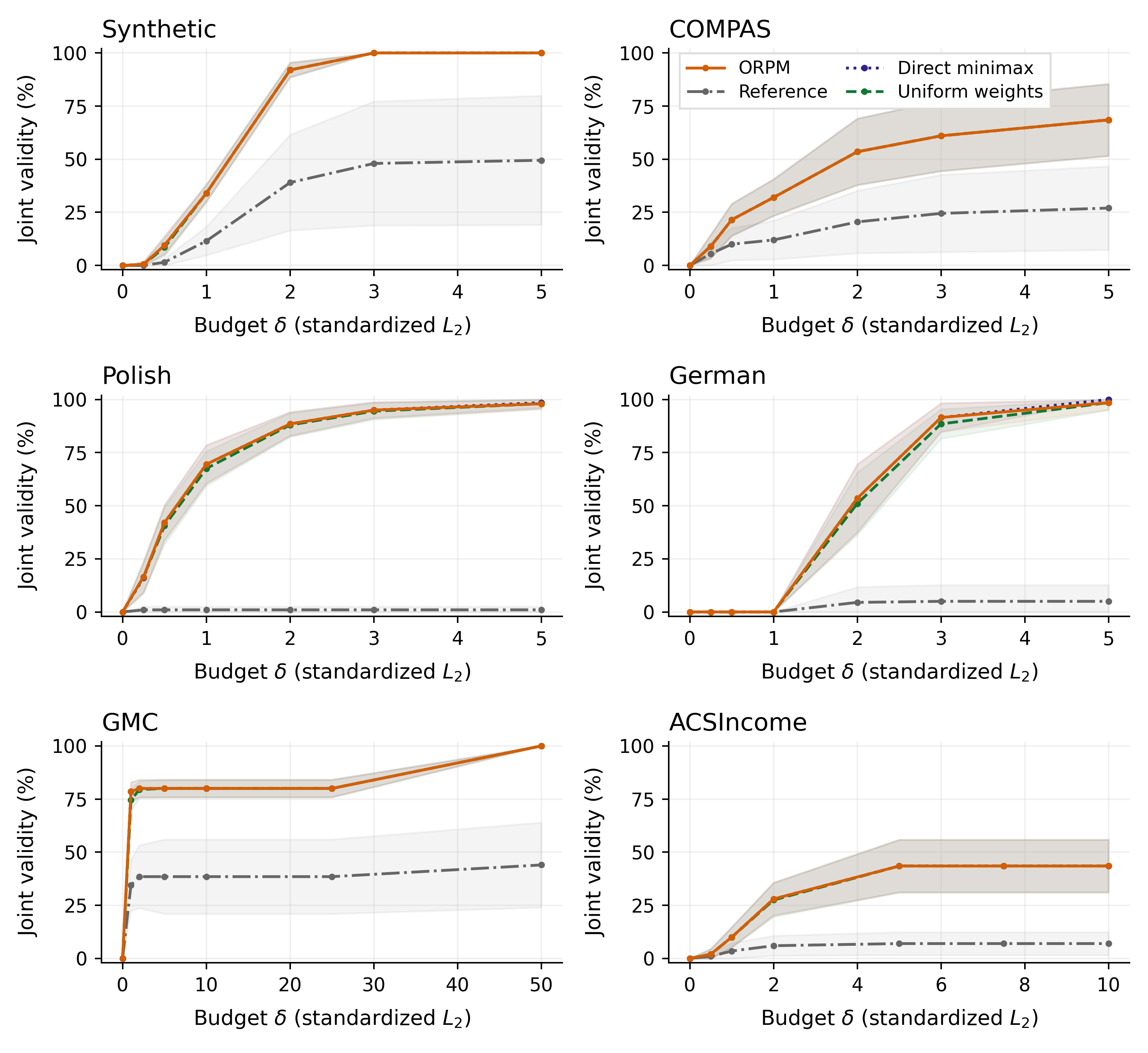}
\caption{Expanded decision-tree budget-validity comparison on all six datasets with 12 competing trees, 10 seeds, and 20 instances per seed. ORPM uses 100 one-second response calls; uniform weighting and direct minimax use one 30-second solve each; the single-tree reference uses leaf enumeration. Shading is pointwise 95\% Student-$t$ uncertainty over seed means, clipped to $[0,100]\%$ for display. }
\label{fig:tree-tradeoff-app}
\end{figure}

\begin{figure}[ht!]
\centering
\includegraphics[width=\linewidth]{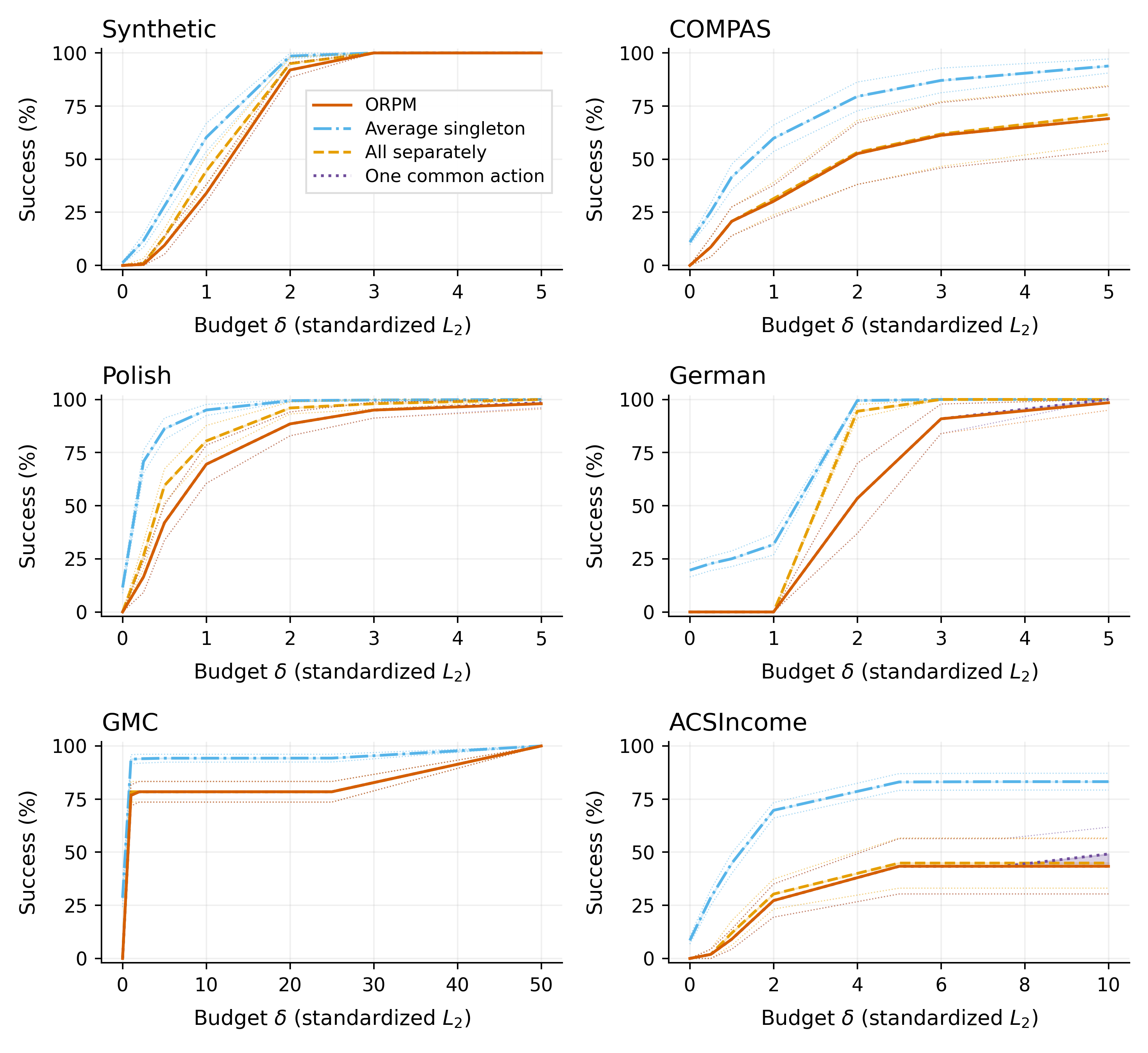}
\caption{Cost-of-multiplicity success curves on all six datasets with twelve decision trees with 10 seeds and 20 instances per seed. Dark fill denotes optimization identification ranges and thin dotted boundaries denote pointwise 95\% sampling limits at their endpoints. This cohort preserves eleven cases outside the trade-off study's standard prefix (1,189/1,200 overlap), so the overall curves need not coincide with Figure~\ref{fig:tree-tradeoff-app}.}
\label{fig:tree-cost-app}
\end{figure}

Figures~\ref{fig:tree-gap-app}-\ref{fig:tree-cost-app} extend the decision-tree results across datasets, using 10 seeds and 20 instances per seed. At the middle budget, mixture-gap certification varies markedly: Polish with twelve trees has no certified attainment of 0.005 by round 200, and COMPAS and ACSIncome also have none, despite positive deterministic validity in the budget experiments (see Figure~\ref{fig:tree-gap-app}). 
In the budget-validity comparison, increasing $\delta$ raises ORPM's observed joint validity on COMPAS with twelve trees from 21.5\% at $\delta=0.5$ to 68.5\% at $\delta=5$, whereas ACSIncome remains at 43.5\% from $\delta=5$ to $10$. Larger budgets therefore improve success in some settings but do not uniformly eliminate failures (see Figure~\ref{fig:tree-tradeoff-app}).  We use additional baselines for the trade-off experiment. 
Let $p_i(z)$ be tree $i$’s favorable-class probability and $\mathcal K_\delta(x)$ the allowed changes within budget $\delta$.
The baseline uniform weighting minimizes the average squared probability loss:
  $
  \min_{z\in\mathcal K_\delta(x)}\frac1m\sum_{i=1}^m(1-p_i(z))^2.
  $
  Every tree receives equal weight. Good average performance can still leave some trees rejecting the action.
The baseline direct minimax minimizes the largest loss:
  $
  \min_{z\in\mathcal K_\delta(x)}\max_i(1-p_i(z))^2.
  $
  It directly targets the least favorable tree, the same objective as ORPM approaches through repeated weighted responses. Uniform weighting and direct minimax each receive one 30-second, single-thread solve, and their returned solutions are not necessarily globally optimal.
The baseline single-tree reference generates an action using only the first tree, then evaluates it against the entire competing set. If that tree already accepts the factual point, it returns the factual point. Otherwise, it enumerates reachable leaves and selects the highest favorable-class probability, breaking ties by smaller $L_2$ distance.
Finally, requiring a common action can be substantially harder than allowing separate recourse: on German with twelve trees at $\delta=2$, average-singleton success is 99.5\%, all-separate feasibility 94.5\%, and common-action and ORPM success 53.4\% (see Figure~\ref{fig:tree-cost-app}). 

\end{document}